\documentclass[12pt,journal,letterpaper,onecolumn,twoside]{IEEEtran}

\usepackage{graphicx}  
\usepackage{epsfig}
\usepackage{psfig}

\usepackage{amsmath}   

\newcommand{\bm}[1]{\mbox{\boldmath{$#1$}}}

\usepackage{amsfonts,amssymb}

\begin{document}

\title{ANSIG --- An Analytic Signature for Arbitrary 2D~Shapes (or Bags of Unlabeled Points)}
\author{Jos\'e J.~Rodrigues,~\IEEEmembership{Student Member,~IEEE,}
        Jo\~ao M.~F.~Xavier,~\IEEEmembership{Member,~IEEE,}\\
        and Pedro M.~Q.~Aguiar,~\IEEEmembership{Senior Member,~IEEE}
\thanks{The authors are with the Institute for Systems and Robotics (ISR), Instituto
Superior T\'{e}cnico (IST), Av.~Rovisco Pais, 1049-001 Lisboa, Portugal.
Contact author: Pedro M. Q. Aguiar. E-mail: {\tt
aguiar@isr.ist.utl.pt}.}
\thanks{This work was partially supported by the Portuguese Foundation for Science and Technology
(FCT), under ISR/IST plurianual funding (POSC program, FEDER), and
grants MODI-PTDC/EEA-ACR/72201/2006 and
SIPM-PTDC/EEA-ACR/73749/2006.}}



\maketitle

\begin{abstract}
In image analysis, many tasks require representing
two-dimensional~(2D) shape, often specified by a set of 2D~points,
for comparison purposes. The challenge of the representation is that
it must not only capture the characteristics of the shape but also
be invariant to relevant transformations. Invariance to geometric
transformations, such as translation, rotation, and scale, has
received attention in the past, usually under the assumption that
the points are previously labeled, {\it i.e.}, that the shape is
characterized by an {\em ordered} set of landmarks. However, in many
practical scenarios, the points describing the shape are obtained
from automatic processes, {\it e.g.}, edge or corner detection, thus
without labels or natural ordering. Obviously, the combinatorial
problem of computing the correspondences between the points of two
shapes in the presence of the aforementioned geometrical distortions
becomes a quagmire when the number of points is large. We circumvent
this problem by representing shapes in a way that is {\em invariant
to the permutation} of the landmarks, {\it i.e.}, we represent {\em
bags of unlabeled 2D points}. Within our framework, a shape is
mapped to an analytic function on the complex plane, leading to what
we call its analytic signature~(ANSIG). To store an ANSIG, it
suffices to sample it along a closed contour in the complex plane.
We show that the ANSIG is a maximal invariant with respect to the
permutation group, {\it i.e.}, that different shapes have different
ANSIGs and shapes that differ by a permutation (or re-labeling) of
the landmarks have the same ANSIG. We further show how easy it is to
factor out geometric transformations when comparing shapes using the
ANSIG representation. Finally, we illustrate these capabilities with
shape-based image classification experiments.
\end{abstract}


\section{Introduction}

This paper deals with the representation of two-dimensional~(2D)
shape. In our context, a 2D~shape is described by the 2D~coordinates
of a set of {\em unlabeled} points, or landmarks. We seek efficient
ways to represent such sets, in particular we seek representations
that are suitable to be used in shape-based recognition tasks.
Besides being discriminative, those representations should be
invariant to (or, at least, deal well with) shape-preserving
geometric transformations. Above all, such representations must deal
with the fact that the landmarks do not have labels. This means that
the representation should be invariant to the order by which the
landmarks are stored, since different orderings of the same set of
landmarks represent the same shape. We thus focus on developing {\em
permutation invariant}, or {\em label invariant}, representations
for 2D~shape.

\subsection{Shape representation}

Many objects are primarily recognized by their shape, rather than
their color or texture~\cite{biederman87,mumford91}. Although this
fact has been confirmed by surveys showing that users would prefer
to retrieve images from shape queries~\cite{schomaker99}, the
majority of content-based image retrieval systems still use color
and texture features to compare images. In fact, shape-based
classification proved to be a very hard task, remaining an open
problem, underlying which is the fundamental question of how to
represent shape.

When the shape is described by a set of {\em labeled} landmarks, an
established theory provides tools to cope with geometric
transformations and shape variations: the {\em statistical theory of
shape}~\cite{small97,kendall99}. Although this theory has lead to
significant results when the shapes to compare are characterized by
feature points whose correspondences from image to image can be
obtained, in many practical scenarios, such correspondences are not
available. We thus focus on {\em unlabeled}~data.

The majority of the methods to cope with shapes described by
unlabeled sets of points focus on representing a ``blob", {\it
i.e.}, a shape that is a simply-connected set of points. A number of
techniques, usually called {\em region-based}, describe these shapes
by using moment descriptors, {\it e.g.}, geometrical~\cite{hu62},
Legendre~\cite{teague80}, Zernike~\cite{teague80,khotanzad90}, and
Tchebichef~\cite{mukundan01} moments. Other approaches, {\em
contour-based}, represent the boundary of the shape using, {\it
e.g.}, curvature scale space~\cite{mokhtarian92},
wavelets~\cite{chauang96}, contour displacements~\cite{adamek04},
splines~\cite{dierckx95}, or Fourier
descriptors~\cite{kauppinen95,zhang05,bartolini05}. Some of these
representations exhibit desired invariance to geometric
transformations but they are restricted to shapes well described by
closed contours.

In image analysis, shape cues come primarily from the image edges.
In general, it is hard to extract complete contours when dealing
with real images~\cite{ghosh05}, thus researchers also developed
local shape descriptors that, at each point of the shape, capture
the relative distribution of the remaining points, {\it e.g.}, {\em
shape contexts}~\cite{belongie02} and {\em distance
multisets}~\cite{grigorescu03}. These local representations proved
to cope with the contour discontinuities typical of the output of
automatic edge detection processes but do not deal with general
shapes and geometric transformations. In this paper we seek ways to
describe shapes characterized by arbitrary sets of points.

A number of approaches to deal with shapes described by general sets
of unlabeled points are motivated by the need to register the
corresponding images, {\it i.e.}, to compute the rigid
transformation that best ``aligns" them. The majority of these
registration methods are inspired by the fact that the solution for
the rigid transformation is easily computed when the labels, {\it
i.e.}, the point correspondences, are known --- the {\em Procrustes
matching problem}. To cope with unlabeled points, they develop
iterative algorithms that compute, in alternate steps, the rigid
registration parameters and the point correspondences. One of the
better known examples is the {\em Iterative Closest Point~(ICP)}
algorithm~\cite{besl92}. More recently, others have proposed
statistical methods that use ``soft"
correspondences~\cite{chui00,luo02,kent04,mcneil06nips,mcneil06icip,mcneil06cvpr},
leading to two-step iterative algorithms based on {\em
Expectation-Maximization~(EM)}~\cite{KN:McLachlanKrishnan}. Although
these methods have dealt with challenging scenarios, including shape
part decomposition~\cite{mcneil06nips} and nonrigid
registration~\cite{chui00}, they have the limitations of iterative
algorithms, including the uncertain convergence and the sensitivity
to initialization.

Rather than attempting to infer the labels of the points describing
the shapes to compare, {\it i.e.}, rather than computing the
permutation between the two sets of points, we seek {\em permutation
invariant} representations. The relevance of permutation invariance
in learning tasks has been recently pointed
out~\cite{jebara03,jebara03aistat,jebara04,jebara06,jebara07}. In
these works, the permutation is factored out after being computed as
the solution of a convex optimization problem over the permutation
matrices. However, this formulation does not deal with geometric
transformations such as the rigid rotation of the point set. A
permutation invariant representation recently proposed describes a
shape by the set of all pairwise distances between the shape
points~\cite{boutin04,lee06,boutin07}. Naturally, the dimensionality
of the this representation limits its applicability to shapes
described by small sets of landmarks, like fingerprint minutiae. We
focus on large point sets, as those typically obtained from the
edges of real images.

\subsection{Our approach: the analytic signature of a shape}

Our approach is rooted on a new {\it permutation invariant}, or {\it
label invariant}, representation for sets of 2D~points. We represent
a 2D~shape by what we call its analytic signature~(ANSIG), an
analytic function defined over the complex plane. We show that
shapes that differ by a re-labeling of the landmarks ({\it i.e.}, by
a re-ordering of the vector containing the point set) have the same
ANSIG. Thus, the ANSIG representation is {\it permutation
invariant}, or {\it label invariant}. Furthermore, we show that this
representation enables discriminating different shapes, {\it i.e.},
geometrically different shapes are in fact represented by different
ANSIGs.

The impact of the ANSIG representation in shape-based classification
tasks is obvious: while methods that use less powerful ({\it i.e.},
``less invariant") representations usually require several
prototypes of each class ({\it i.e.}, several training examples to
perform statistical learning) and sophisticated classification
schemes, in our case, shape-based classification boils down to
comparing the ANSIG of a candidate shape with the one of a single
prototype shape per class. As any analytic function, the ANSIG of a
shape is completely described by the values it takes in a closed
contour on the complex plane. Thus, we store ANSIGs by simply
sampling them on the unit-circle. To compare ANSIGs, {\it i.e.}, to
evaluate shape similarity, it suffices to compute the angle between
the vectors collecting those samples.


Although the ANSIG representation is not invariant with respect to
geometric transformations, such as translation, rotation, and scale,
they are easily taken care of. While an adequate pre-processing step
factors out translation and scale, the rotation that best aligns two
shapes is easily obtained from their ANSIGs. In fact, the rotation
that maximizes the above mentioned similarity is efficiently
computed by using the Fast Fourier Transform~(FFT) algorithm.

In many practical scenarios the set of points describing a shape is
obtained from an image, as the output of an automatic process, {\it
e.g.}, edge detection or simple thresholding. Thus, it is natural to
expect that, besides being noisy, the sets of 2D~points to compare
have distinct cardinality (particularly when comparing shapes
obtained from images of different sizes). We show how the ANSIG
representation also deals well with this kind of perturbations.

To illustrate the invariance properties of the ANSIG representation,
we present experiments with synthetic data. To demonstrate its
usefulness in shape-based classification, we report results obtained
with real images. Other experiments are in preliminary versions of
this work~\cite{ANSIG1,ANSIG2}.

\subsection{Paper organization}

In Section~\ref{sec:perminv}, we address the construction of
permutation invariant representations for a set of 2D~points. We
show that what we call the analytic signature~(ANSIG) of a 2D~shape
is such a representation and discuss how to store it.
Section~\ref{sec:geometric} describes how shape-preserving
transformations are easily taken care of by the ANSIG
representation. In Section~\ref{sec:implementation}, we discuss
implementation issues that arise from the need to efficiently
compare ANSIGs, for the purpose of shape-based classification.
Section~\ref{sec:exp}, contains experiments and
Section~\ref{sec:conc} concludes the paper.

\section{Permutation invariance: the analytic signature of a shape}
\label{sec:perminv}

We consider that a 2D~shape is described by a set of $n$ unlabeled
points, or landmarks, in the plane. Thus, under the usual
identification of ${\mathbb R}^2$ with the complex plane ${\mathbb
C}$, a 2D~shape can be represented by a vector
\begin{equation}
{\boldsymbol z} = \begin{bmatrix} z_1 \\ z_2 \\ \vdots \\ z_n
\end{bmatrix} = \begin{bmatrix} x_1 + i y_1 \\ x_2 + i y_2 \\ \vdots
\\ x_n + i y_n \end{bmatrix} \in {\mathbb C}^n. \label{shapevector}
\end{equation}
However, since there are not labels for the landmarks, the order by
which they are stored is irrelevant and the choice
in~(\ref{shapevector}) is not unique, {\it i.e.}, the same shape is
equivalently represented by any vector in the set
\begin{equation}
\{ {\boldsymbol \Pi} {\boldsymbol z}:\,{\boldsymbol \Pi} \in \Pi(n)
\}, \label{orbit}
\end{equation}
where $\Pi(n)$ denotes the set of $n\times n$ permutation
matrices\footnote{A permutation matrix is a square matrix with
exactly one entry equal to 1 per row and per column and the
remaining entries equal to 0. Each element in $\Pi(n)$ represents a
specific permutation: when multiplied by a vector ${\boldsymbol z}$
produces a vector ${\boldsymbol \Pi} {\boldsymbol z}$ that has the
same entries of ${\boldsymbol z}$ but arranged in a possibly
different order. The cardinality of $\Pi(n)$ is~$n!$.}.

Our goal is to develop efficient ways to represent the vector set
in~(\ref{orbit}).

\subsection{Shapes as points in a quotient space}

In group theory parlance, we say that a shape defined as a set like
in~(\ref{orbit}) is a point in the quotient space ${\mathbb C}^n
/\Pi(n)$. Here, we view ${\Pi}(n)$ as a group, whose action on
${\mathbb C}^n$ is the left matrix multiplication, {\it i.e.}, the
group operation ${\Pi}(n)\times{\mathbb C}^n\rightarrow{\mathbb
C}^n$ is simply ${\boldsymbol \Pi}\cdot{\boldsymbol z}={\boldsymbol
\Pi}{\boldsymbol z}$.
This group action induces a partition of~${\mathbb C}^n$ into
disjoint orbits: the orbit passing through~${\boldsymbol z}$
collects all possible ways of representing the shape in vector
${\boldsymbol z}$, {\it i.e.}, it is the set in~(\ref{orbit}). Each
shape corresponds then to an orbit and we denote the one
corresponding to the shape in vector ${\boldsymbol z}$, {\it i.e.},
the orbit passing through~${\boldsymbol z}$, by $[{\boldsymbol z}]$.
The quotient space is then the set of shapes, {\it i.e.}, the set of
all orbits:
\begin{equation}{\mathbb C}^n /
\Pi(n) = \{ [ {\boldsymbol z} ]:\, {\boldsymbol z} \in {\mathbb C}^n
\}. \end{equation}

\begin{figure}[ht]
\centerline{\psfig{figure=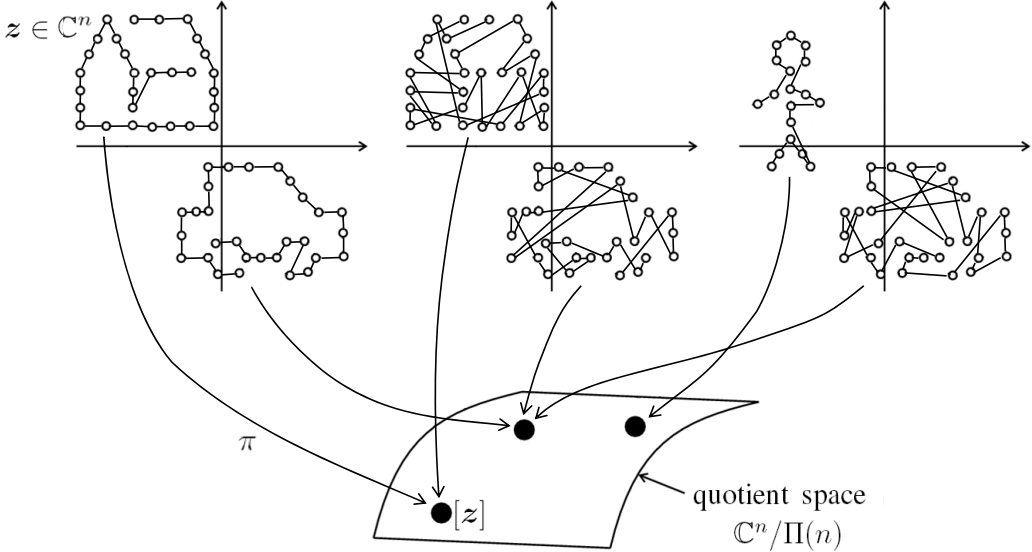,width=14cm}}
\vspace*{-.5cm}\caption{Shapes as points in a quotient space. In the
shape drawings, the line connecting the landmarks indicates their
labeling, {\it i.e.}, the order by which the points are stored.
While shape instances that only differ by a re-labeling of the
landmarks are mapped to the same point in the quotient space,
geometrically different shapes are mapped to distinct
points.}\label{fig:gtheory}
\end{figure}

Fig.~\ref{fig:gtheory} illustrates the scenario, where the canonical
(or quotient) map $\pi\,:\,{\mathbb C}^n \rightarrow {\mathbb C}^n /
\Pi(n)$ maps each shape vector instance to its orbit, {\it i.e.},
\begin{equation}
\pi({\boldsymbol z}) = [ {\boldsymbol z} ]=\{ {\boldsymbol \Pi}
{\boldsymbol z}:\,{\boldsymbol \Pi} \in \Pi(n) \}.
\end{equation}
From the definitions above, it is clear that two vectors
${\boldsymbol z}$ and ${\boldsymbol w}$ that represent the same
shape, {\it i.e.}, that correspond to distinct labelings of the same
landmarks, are mapped by $\pi$ to the same point in the quotient
space: $\pi({\boldsymbol z}) = \pi( {\boldsymbol w})$. Also, two
vectors ${\boldsymbol z}$ and ${\boldsymbol w}$ that do not
represent the same shape, {\it i.e.}, whose entries are not simply
related by a permutation, will be mapped by $\pi$ to distinct points
in the quotient space: $\pi({\boldsymbol z}) \neq \pi( {\boldsymbol
w} )$, see Fig.~\ref{fig:gtheory}. This characteristic is usually
referred to as a \textit{maximal invariance} property, meaning that
\begin{equation}\pi({\boldsymbol
z})=\pi({\boldsymbol w}) \; \Leftrightarrow \; \mbox{${\boldsymbol
z}$ and ${\boldsymbol w}$ represent the same
shape}\,.\label{eq:maxinv}
\end{equation}

Although the maximal invariance property in \eqref{eq:maxinv} is
precisely what we look for (the map $\pi$ detects whether or not
${\boldsymbol z}$ and ${\boldsymbol w}$ represent the same shape),
this mechanism is hardly implementable, due to the rather abstract
nature of the quotient space and map.

\subsection{Polynomial signature of a shape}

We now develop a version of the objects introduced above that is
suitable for use in practice. In particular, we propose to replace
the abstract map $\pi\,:\,{\mathbb C}^n \rightarrow {\mathbb C}^n /
\Pi(n)$ by a map from ${\mathbb C}^n$ to the set ${\mathcal A}$ of
analytic functions on the complex plane:
\begin{equation}
{\mathcal A} = \{ f:\,{\mathbb C} \rightarrow {\mathbb C}:\,f \mbox{
is analytic} \}.\end{equation} Our surrogate map must exhibit
maximal invariance with respect to the group~$\Pi(n)$, just like the
quotient map~$\pi$ in~\eqref{eq:maxinv}. This means that it must map
${\boldsymbol z}$ and ${\boldsymbol w}$ to the same analytic
function if and only if ${\boldsymbol z}$ and ${\boldsymbol w}$
represent the same shape, {\it i.e.}, iff $[{\boldsymbol
z}]=[{\boldsymbol w}]$, {\it i.e.}, iff ${\boldsymbol z} =
{\boldsymbol \Pi} {\boldsymbol w}$, for some permutation
matrix~${\boldsymbol \Pi}$.

A possible choice for the map from ${\mathbb C}^n$ to ${\mathcal A}$
is such that a shape vector ${\boldsymbol z}$ is mapped to a complex
polynomial $p({\boldsymbol z},\cdot)$, leading to what we call the
{\em polynomial signature} of a shape. This map ${\boldsymbol z}
\mapsto p({\boldsymbol z},\cdot)$ is defined by the following
expressions, where $\xi$ is a dummy complex variable:
\begin{equation} p({\boldsymbol z},\xi) = \prod_{m=1}^n
{(\xi-z_m)}\,. \label{polynomial}
\end{equation}

\noindent{\bf Maximal invariance of the polynomial signature.} It is
clear that the invariance with respect to the permutation group
holds. In fact, $p({\boldsymbol z},\cdot) = p({\boldsymbol
w},\cdot)$ whenever ${\boldsymbol z}, {\boldsymbol w}$  differ only
by a permutation of their entries, {\it i.e.}, whenever
${\boldsymbol z} = {\boldsymbol \Pi} {\boldsymbol w}$, because, from
definition~(\ref{polynomial}), it follows immediately that
$p({\boldsymbol \Pi}{\boldsymbol w},\cdot)=p({\boldsymbol
w},\cdot)$, due to the commutativity and associativity of the
complex product.

To establish the maximal invariance of the polynomial signature,
consider two vectors ${\boldsymbol z}$ and ${\boldsymbol w}$ that
have equal signatures, {\it i.e.}, such that the polynomials
$p({\boldsymbol z},\cdot)$ and $p({\boldsymbol w},\cdot)$ are equal.
This equality means that those polynomials share the same system of
roots, including their multiplicities. Since,
from~\eqref{polynomial}, the roots of a polynomial signature are the
complex points in the shape vector, the equality of $p({\boldsymbol
z},\cdot)$ and $p({\boldsymbol w},\cdot)$ implies that the
(multi-)sets $\{ z_1, z_2, \ldots, z_n \}$ and $\{ w_1, w_2, \ldots,
w_n \}$ are equal. This way we conclude that the vectors
${\boldsymbol z} = [\,z_1\, z_2\, \cdots\, z_n\,]^T$ and
${\boldsymbol w} = [\,w_1\, w_2\, \cdots\, w_n\,]^T$ are equal up to
a permutation, thus proving the maximal invariance of the polynomial
signature in~(\ref{polynomial}).

The practical use of the polynomial signature is limited by its
numerical stability. In fact, since a shape with $n$ points is
represented by a $n^{\mbox{\small th}}$ degree polynomial, when $n$
is large, the signature exhibits extremely large variations,
resulting numerically unstable ({\it e.g.}, sensitive to the noise).

\subsection{The analytic signature (ANSIG) of a shape}
\label{subsec:expansig}

Inspired by the polynomial signature, we now develop another
analytic map $a\,:\,{\mathbb C}^n \rightarrow {\mathcal A}$ that,
besides maintaining the maximal invariance property, results
numerically stable and robust, leading to what we call the analytic
signature~(ANSIG) of a shape. This map ${\boldsymbol z} \mapsto
a({\boldsymbol z},\cdot)$, {\it i.e.}, the~ANSIG of the shape
described by vector ${\boldsymbol z}$, is defined by
\begin{equation} a({\boldsymbol z},\xi) = \frac{1}{n} \sum_{m=1}^n
e^{z_m \xi}\,. \label{ansig}
\end{equation}

\noindent{\bf Maximal invariance of the ANSIG.} Just like for the
polynomial signature, the invariance of the ANSIG with respect to
the permutation group is obvious. In fact, from~(\ref{ansig}),
$a({\boldsymbol z},\cdot) = a({\boldsymbol w},\cdot)$, whenever
${\boldsymbol z} = {\boldsymbol \Pi} {\boldsymbol w}$, due to the
commutativity and associativity of the complex sum.

To establish the maximal invariance, consider two shape vectors
${\boldsymbol z}$ and ${\boldsymbol w}$ that have equal ANSIGs,
$a({\boldsymbol z},\cdot) = a({\boldsymbol w}, \cdot)$. We will show
that ${\boldsymbol z}$ and ${\boldsymbol w}$ have also the same
polynomial signature, $p({\boldsymbol z},\cdot) = p({\boldsymbol w},
\cdot)$, {\it i.e.}, that they represent the same shape, differing
only by a permutation of their entries. Consider then the equal
ANSIGs $a({\boldsymbol z},\cdot)$ and $a({\boldsymbol w}, \cdot)$,
given by \eqref{ansig}, as analytic functions on the complex plane.
Obviously, their derivatives at the origin also coincide:
\begin{equation}
\left. \frac{d^k}{d \xi^k} a({\boldsymbol z},\xi) \right|_{\xi = 0}
= \left. \frac{d^k}{d \xi^k} a({\boldsymbol w},\xi) \right|_{\xi =
0}\,, \quad k = 1, 2, \ldots, n. \label{proof1}
\end{equation}
Using the definition of the ANSIG map~$a$ in~(\ref{ansig}), we write
the system of equations~(\ref{proof1}) in terms of the entries of
the vectors ${\boldsymbol z}$ and ${\boldsymbol w}$:
\begin{equation}
z_1^k + z_2^k + \cdots + z_n^k  =  w_1^k + w_2^k + \cdots + w_n^k\,,
\quad k = 1, 2, \ldots, n. \label{proof2}
\end{equation}

We now show that the set of equalities \eqref{proof2} implies that
the vectors ${\boldsymbol z}$ and ${\boldsymbol w}$ have the same
polynomial signature, thus represent the same shape. Start by noting
that, since the $k^{\mbox{\small th}}$ moment of an $n^{\mbox{\small
th}}$ order polynomial with roots $\{r_1,r_2, \ldots, r_n\}$ is
given~by
\begin{equation}
    \mu_k=r_1^k+r_2^k+\cdots+r_n^k\,,\label{eq:moments}
\end{equation}
the system of equations in~(\ref{proof2}) expresses the equality of
the first $n$ moments of the polynomial signatures $p({\boldsymbol
z},\cdot)$ and $p({\boldsymbol w}, \cdot)$, see their definition
in~(\ref{polynomial}). We now use the so-called {\it Newton's
identities}, which relate the first $n$ moments $\left\{\mu_1,
\mu_2, \ldots,\mu_n\right\}$ of a polynomial
$a_0+a_1\xi+a_2\xi^2+\cdots+a_n\xi^n$, with its coefficients
$\left\{a_0,a_1,\ldots,a_n\right\}$, see, {\it e.g.},
\cite{hornandjohnson}:
%
\begin{equation}
a_n\mu_k+a_{n-1}\mu_{k-1}+\cdots+a_{n-k+1}\mu_1=-ka_{n-k}\;,\qquad
k=1,2,\ldots, n\,.\label{eq:identities}
\end{equation}
For monic polynomials, {\it i.e.}, when $a_n=1$, such is the case of
the polynomial signatures by construction, see~(\ref{polynomial}),
the {\it Newton's identities} \eqref{eq:identities} are written in
matrix format~as:
\begin{equation}
   \begin{bmatrix}
                                   1 & 0 & \cdots & 0 & 0 \\
                                   a_{n-1} & 1 & \ddots & 0 & 0 \\
                                \vdots & \vdots & \ddots & \ddots & \vdots \\
                                 a_2 & a_3 & \cdots & 1 & 0 \\
                                 a_1 & a_2 & \cdots & a_{n-1} & 1
   \end{bmatrix}
   \begin{bmatrix} \mu_1\\ \mu_2\\ \vdots\\ \mu_{n-1}\\ \mu_n
   \end{bmatrix}=-
   \begin{bmatrix} a_{n-1}\\ 2a_{n-2}\\ \vdots\\ (n-1)a_1\\
   na_0\end{bmatrix}\,.
   \end{equation}
From the structure of the matrix equality above, it is clear that
the moments of a polynomial uniquely determine its coefficients,
{\it i.e.}, they fully specify the polynomial. In fact, it is
immediate that $\mu_1$ determines $a_{n-1}$, $\mu_1$ and $\mu_2$
determine $a_{n-1}$ and $a_{n-2}$, etc.

%

The moment equalities in~(\ref{proof2}) imply then that the
polynomial signatures $p({\boldsymbol z},\cdot)$ and $p({\boldsymbol
w},\cdot)$ defined in \eqref{polynomial} are identical. Thus, using
the already established maximal invariance of the polynomial
signature, we conclude that ${\boldsymbol z} = [\,z_1\, z_2\,
\cdots\, z_n\,]^T$ and ${\boldsymbol w} = [\,w_1\, w_2\, \cdots\,
w_n\,]^T$ are equal up to a permutation, proving the maximal
invariance of our ANSIG map~$a$ in~(\ref{ansig}).

\subsection{Storing the ANSIG}

The ANSIG $a:\,{\mathbb C}^n \rightarrow {\mathcal A}$ maps a shape,
{\it i.e.}, a vector ${\boldsymbol z}$, to an analytic function
$a({\boldsymbol z},\cdot)$ on the complex plane. Since the space of
analytic functions is infinite-dimensional, our map seems inadequate
to a computer implementation. However, we can store any $f \in
{\mathcal A}$ by exploiting the well-known Cauchy's integral
formula, whose major consequence is that any analytic function is
unambiguously determined by the values it takes on a simple closed
contour, see, {\it e.g.}, \cite{ahlfors}. In particular, if we
choose the contour to be the unit-circle, {\it i.e.}, the
unit-radius circle centered at the origin, ${\mathbb S}^1 = \{ z \in
{\mathbb C}:\, | z | = 1 \}$, the analytic function $f$ is uniquely
determined by $\{ f(e^{j \varphi}):\,\varphi \in [0,2\pi] \}$. In
practice, this is approximated by sampling $f$ on a finite set of
$K$ points uniformly distributed in the unit-circle\footnote{In all
the experiments reported in this paper, we used $K=512$.}, {\it
i.e.}, on $\left\{ 1 , W_K, W_K^2, \ldots, W_K^{K-1} \right\}$,
where
\begin{equation}
W_K = e^{j \frac{2\pi}{K} }\,. \label{wk}
\end{equation}

In summary, we approximate the ANSIG map $a\,:\,{\mathbb C}^n
\rightarrow {\mathcal A}$ by its discrete counterpart
$a_K\,:\,{\mathbb C}^n \rightarrow {\mathbb C}^K$, where the
discrete version of the ANSIG of ${\boldsymbol z}$ is then given by
\begin{equation}
a_K({\boldsymbol z}) =\begin{bmatrix} a({\boldsymbol z},1) &
a({\boldsymbol z}, W_K) & a({\boldsymbol z}, W_K^2) &
 \cdots & a\left({\boldsymbol z}, W_K^{K-1} \right) \end{bmatrix}^T \,.
\label{discrete}
\end{equation}

\subsection{Illustration}

\begin{figure}[hbt]
  \centerline{\epsfig{figure=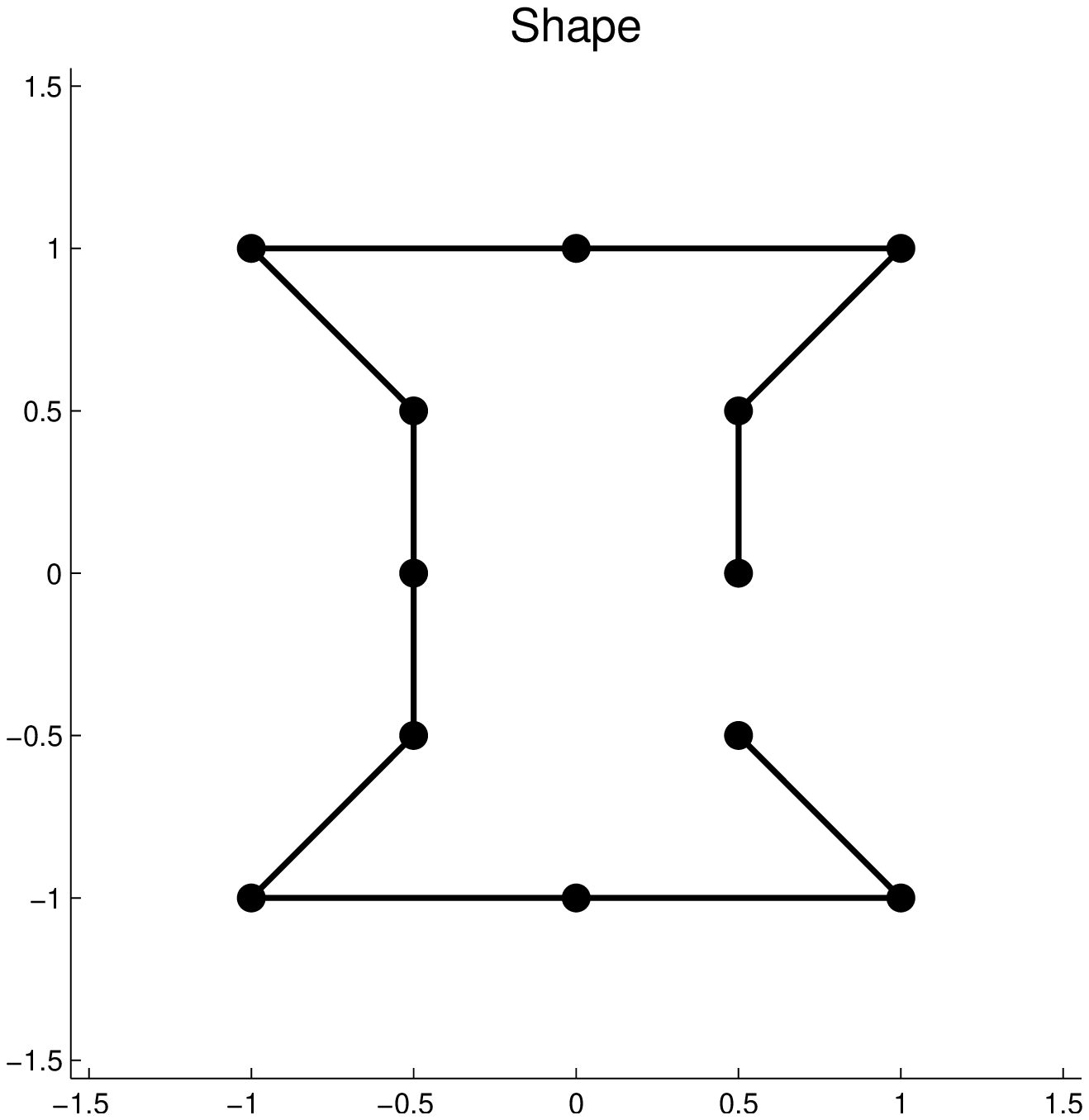, width=5cm}\hspace*{1cm}
  \epsfig{figure=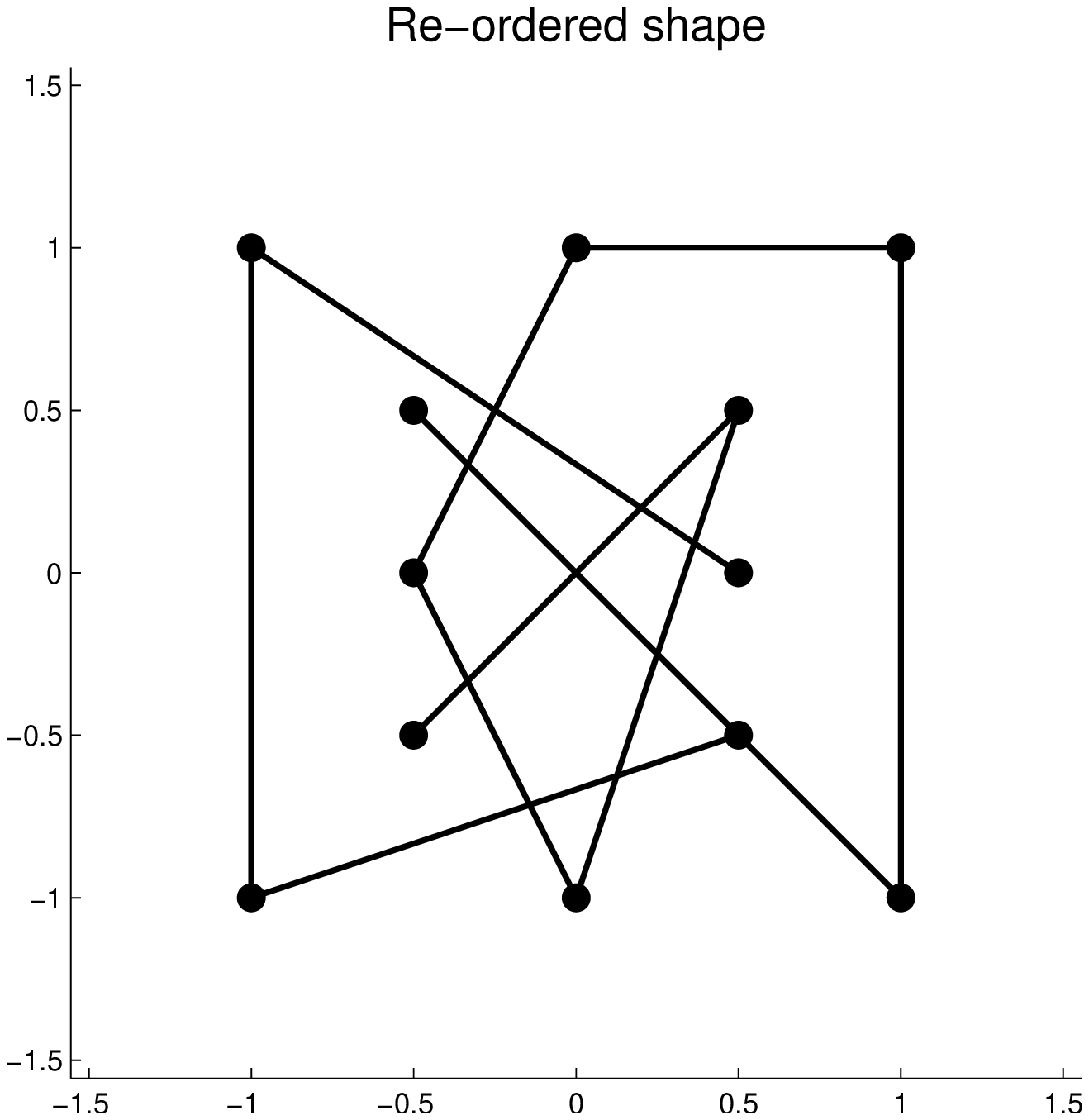, width=5cm}}
\vspace*{-.5cm}\caption{Two observations of the same shape,
described by the positions of a set of landmarks, here represented
by the black dots. The line connecting the dots, represented only
for illustrative purposes, indicates the order by which the points
are stored. Thus, although the left and right images represent the
same shape, this is not trivially inferred from the stored data.
\label{fig:perm_star}}
\end{figure}

To illustrate the invariance of the ANSIG to the landmark labels, we
use two shapes whose only difference is the order by which the
landmarks are stored. In Fig.~\ref{fig:perm_star}, we represent such
a pair of shapes. Note that the set of landmarks (black dots) of the
left image is the same than that of the right one. To indicate the
order by which the landmarks are stored, we use a line connecting
them, thus the images look different, illustrating the quagmire of
having to compare shapes when the correspondence between the sets of
landmarks is unknown.

\begin{figure}[hbt]
\centerline{\epsfig{figure=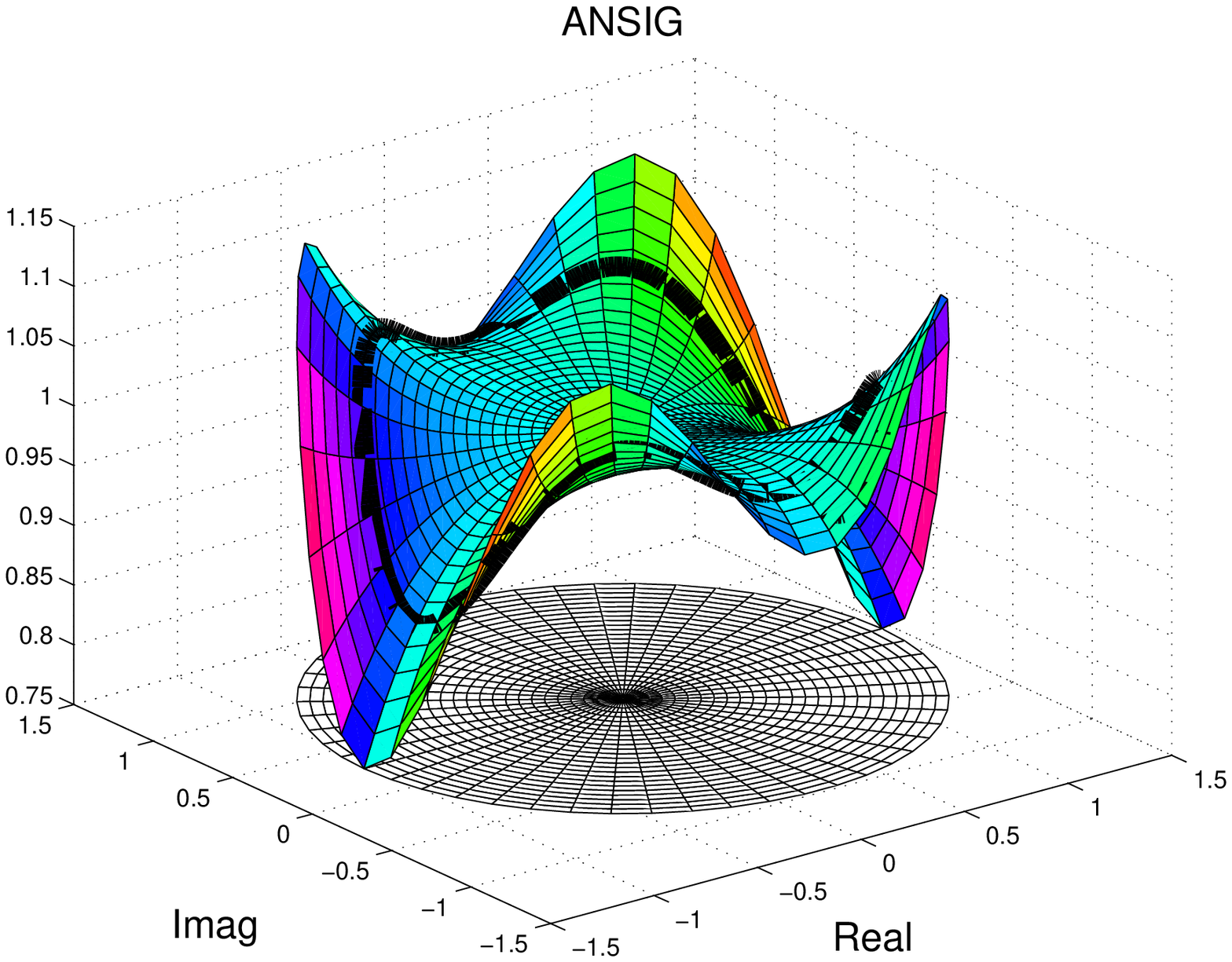,height=5cm}
\epsfig{figure=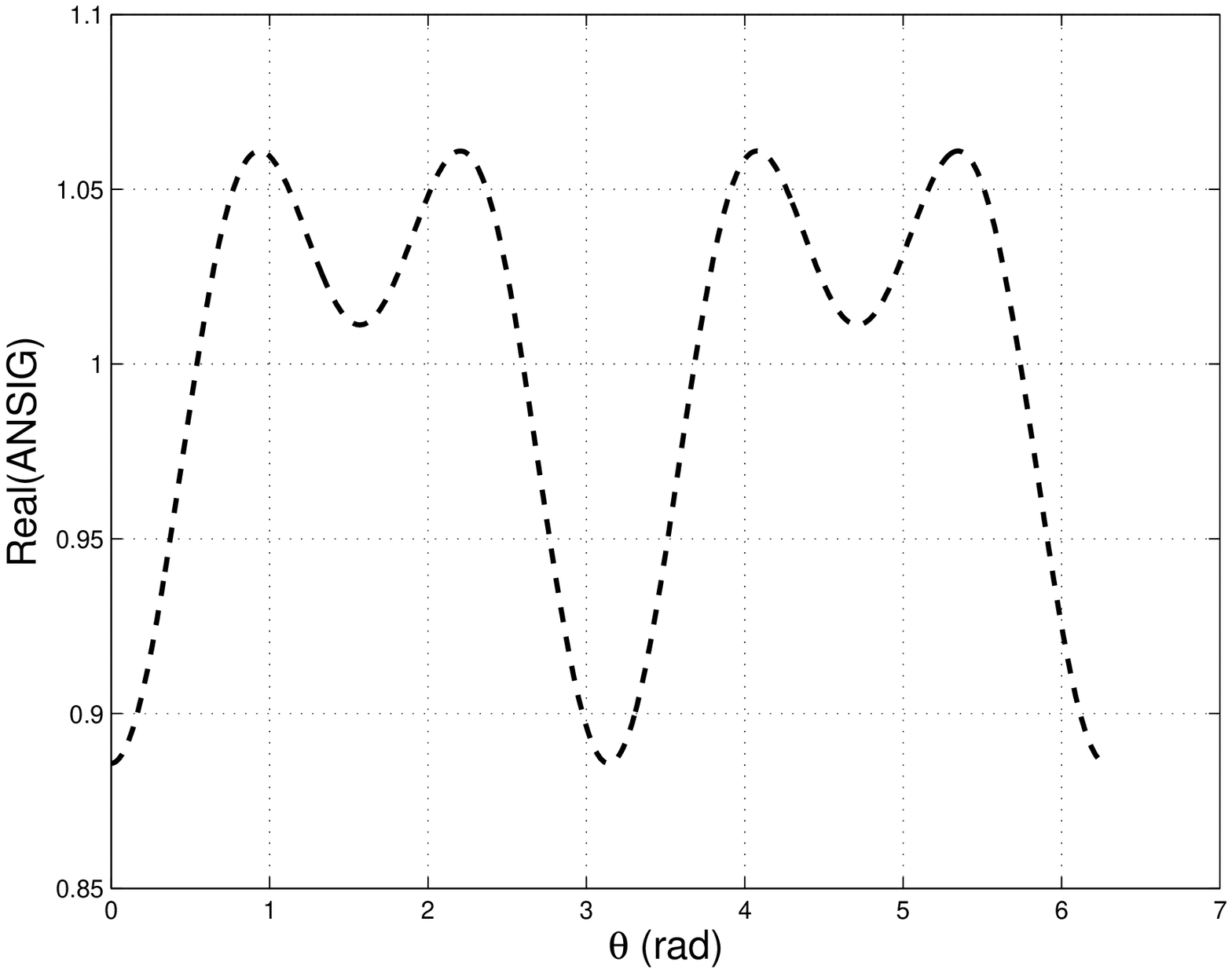,
height=4cm}\epsfig{figure=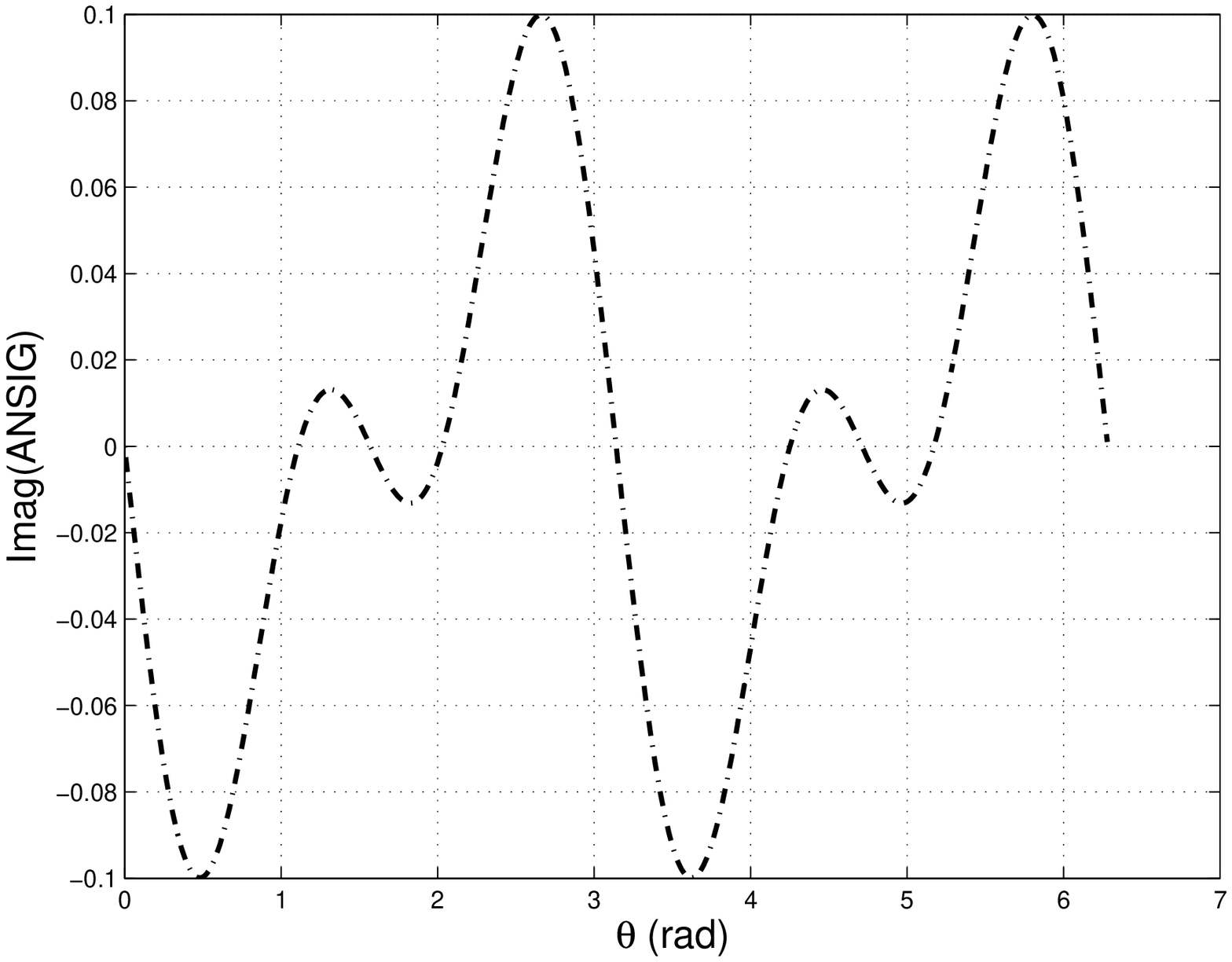,
height=4cm}} \vspace*{-.5cm}\caption{The ANSIG of the shapes in
Fig.~\ref{fig:perm_star}. In the left, the real part is represented
by the surface high and the imaginary part by its color. In the
middle and right plots, the discrete version of the ANSIG, {\it
i.e.}, its unit-circle samples. Although the vectors containing the
shapes in Fig.~\ref{fig:perm_star} are distinct, due to re-ordering,
their ANSIGs results equal, as
desired.\label{fig:perm_star_ANSIG_plane}}
\end{figure}

We compute the ANSIG of each of the shapes in
Fig.~\ref{fig:perm_star}. As expected, although the vectors
containing the landmarks for these two shapes are completely
different (due to re-ordering), we obtain the same ANSIG, which is
represented in Fig.~\ref{fig:perm_star_ANSIG_plane}. This
illustrates that the ANSIG is a permutation invariant
representation. Obviously, the discrete counterparts of the ANSIGs,
{\it i.e.}, their samples on the unit-circle, also represented in
Fig.~\ref{fig:perm_star_ANSIG_plane}, are also equal. Since the
ANSIGs of geometrically different shapes result distinct, this
signature is a substitute of the abstract quotient map $\pi$ in
Fig.~\ref{fig:gtheory}. The reader may also wonder why the middle
and right plots in Fig.~\ref{fig:perm_star_ANSIG_plane} are periodic
(period~$\pi$). This fact is due to the rotational symmetry (of
$\pi$) of the shapes represented in Fig.~\ref{fig:perm_star}. We
postpone to the following section a more detailed comment on this
aspect.

%

\section{Quotienting out shape-preserving geometric transformations}
\label{sec:geometric}

We now address how the ANSIG representation handles shape-preserving
geometric transformations, such as translation, rotation and scale.
For reasons that will become clear below, we require that the shape
vector ${\boldsymbol z} = [ \, z_1 \, z_2 \, \cdots \, z_n\, ]^T$
does not have all its entries with the same value, {\it i.e.}, we
require that ${\boldsymbol z} \in {\mathbb C}_\ast^n$, where
${\mathbb C}_\ast^n = {\mathbb C}^n - \{ [\,z\,z\,\cdots\,z\,]^T:\,z
\in {\mathbb C} \}$ (note that only the pathological cases when all
the landmarks collapse into a single point are excluded).

As introduced in the previous section, a shape is an orbit generated
by a group~$G$ of plausible transformations. Our main goal is to
conceive a computationally efficient mechanism to detect if two
given vectors~${\boldsymbol z}, {\boldsymbol w} \in {\mathbb
C}_\ast^n$ are in the same orbit, {\it i.e.}, if they represent the
same shape. In the previous section, we considered the permutation
group~$G = \Pi(n)$, {\it i.e.}, we considered that ${\boldsymbol z}$
and ${\boldsymbol w}$ represent the same shape if they are equal up
to a permutation. We introduced a maximal invariant with respect to
this group: the ANSIG $a:\,{\mathbb C}_\ast^n \rightarrow {\mathcal
A}$ is such that $a({\boldsymbol z}, \cdot)=a({\boldsymbol w},
\cdot)$ if and only if ${\boldsymbol z}$ and ${\boldsymbol w}$ are
in the same orbit. Building on this result, we now extend the
group~$G$ to also accommodate shape-preserving geometric
transformations.

\subsection{Group of geometric transformations}

Naturally, two vectors ${\boldsymbol z}, {\boldsymbol w} \in
{\mathbb C}_\ast ^n$ represent the same shape if they are equal, up
to, not only a permutation of their entries, but also a translation,
rotation, and scale factor, affecting the set of points they
represent in the plane. To take this into account, we consider the
group of transformations $G = \Pi(n)\times {\mathbb C} \times
{\mathbb S}^1\times {\mathbb R}^+  $, where $\Pi(n)$ is the set of
permutation matrices, ${\mathbb S}^1$ is the unit-circle in the
complex plane, and ${\mathbb R}^+$ is the set of positive real
numbers. We then consider the action of $G$ on ${\mathbb C}_\ast^n$
as the map $G \times {\mathbb C}_\ast^n \rightarrow {\mathbb
C}_\ast^n$, defined~by
\begin{equation}
\left({\boldsymbol \Pi},t,e^{j\theta},\lambda \right) \cdot
{\boldsymbol z} =  \lambda e^{j\theta} {\boldsymbol \Pi}{\boldsymbol
z} + t {\boldsymbol 1}_n\,,\label{eq:groupaction}
\end{equation}
where ${\boldsymbol 1}_n = [\,1\, 1\, \cdots\, 1\,]^T$ is the
$n$-dimensional vector with all entries equal to~$1$. It is clear
from~(\ref{eq:groupaction}) that the action of an element of the
group $G$ on a given vector ${\boldsymbol z}$ corresponds to a
permutation (${\boldsymbol \Pi}$), translation ($t$), rotation
($\theta$), and scaling ($\lambda$), applied to the shape
in~${\boldsymbol z}$. Naturally, the group operation takes into
account the composition of rigid geometric transformations:
\begin{eqnarray}
\left({\boldsymbol \Pi}_1,t_1,e^{j\theta_1},\lambda_1 \right) \cdot
\left({\boldsymbol \Pi}_2,t_2,e^{j\theta_2},\lambda_2 \right) \cdot
{\boldsymbol z} & = &
 \left({\boldsymbol
\Pi}_1,t_1,e^{j\theta_1},\lambda_1 \right) \cdot \left(\lambda_2
e^{j\theta_2} {\boldsymbol \Pi}_2{\boldsymbol
z} + t_2 {\boldsymbol 1}_n\right)\nonumber\\
&=&
 \lambda_1 e^{j\theta_1}{\boldsymbol \Pi}_1 \left(\lambda_2 e^{j\theta_2} {\boldsymbol \Pi}_2{\boldsymbol z}
+ t_2 {\boldsymbol 1}_n\right)+t_1 {\boldsymbol 1}_n\nonumber\\
 &=&
 \lambda_1\lambda_2 e^{j(\theta_1+\theta_2)}{\boldsymbol \Pi}_1{\boldsymbol \Pi}_2{\boldsymbol z}
+ \lambda_1 e^{j\theta_1}t_2 {\boldsymbol 1}_n+t_1 {\boldsymbol 1}_n\nonumber\\
 &=&
 \left(
{\boldsymbol \Pi}_1 {\boldsymbol \Pi}_2, \lambda_1 e^{j\theta_1} t_2
+ t_1, e^{j(\theta_1+\theta_2)},\lambda_1
\lambda_2\right)\cdot{\boldsymbol
z}\,.\label{eq:compgeom}\end{eqnarray}

The shapes that are equal to the instance ${\boldsymbol z}$, up to a
permutation, translation, rotation, and scale, form the orbit~$[
{\boldsymbol z} ]$, the set obtained by considering the action of
all elements of~$G$ on~${\boldsymbol z}$,
\begin{equation}
[ {\boldsymbol z} ] = \left\{ \left({\boldsymbol \Pi}, t,
e^{j\theta},\lambda\right) \cdot {\boldsymbol
z}:\,\left({\boldsymbol \Pi}, t, e^{j\theta},\lambda\right) \in G
\right\}.
\end{equation}
The problem of deciding if two given vectors ${\boldsymbol z}$,
${\boldsymbol w}$ represent the same shape, {\it i.e.}, if they are
in the same orbit, corresponds to checking if the value of the
optimization problem
\begin{equation}\min_{\left({\boldsymbol \Pi}, t,
e^{j\theta},\lambda\right)\in G} \;\left\| {\boldsymbol z} -
\left({\boldsymbol \Pi}, t, e^{j\theta},\lambda\right) \cdot
{\boldsymbol w} \right\| \label{optimproblem} \end{equation} is zero
(or, in practice, below a small threshold, due to the noise). To the
best of our knowledge, solving~(\ref{optimproblem}) without making
use of permutation invariant representations requires an exhaustive
search over the set $\Pi(n)$, which has cardinality~$n!$. Clearly,
this is not feasible, even for moderate values of the number of
landmarks, say $n=100$ ($100!\simeq10^{158}$). In opposition, in our
experiments, we use shapes described by very large sets of points,
{\it e.g.}, with~$n$ up to $40000$.

\subsection{Translation and scale}
\label{subsec:transscale}

We use the ANSIG map to devise a scheme that circumvents the
combinatorial search in~\eqref{optimproblem}. We start by
quotienting out all the transformations, except the rotation,
through a map~$\phi:\,{\mathbb C}_\ast^n \rightarrow {\mathcal A}$.
Then, we show that~$\phi$ is equivariant with respect to the
rotations, {\it i.e.}, that its action commutes with the one of the
group of rotations. This will enable a computationally simple scheme
to decide from $\phi({\boldsymbol z})$ and $\phi({\boldsymbol w})$
if the vectors ${\boldsymbol z}$ and ${\boldsymbol w}$ correspond to
the same shape.

To factor out translation and scale, we pre-process the shape
vector, through centering and normalization, before computing its
signature. This corresponds to considering the composite map
$\phi:\,{\mathbb C}_\ast^n \rightarrow {\mathcal A}$, defined~by
\begin{equation}
\phi({\boldsymbol z} , \cdot ) = a\left( \sqrt{n}
\frac{\mbox{${\boldsymbol z} - \overline{\boldsymbol
z}$}}{\mbox{$\left\| {\boldsymbol z} - \overline{\boldsymbol z}
\right\|$}} , \cdot \right)\,, \label{mapm}
\end{equation}
where $a$ is given by~(\ref{ansig}), $\overline{\boldsymbol z}$ is a
vector with all entries equal to the mean value of~$\boldsymbol z$,
{\it i.e.}, $\overline{\boldsymbol z} = \frac{1}{n} {\boldsymbol
1}_n^T {\boldsymbol z}{\boldsymbol 1}_n$ , and
$\left\|\cdot\right\|$ denotes the 2-norm, {\it i.e.},
$\left\|{\boldsymbol z}\right\|=\sqrt{{\boldsymbol z}^T{\boldsymbol
z}}$. It becomes clear now why we excluded the shape vectors with
all equal entries, {\it i.e.}, why we imposed ${\boldsymbol z} \in
{\mathbb C}_\ast^n$: this guarantees ${\boldsymbol z} \neq
\overline{\boldsymbol z}$, thus $\left\| {\boldsymbol z} -
\overline{\boldsymbol z} \right\|\neq0$ in~(\ref{mapm}). Since in
the remaining of the paper we focus in comparing shapes affected by
the above mentioned geometric transformations, we make $\phi$
supersede $a$, {\it i.e.}, we refer to the analytic function
$\phi({\boldsymbol z} , \cdot )$ as the analytic signature (ANSIG)
of the shape ${\boldsymbol z}$, and to the composite map $\phi$ in
\eqref{mapm} as the ANSIG map.

The reader may wonder why the factor~$\sqrt{n}$ in~(\ref{mapm}). In
fact, this factor has nothing to do with factoring out translation
or scale --- it is constant for shapes described by the same
number~$n$ of landmarks. The motivation for this factor is precisely
to enable dealing with shapes described by different numbers of
landmarks, providing robustness to over/under sampling the shapes.
To demonstrate this capability, consider an extreme example, where
each landmark in the $n$-dimensional vector ${\boldsymbol z} \in
{\mathbb C}_\ast^n$ is repeated $p$ times, leading to the $p
n$-dimensional vector ${\boldsymbol z}_p$, {\it i.e.},
\begin{equation}
{\boldsymbol z}_p = {\boldsymbol 1}_p \otimes {\boldsymbol
z}\label{eq:zp}\,,
\end{equation}
where $\otimes$ is the Kronecker product. Naturally, ${\boldsymbol
z}$ and ${\boldsymbol z}_p$ represent the same shape. We now show
that the factor~$\sqrt{n}$ in~(\ref{mapm}) guarantees that they have
the same ANSIG\footnote{Although this extreme example does not
happen in practice, it is useful to illustrate what may happen when
comparing shapes that, although similar, have been (very)
differently sampled, leading to vectors of (very) different
dimensions. This is frequent, for example, when the shapes to
compare come from the edge maps of images of very different
resolutions.}.

To simplify the notation, name ${\boldsymbol w}$ the post-processed
version of the shape vector ${\boldsymbol z}$, {\it i.e.},
\begin{equation}
{\boldsymbol w} = \sqrt{n} \frac{\mbox{${\boldsymbol z} -
\overline{\boldsymbol z}$}}{\mbox{$\left\| {\boldsymbol z} -
\overline{\boldsymbol z} \right\|$}}\,.\label{w}
\end{equation}
From the relation between ${\boldsymbol z}_p$ and ${\boldsymbol z}$
in~\eqref{eq:zp}, it immediately follows that
\begin{equation}{\boldsymbol z}_p - \overline {\boldsymbol
z}_p = {\boldsymbol 1}_p \otimes \left( {\boldsymbol z} -
\overline{\boldsymbol z} \right) \qquad \mbox{and}\qquad\left\|
{\boldsymbol z}_p - \overline {\boldsymbol z}_p \right\| = \sqrt{p}
\left\| {\boldsymbol z} - \overline {\boldsymbol z} \right\|\,,
\end{equation}
thus the post-processed version of ${\boldsymbol z}_p$ is just the
$p$-times replication of the one of ${\boldsymbol z}$ in \eqref{w},
\begin{equation}
\sqrt{p n} \frac{\mbox{${\boldsymbol z}_p - \overline{\boldsymbol
z}_p$}}{\mbox{$\left\| {\boldsymbol z}_p - \overline{\boldsymbol
z}_p \right\|$}}=\sqrt{n} \frac{\mbox{${\boldsymbol 1}_p \otimes
({\boldsymbol z} - \overline{\boldsymbol z})$}}{\mbox{$\left\|
{\boldsymbol z} - \overline{\boldsymbol z} \right\|$}}={\boldsymbol
1}_p \otimes {\boldsymbol w}\,.
\end{equation}

The ANSIG of ${\boldsymbol z}_p$ is now successively written as
\begin{eqnarray}
\phi({\boldsymbol z}_p, \xi) & = & a\left( {\boldsymbol 1}_p \otimes
{\boldsymbol w} , \xi \right)\label{inicio}
 \\
& = & \frac{1}{p n} \left( \rule{0ex}{3ex} \right.
\underbrace{\sum_{m=1}^{n} e^{w_m \xi}  + \sum_{m=1}^{n} e^{w_m \xi}
+ \cdots + \sum_{m=1}^{n} e^{w_m \xi}}_{\mbox{$p$ times}} \left.
\rule{0ex}{3ex} \right) \label{esta} \\
 & = & \frac{1}{n}
\sum_{m=1}^{n} e^{w_m \xi}  \\ & = & a({\boldsymbol w},
\xi)\label{esta2}
 \\ & = & \phi({\boldsymbol z}, \xi)\,,\label{fim}
\end{eqnarray}
establishing the desired equality between the signatures of
${\boldsymbol z}_p$ and ${\boldsymbol z}$. Equalities \eqref{inicio}
and \eqref{fim} come from the definition of~$\phi$ in (\ref{mapm});
in (\ref{esta}) and~(\ref{esta2}) we used the definition of $a$ in
(\ref{ansig}).

\noindent\textbf{Maximal invariance with respect to permutation,
translation, and scale.} We now show that $\phi$ is a maximal
invariant with respect to all transformations, except rotation, {\it
i.e.}, that two shapes have the same ANSIG if and only if they are
equal up to permutation, translation, and scale.

The maximal invariance of $\phi$ can be formalized~as
\begin{equation} \phi({\boldsymbol z} , \cdot) = \phi({\boldsymbol
w} , \cdot)\; \Leftrightarrow\; {\boldsymbol z} = \left({\boldsymbol
\Pi},t,1,\lambda\right) \cdot {\boldsymbol w}\,, \label{fundprop}
\end{equation} for some $({\boldsymbol \Pi},t,1,\lambda) \in G$. The
invariance, {\it i.e.}, the sufficiency part ($\Leftarrow$) of
(\ref{fundprop}), is straightforward:
\begin{eqnarray}
\phi({\boldsymbol z} , \cdot) &=& \phi\left(\left({\boldsymbol
\Pi},t,1,\lambda\right) \cdot {\boldsymbol w} , \cdot\right) \\
 &=& \phi\left(\lambda{\boldsymbol \Pi}{\boldsymbol w}+t{\boldsymbol 1}_n ,
 \cdot\right)\\
 &=& a\left(\sqrt{n}\frac{\lambda{\boldsymbol \Pi}{\boldsymbol w}+t{\boldsymbol 1}_n-\overline{\lambda{\boldsymbol \Pi}{\boldsymbol w}+t{\boldsymbol 1}_n}}
 {\left\|\lambda{\boldsymbol \Pi}{\boldsymbol w}+t{\boldsymbol 1}_n-\overline{\lambda{\boldsymbol \Pi}{\boldsymbol w}+t{\boldsymbol 1}_n}\right\|} ,
 \cdot\right)\\
 &=& a\left(\sqrt{n}\frac{{\boldsymbol \Pi}\left({\boldsymbol w}-\overline{{\boldsymbol
 w}}\right)}
 {\left\|{\boldsymbol w}-\overline{{\boldsymbol
 w}}\right\|} ,
 \cdot\right)\label{eq:vec}\\
 &=& a\left(\sqrt{n}\frac{{\boldsymbol w}-\overline{{\boldsymbol
 w}}}
 {\left\|{\boldsymbol w}-\overline{{\boldsymbol
 w}}\right\|} ,
 \cdot\right)\label{eq:ainv}\\
 &=& \phi({\boldsymbol w} , \cdot)\,,
\end{eqnarray}
where we just used the definition of $\phi$, standard manipulation
of matrices and vectors in \eqref{eq:vec}, and the invariance of $a$
with respect to permutations in \eqref{eq:ainv}.

We now prove the maximal invariance, {\it i.e.}, the necessity
 part ($\Rightarrow$) of the equivalence~(\ref{fundprop}). Naturally,
 the maximal invariance of $\phi$ also hinges on the
maximal invariance of $a$ established in the previous section.
Suppose that $\phi({\boldsymbol z}, \cdot) = \phi({\boldsymbol w} ,
\cdot)$. Then, from the definition of~$\phi$ in~(\ref{mapm}),
\begin{equation}
a\left( \sqrt{n} \frac{\mbox{${\boldsymbol z} -
\overline{\boldsymbol z}$}} {\mbox{$\left\| {\boldsymbol z} -
\overline{\boldsymbol z} \right\|$}} , \cdot \right) = a\left(
\sqrt{n} \frac{\mbox{${\boldsymbol w} - \overline{\boldsymbol
w}$}}{\mbox{$\left\| {\boldsymbol w} - \overline{\boldsymbol w}
\right\|$}} , \cdot \right)\,, \label{interm1}
\end{equation}
and, by the maximal invariance of $a$ with respect to permutations,
equality~(\ref{interm1}) implies that
\begin{equation}
\sqrt{n}  \frac{\mbox{${\boldsymbol z} -
\overline{\boldsymbol
z}$}}{\mbox{$\left\| {\boldsymbol z} -
 \overline{\boldsymbol z} \right\|$}} =
 {\boldsymbol \Pi}  \sqrt{n} \frac{\mbox{${\boldsymbol w} -
 \overline{\boldsymbol w}$}}{\mbox{$\left\| {\boldsymbol w} -
 \overline{\boldsymbol w} \right\|$}}
 \,,\label{interm2}
\end{equation}
for some ${\boldsymbol \Pi} \in \Pi(n)$. The relation between
${\boldsymbol z}$ and ${\boldsymbol w}$ in~(\ref{interm2}) can be
rewritten as
\begin{equation}
{\boldsymbol z}=\frac{\left\| {\boldsymbol z}- \overline{\boldsymbol
z} \right\|}{\left\| {\boldsymbol w} - \overline{\boldsymbol w}
\right\|}  {\boldsymbol \Pi}{\boldsymbol w} - \frac{\left\|
{\boldsymbol z}- \overline{\boldsymbol z} \right\|}{\left\|
{\boldsymbol w} - \overline{\boldsymbol w}
\right\|}\overline{\boldsymbol w}+\overline{\boldsymbol
z}\,.\label{aux}
\end{equation}
It is now clear that this relation is of the form ${\boldsymbol z} =
\lambda{\boldsymbol \Pi}  {\boldsymbol w} + t {\boldsymbol 1}_n$,
establishing that ${\boldsymbol z} = ({\boldsymbol \Pi},t,1,\lambda)
\cdot {\boldsymbol w}$ for a particular choice of $({\boldsymbol
\Pi},t,1,\lambda) \in G$, thus completing the proof of
\eqref{fundprop}; just define
\begin{equation}
\lambda = \frac{\mbox{$\left\| {\boldsymbol z} -
\overline{\boldsymbol z} \right\|$}}{\mbox{$\left\| {\boldsymbol w}
- \overline{\boldsymbol w} \right\|$}} > 0 \qquad \mbox{ and }\quad
t = \frac{1}{n}{\boldsymbol 1}_n^T\left({\boldsymbol
z}-\lambda{\boldsymbol w}\right)\, \in{\mathbb C}\,.\label{defs}
\end{equation}

\subsection{Rotation}
\label{subsec:rotation}

We have seen that the ANSIG map~$\phi$ quotients out all
transformations in~$G$, except the rotation. We now show that
although $\phi$ is not invariant to rotations, it is equivariant,
{\it i.e.}, that its action commutes with the one of the group of
rotations. Naturally, this equivariance is inherited from  the map
$a$ in~(\ref{ansig}). Let us then start by looking at how the
rotation of a shape ${\boldsymbol z}$ affects the analytic function
produced by the map $a$. The rotated version of the shape is simply
$e^{j\theta} {\boldsymbol z}$. From the definition of $a$
in~(\ref{ansig}), it follows that
\begin{equation}
a(e^{j\theta} {\boldsymbol z} , \xi)  =  \frac{1}{n} \sum_{k=1}^n
e^{e^{j \theta} z_k \xi} = \frac{1}{n} \sum_{k=1}^n e^{ z_k \left(
e^{j \theta} \xi \right)} = a({\boldsymbol z} , e^{j\theta}
\xi)\,.\label{propa}
\end{equation}

\noindent{\bf Equivariance with respect to rotation.} The property
just derived \eqref{propa} leads immediately to the equivariance of
the ANSIG map~$\phi$, through the following sequence of equalities:
\begin{eqnarray}
\phi\left( ({\boldsymbol \Pi},t,e^{j\theta},\lambda) \cdot
{\boldsymbol z} , \xi \right) & = & \phi\left( \lambda
e^{j\theta}{\boldsymbol \Pi}
 {\boldsymbol z} + t {\boldsymbol 1}_n , \xi
\right) \label{minvbegin} \\ & = & a\left( \sqrt{n}
\frac{e^{j\theta}{\boldsymbol \Pi}  \left({\boldsymbol z} -
\overline{\boldsymbol z}\right)}{\left\| {\boldsymbol z} -
\overline{\boldsymbol z} \right\|} ,
\xi \right) \label{eq:mat} \\
& = & a\left( \sqrt{n} \frac{\mbox{$ e^{j\theta} ({\boldsymbol z} -
\overline{\boldsymbol z})$}}{\mbox{$\left\| {\boldsymbol z} -
\overline{\boldsymbol z} \right\|$}} , \xi \right)
\label{minv1} \\
& = & a\left( \sqrt{n} \frac{\mbox{${\boldsymbol z} -
\overline{\boldsymbol z}$}}{\mbox{$\left\| {\boldsymbol z} -
\overline{\boldsymbol z} \right\|$}} ,e^{j\theta} \xi  \right) \label{minv2} \\
& = & \phi({\boldsymbol z} , e^{j\theta}\xi )\,, \label{minvend}
\end{eqnarray}
where \eqref{eq:mat} and \eqref{minvend} use the definition of
$\phi$ in \eqref{mapm} and simple manipulations, \eqref{minv1} uses
the invariance of $a$ with respect to permutations, and
\eqref{minv2} comes from \eqref{propa}.

We thus conclude that the rotation of a shape propagates to its
signature, {\it i.e.}, the ANSIG of the rotated shape is simply a
rotated version (in the complex plane) of the ANSIG of the original
shape. Obviously, the equivariance is inherited by the restriction
of the ANSIG to the unit-circle, $\phi_{{\mathbb S}^1}:\,{\mathbb
C}_\ast^n \times {\mathbb S}^1 \rightarrow {\mathbb C}$,
$\phi_{{\mathbb S}^1}({\boldsymbol z} , \xi) = \phi({\boldsymbol z},
\xi)$:
\begin{equation} \phi_{{\mathbb S}^1}\left( \left( {\boldsymbol
\Pi},t, e^{j\theta},\lambda \right) \cdot {\boldsymbol z} ,
\xi\right) = \phi_{{\mathbb S}^1}\left({\boldsymbol z} , e^{j\theta}
\xi\right)\,. \label{equivariant}
\end{equation}

%

\noindent\textbf{Equality of orbits.} In summary, two vectors are in
the same orbit of the group G of relevant transformations, {\it
i.e.}, they represent the same shape, up to permutation,
translation, rotation, and scale, iif (the restriction to the
unit-circle of) their ANSIGs are equal, up to a rotation:
\begin{equation}
\mbox{${\boldsymbol z}$ and ${\boldsymbol w}$ represent the same
shape}\;\Leftrightarrow\; \phi_{{\mathbb S}^1}({\boldsymbol z},
\cdot) = \phi_{{\mathbb S}^1}\left({\boldsymbol w}, e^{j\theta} \,
 \cdot \right),\,\mbox{for some }\theta \in [0,2\pi].
\label{detecting} \end{equation}

The necessity part of this claim ($\Rightarrow$) comes immediately
from \eqref{equivariant}. To prove the sufficiency ($\Leftarrow$),
suppose that $\phi_{{\mathbb S}^1}({\boldsymbol z}, \cdot) =
\phi_{{\mathbb S}^1}\left({\boldsymbol w},
e^{j\theta}\,\cdot\right)$, for some $\theta \in [0,2\pi]$. This
means that
\begin{equation}
\phi({\boldsymbol z}, e^{j \varphi}) = \phi({\boldsymbol w} ,
e^{j\theta}e^{j \varphi} )\,, \quad \mbox{ for all } \varphi \in
[0,2\pi] \label{exp1}.
\end{equation}
Using the equivariance of $\phi$, derived
in~(\ref{minvbegin}-\ref{minvend}), in the right-hand side
of~(\ref{exp1}), we get
\begin{equation}
\phi({\boldsymbol z},  e^{j \varphi}) = \phi(
e^{j\theta}{\boldsymbol w} ,  e^{j \varphi} )\,, \quad \mbox{ for
all } \varphi \in [0,2\pi] \label{exp2}\,,
\end{equation}
stating that the analytic functions $\phi({\boldsymbol z},\cdot)$
and $\phi(e^{j\theta}{\boldsymbol w} ,\cdot)$ coincide on the
unit-circle, thus, on the entire complex plane:
\begin{equation}
\phi({\boldsymbol z},\cdot) = \phi(e^{j\theta}{\boldsymbol w}
,\cdot)\,. \label{tmp}\end{equation} From the maximal invariance of
the map~$\phi$, stated in~(\ref{fundprop}), equality~(\ref{tmp}) is
equivalent to
\begin{equation} {\boldsymbol z} = ({\boldsymbol
\Pi},t,1,\lambda) \cdot e^{j\theta} {\boldsymbol w}\,, \label{final}
\end{equation} for some $({\boldsymbol \Pi},t,1,\lambda)\in G$.
Using the composition of elements of~G, expressed in
\eqref{eq:compgeom}, equality~(\ref{final}) can be rewritten as
${\boldsymbol z} = ({\boldsymbol \Pi},t,e^{j\theta},\lambda) \cdot
{\boldsymbol w}$,
which shows that ${\boldsymbol z}$ and ${\boldsymbol w}$ are in the
same orbit, {\it i.e.}, that they represent the same shape.

\subsection{Illustrations}

We now illustrate the behavior of the ANSIG representation in what
respects to the properties studied in this section, {\it i.e.},
dealing with shape-preserving geometric transformations and with
shapes described by point sets of different cardinality ({\it i.e.},
with different sampling density).

\begin{figure}[hbt]
\centerline{\epsfig{figure=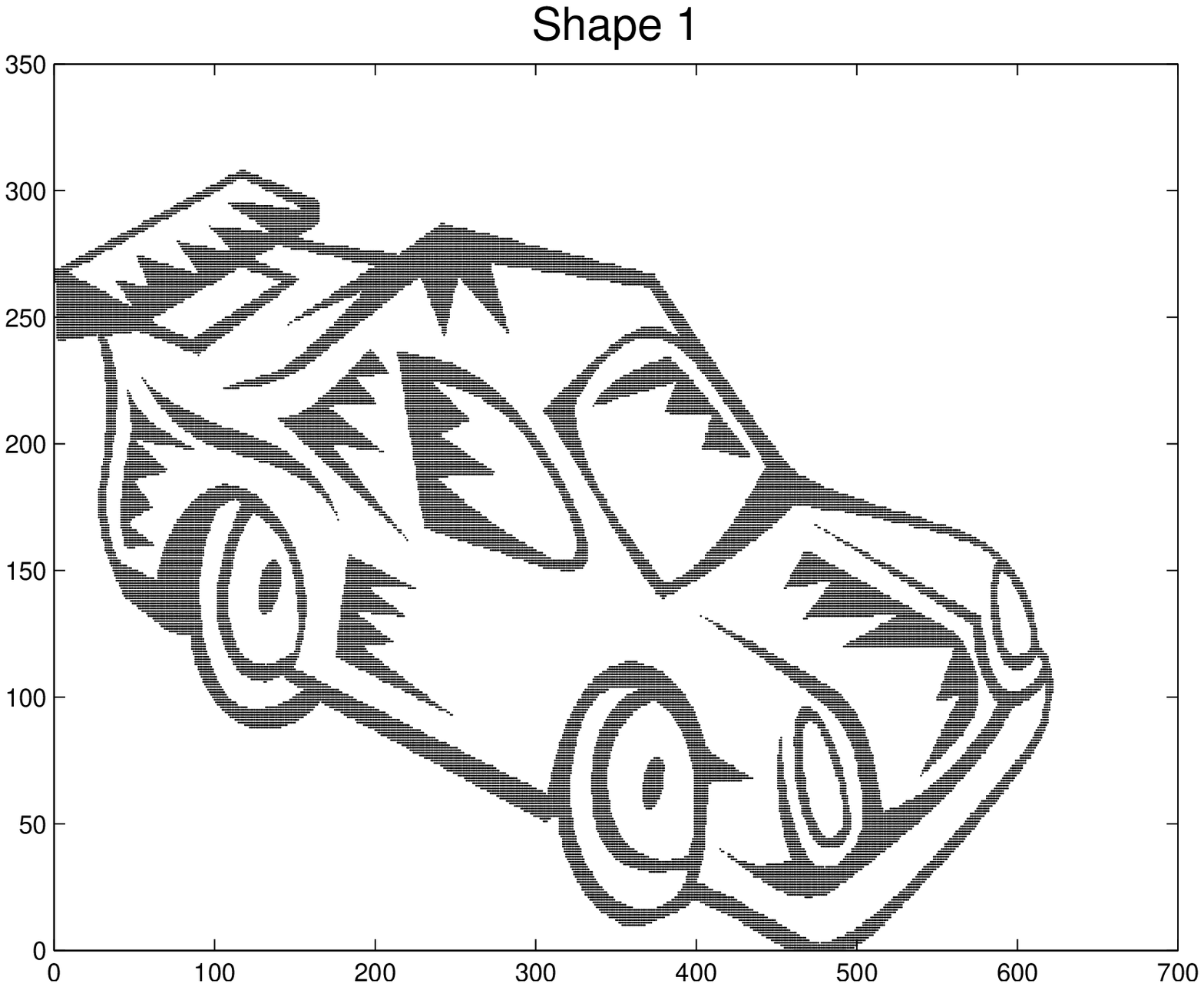,
width=5cm}\hspace*{1cm} \epsfig{figure=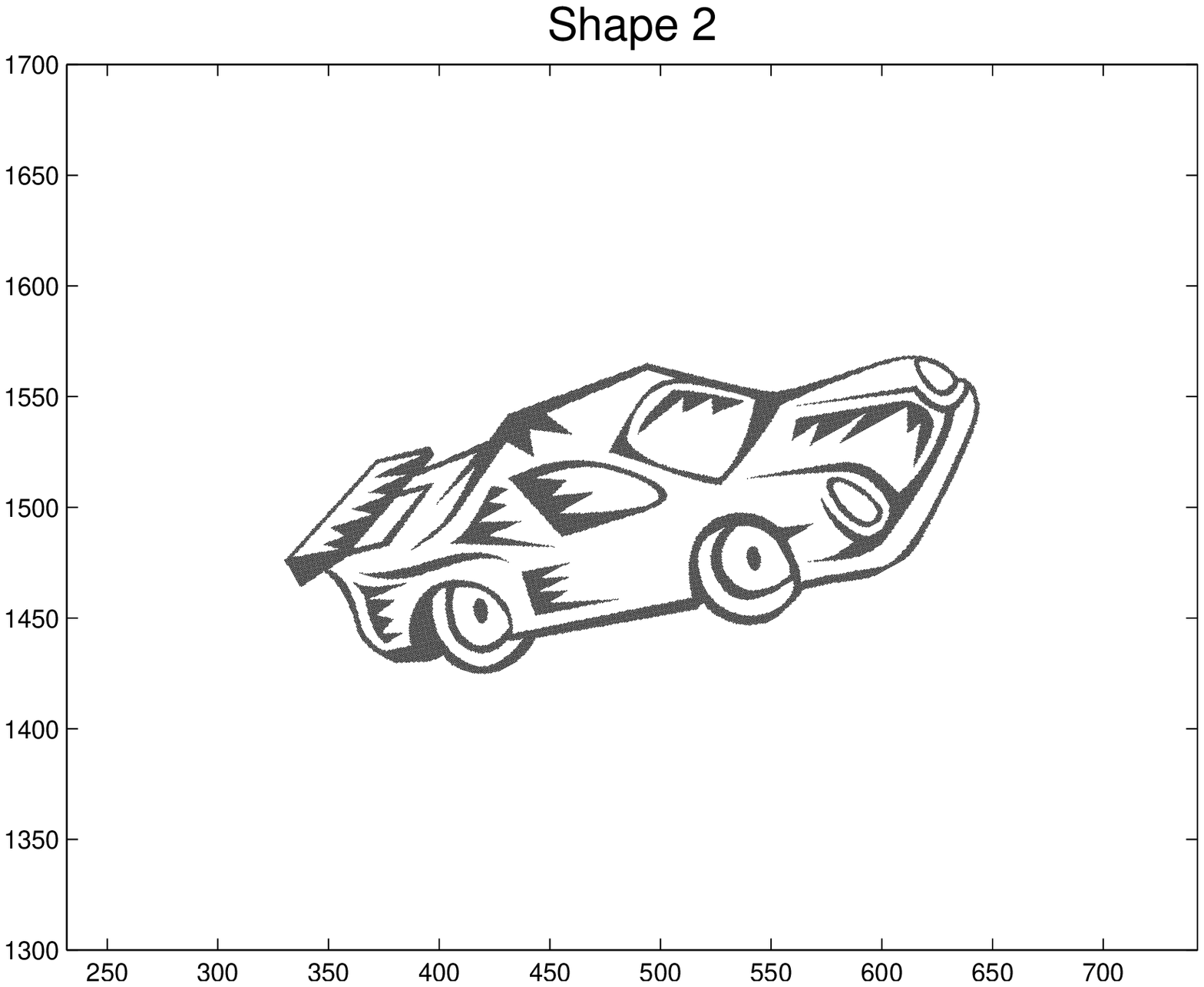,
width=5cm}} \vspace*{-.5cm}\caption{Two shapes that differ by a
geometric transformation: the right one is obtained by translating,
rotating, and resizing the left one. The ordering of the points
describing the shapes is also made distinct on purpose, as in the
example of Fig.~\ref{fig:perm_star}.\label{fig:illustration1}}
\end{figure}

\noindent{\bf Shape-preserving transformations.}  We use the binary
images shown in Fig.~\ref{fig:illustration1} which, besides a point
permutation (not seen in the images), also differ by translation,
rotation, and scale factors (easily perceived due to different
position, orientation, and scale of the shapes).

\begin{figure}[hbt]
\centerline{\epsfig{figure=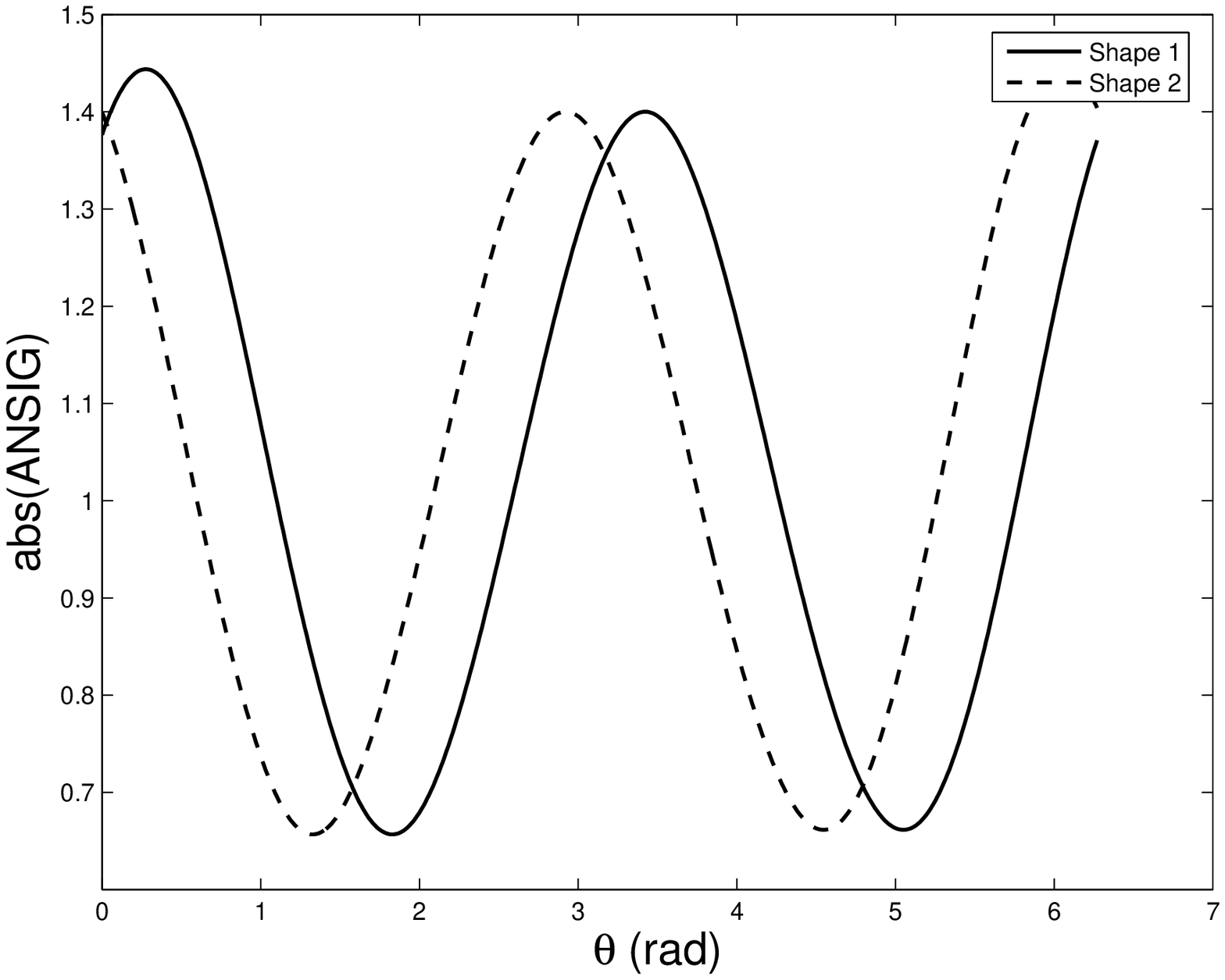,
width=5cm}\hspace*{1cm}
\epsfig{figure=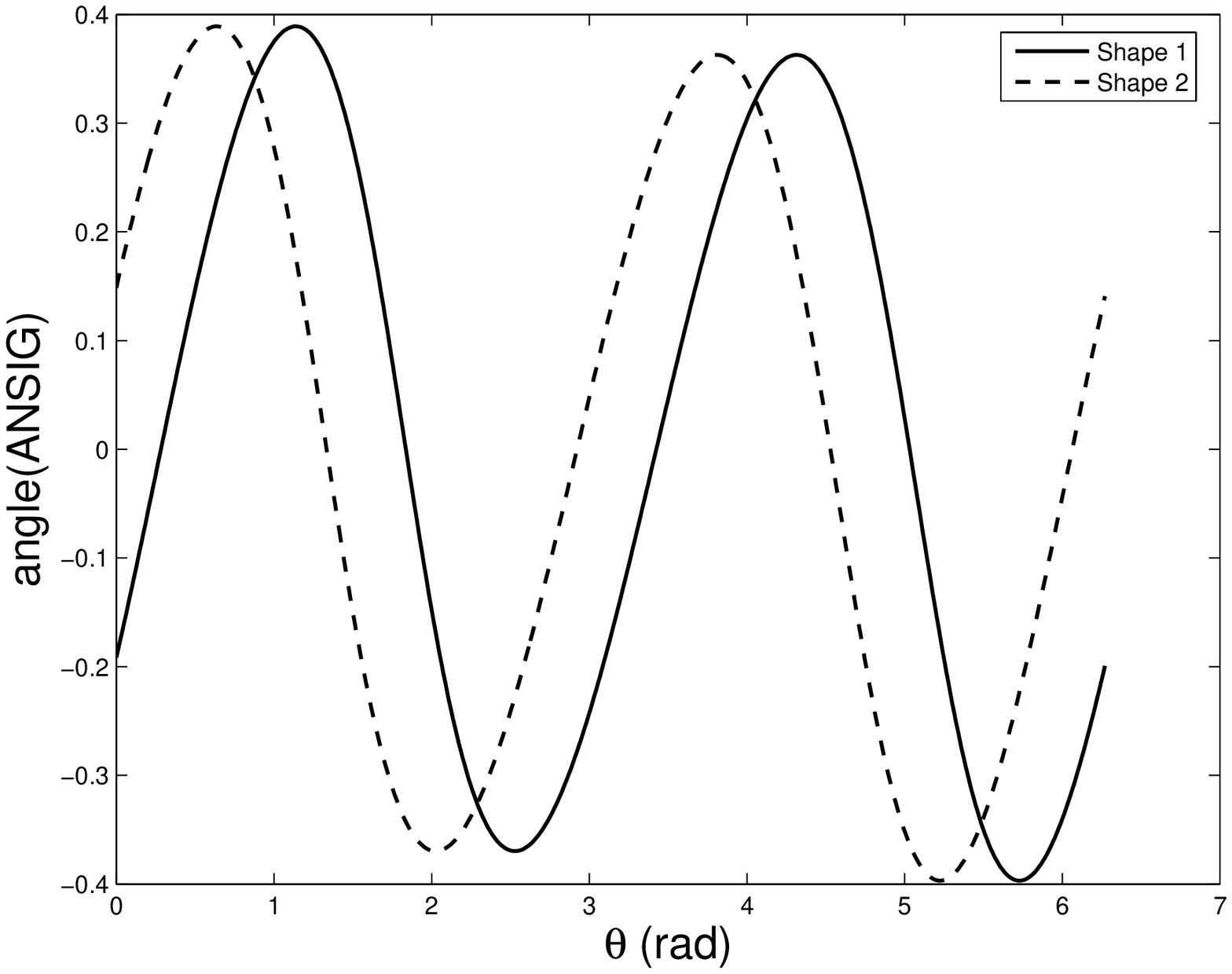,width=5cm}}
\vspace*{-.5cm}\caption{Magnitude and phase of the ANSIGs of the two
shapes of Fig.~\ref{fig:illustration1}. Note that, although these
shapes differ by position, rotation, scale, and point labeling,
their ANSIGs only differ by a (circular) translation that can be
easily computed.\label{fig:illustration3}}
\end{figure}

We computed the (unit-circle samples of the) ANSIGs of the shapes in
Fig.~\ref{fig:illustration1}, obtaining the magnitude and phase
plots in Fig.~\ref{fig:illustration3}. Note that, in spite of the
different vectors describing the shapes in
Fig.~\ref{fig:illustration1}, both ANSIG plots in
Fig.~\ref{fig:illustration3} only differ by a (circular)
translation. This is in agreement with what we concluded in this
section, see expression~(\ref{equivariant}): permutation,
translation, and scale, are factored out; the rotation of a shape
induces the same rotation on its ANSIG, thus a (circular)
translation of the magnitude and phase plots of its restriction to
the unit-circle. The rotation that aligns the shapes can be
efficiently computed, as will be described in
Section~\ref{sec:implementation}.

\noindent{\bf Different sampling density.} The experiment above used
shapes described by sets of points of the same cardinality. We now
illustrate that the ANSIG also copes with shapes described by point
sets of (very) different cardinality, as also discussed in this
section. This characteristic is very important in practice, {\it
e.g.}, to deal with shapes obtained from images of different
resolutions (due do the pixelization, the same shape usually leads
to point sets of different cardinality).

\begin{figure}[hbt]
\centerline{\epsfig{figure=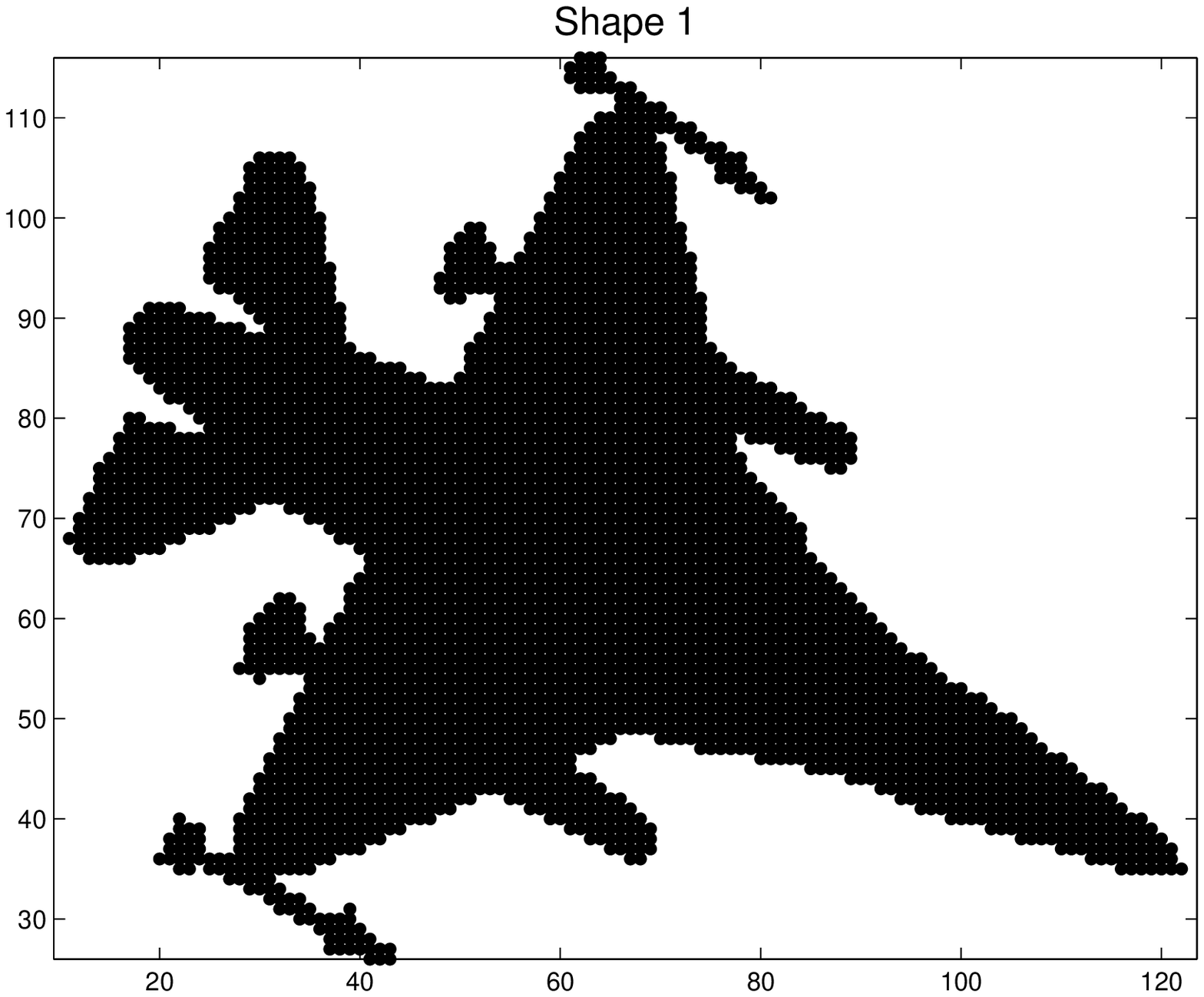,
width=5cm}
\hspace*{1cm}\epsfig{figure=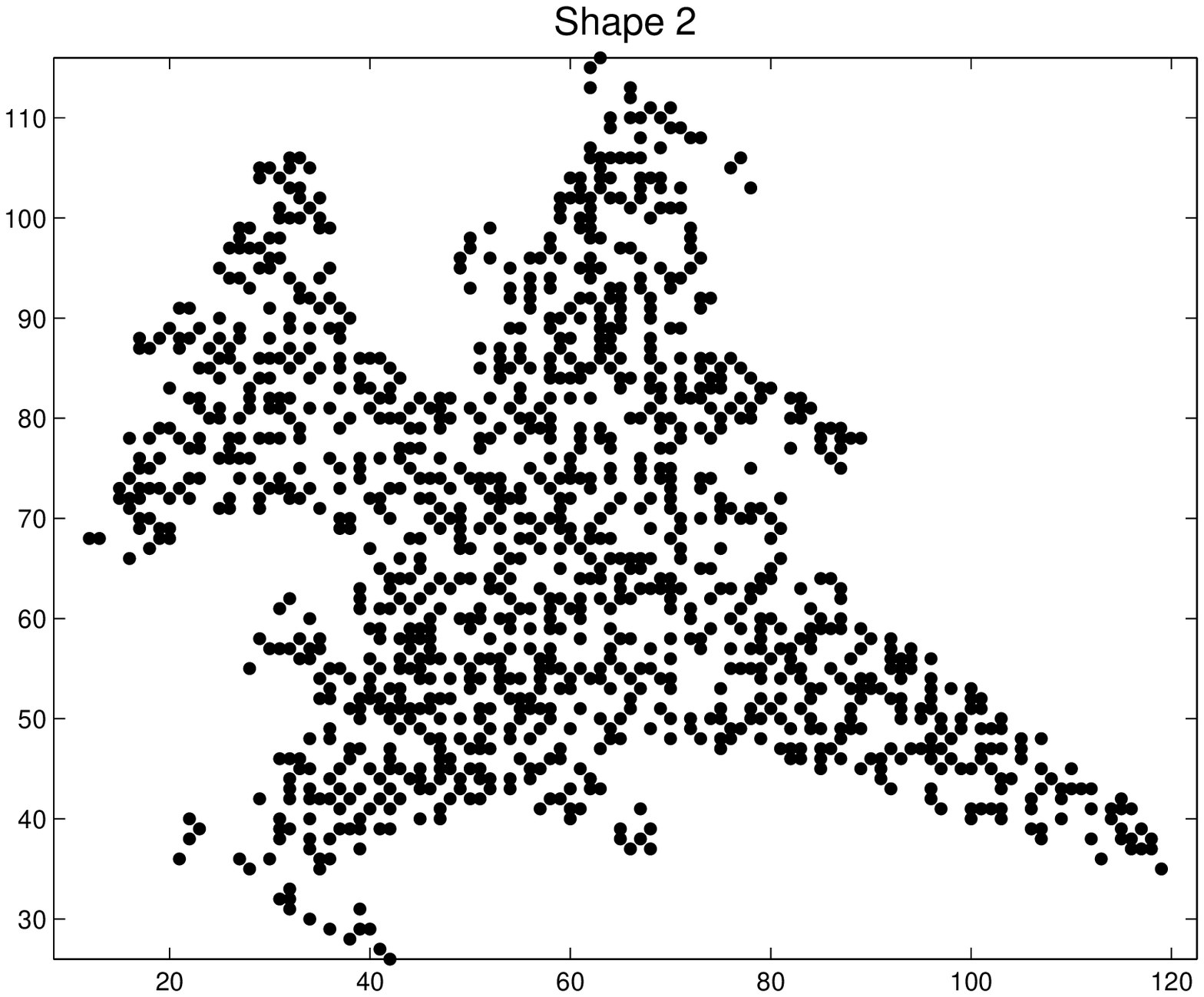,
width=5cm}}
    \vspace*{-.5cm}\caption{Similar shapes with different sampling density. The right image has 30\% of the points of the left one.\label{fig:sampling}}
    \end{figure}

\begin{figure}[hbt]
\centerline{\epsfig{figure=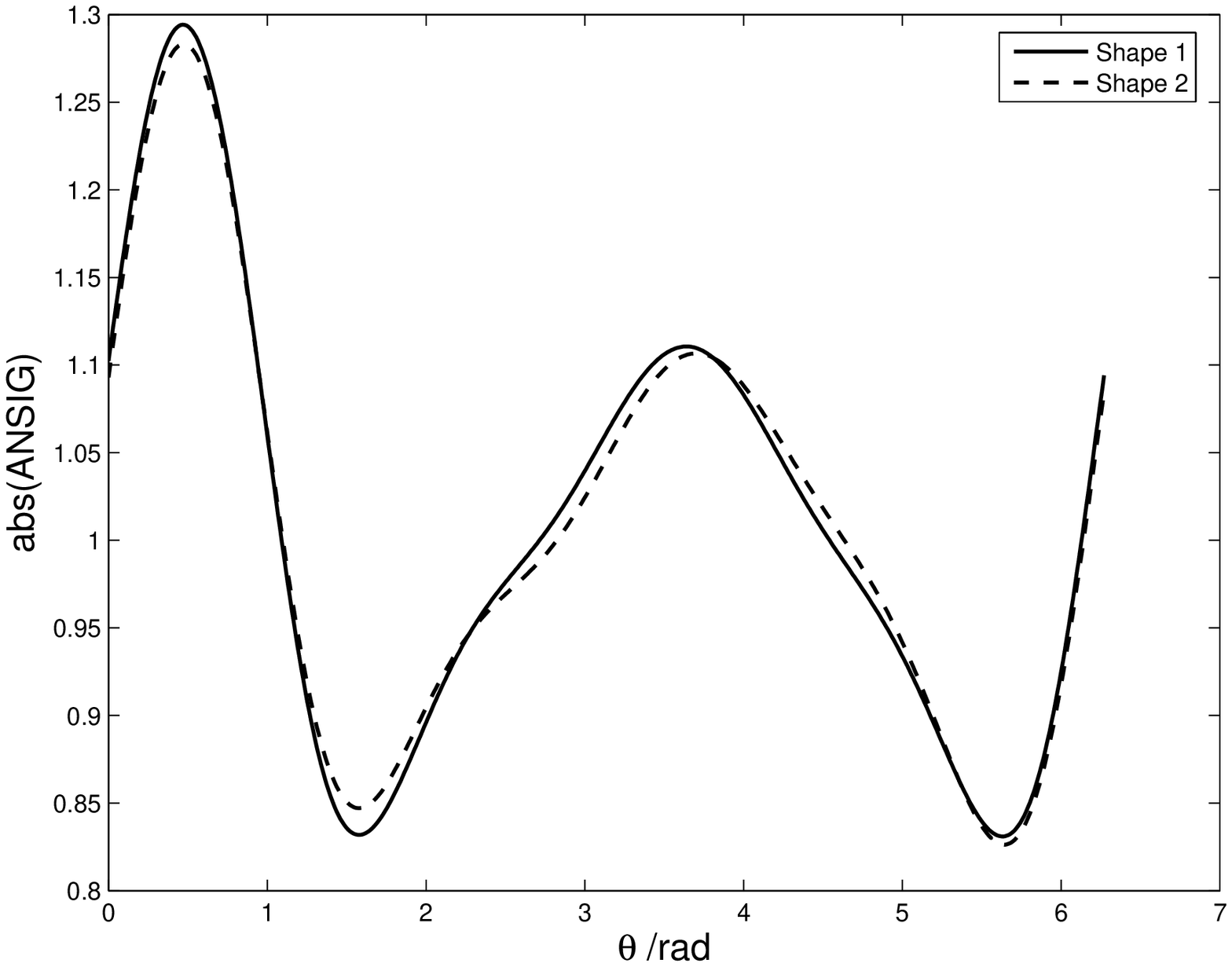,
width=5cm}\hspace*{1cm}\epsfig{figure=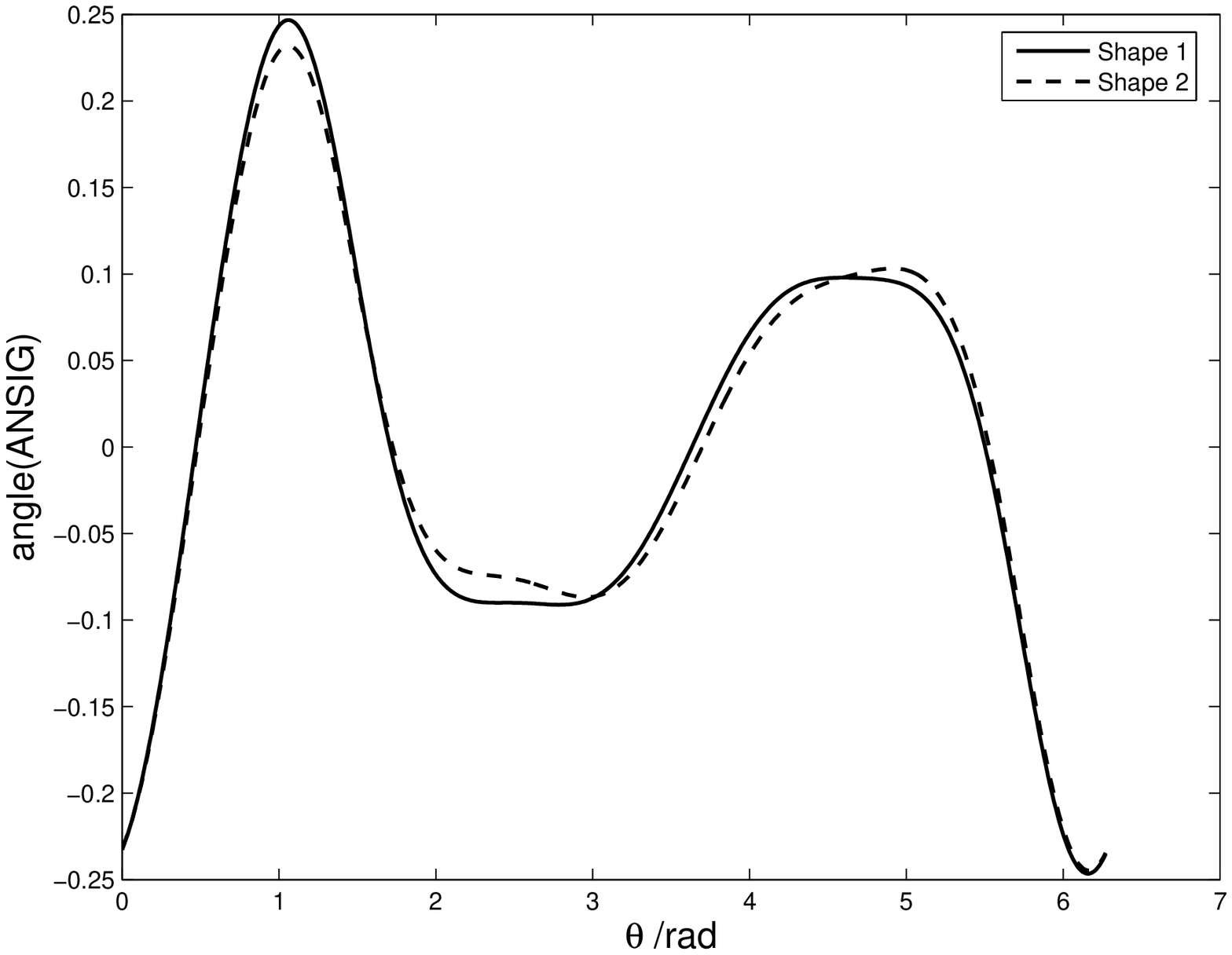,
width=5cm}}
    \vspace*{-.5cm}\caption{Magnitude and phase of the ANSIGs of the shapes in
Fig.~\ref{fig:sampling}. Although the number of points describing
these shapes differ significantly, their ANSIGs result similar, as
desired.\label{fig:sampling2}}
    \end{figure}

We use the shapes represented in Fig.~\ref{fig:sampling}. Both
represent the same object but the shape in the right is described by
just $30\%$ of the points of the one in the left. In
Fig.~\ref{fig:sampling2}, we represent the ANSIGs of both shapes. As
easily seen, the ANSIGs of the differently sampled shapes almost
coincide, as desired, and anticipated in this section, see the
insight provided by expressions~(\ref{inicio}-\ref{fim}). All our
tests confirmed that the ANSIG representation deals well with
different sampling densities, at leat as far as the corresponding
shapes are recognizable by an human.

We finally comment on the fact that the ANSIG plots in
Fig.~\ref{fig:perm_star_ANSIG_plane} of the previous section are
periodic (they exhibit not only the $2\pi$ periodicity of any ANSIG
but also a smaller fundamental period of $\pi$). This periodicity is
due the fact that the shape corresponding to those plots is
invariant under rotations of multiples of $\pi$ (see
Fig.~\ref{fig:perm_star}). Since shape rotation leads to a
translation of the ANSIG plots, the plots for the shape in
Fig.~\ref{fig:perm_star} result invariant to translations of
multiples of $\pi$, {\it i.e.}, they are periodic with period $\pi$,
therefore two periods appear in the interval $[0,2\pi]$. Naturally,
this does not happen with shapes that do not exhibit rotational
symmetry, {\it e.g.}, the ones in Figs.~\ref{fig:illustration1}
and~\ref{fig:sampling}, whose ANSIG plots in
Figs.~\ref{fig:illustration3} and ~\ref{fig:sampling2} exhibit only
$2\pi$ periodicity.

\section{Shape-based recognition}
\label{sec:implementation}

In this section we describe how the ANSIG representation can be used
for rigid shape recognition. Since our representation is invariant
to (or deals gracefully with) the relevant transformations (point
labeling, translation, rotation, and scale), we are able to perform
shape classification using a very simple scheme: the shape database
is just composed by the ANSIGs of prototype shapes (a single
prototype per shape class); classification boils down to comparing
the ANSIG of a candidate shape with the ones in the database.

Our experiments show that the invariance of the ANSIG representation
enables dealing with noise and distortions typical of shapes
obtained from real images, with the simple strategy just outlined,
avoiding this way computationally complex learning schemes. We
emphasize, however, that the ANSIG representation can be
incorporated in more sophisticated classification procedures if
non-rigid shape classification is the goal. For example, storing the
ANSIGs of several prototypes per class and using k-NN (nearest
neighbors) classification is straightforward.

\subsection{Efficient comparison of ANSIGs}
\label{subsec:comparison}

As described in Section~\ref{sec:perminv}, a practical way to
implement the map $a:\,{\mathbb C}_\ast^n \rightarrow {\mathcal A}$
is through its unit-circle sampled version $a_K:\,{\mathbb C}^n_\ast
\rightarrow {\mathbb C}^K$, introduced in~(\ref{discrete}). Thus, in
practice, the unit-circle ANSIG restriction $\phi_{{\mathbb S}^1}$
is approximated by its sampled version $\phi_K:\,{\mathbb C}_\ast^n
\rightarrow {\mathbb C}^K$, given~by
\begin{equation}
\phi_K({\boldsymbol z} ) = a_K\left( \sqrt{n}
\frac{\mbox{${\boldsymbol z} - \overline{\boldsymbol
z}$}}{\mbox{$\left\| {\boldsymbol z} - \overline{\boldsymbol z}
\right\|$}}  \right)\,, \end{equation} where, we recall, $K$ denotes
the number of samples on the unit-circle. Naturally, with $K$
sufficiently large, any rotation $e^{j \theta}$ is well approximated
by a point in the unit-circle sampling grid, {\it i.e.}, $e^{j
\theta} \simeq e^{j\frac{2\pi}{K}k}=W_K^k$, thus the equivariance
of~$\phi_{{\mathbb S}^1}$, expressed in~(\ref{equivariant}),
gracefully transfers to its discrete version~$\phi_K$, as
\begin{equation}
\phi_K\left( \left( {\boldsymbol \Pi}, \lambda, v, e^{i\theta}
\right) \cdot {\boldsymbol z} \right)  \simeq\phi_K ( {\boldsymbol
z})\,{\sf mod}\,k\,, \label{equivariant2}
\end{equation}
where ${\sf mod}\,k$ denotes a $k$-step cyclic shift.

Our test for deciding if two point vectors ${\boldsymbol z}$ and
${\boldsymbol w}$ correspond to the same shape, {\it i.e.}, for the
equality of the orbits of~${\boldsymbol z}$ and ${\boldsymbol w}$,
expressed in~(\ref{detecting}), leads to checking if
$\phi_K({\boldsymbol w})$ is a cyclic-shifted version of
$\phi_K({\boldsymbol z})$, {\it i.e.}, if
\begin{equation}
\phi_K({\boldsymbol z}) \simeq \phi_K({\boldsymbol w})\,{\sf
mod}\,k\,, \qquad \mbox{ for some }\; k = 0, 1, \ldots, K-1.
\label{equivariant3}
\end{equation}
This test can be carried out by computing
the cyclic-shift~$k^*$ that best ``aligns" the vectors,
\begin{equation} k^* =
\arg\min_{k = 0, 1, \ldots, K-1} \left\| \phi_K({\boldsymbol z}) -
\phi_K ({\boldsymbol w})\,{\sf mod}\,k \right\|^2 \,, \label{align}
\end{equation}
and then checking the similarity of the corresponding ``aligned"
versions.

Solving~(\ref{align}) by computing $\phi_K({\boldsymbol w}) \, {\sf
mod}\,k$ and comparing it with $\phi_K({\boldsymbol z})$, for each
$k$, leads to an algorithm with computational complexity ${\mathcal
O}(K^2)$. We now present a computationally simpler scheme, based on
the Fast Fourier Transform (FFT). We denote the Discrete Fourier
Transform (DFT) of a vector ${\boldsymbol v} \in {\mathbb C}^K$ by
the vector~$\widehat{\boldsymbol v} \in {\mathbb C}^K $, given by
\begin{equation}
\widehat{\boldsymbol v} = {\boldsymbol D}_K^H {\boldsymbol
v}\,,\qquad\mbox{with}\qquad{\boldsymbol d}_k = \frac{1}{\sqrt{K}}
\begin{bmatrix} 1 & W_K^k & W_K^{2 k} &  W_K^{3 k} & \cdots &
W_K^{(K-1) k}
\end{bmatrix}^T\,.\label{fft}
\end{equation}
In \eqref{fft}, ${\boldsymbol D}_K$ is the $K\times K$ DFT
matrix~\cite{opp}, ${\boldsymbol D}_K^H$ denotes its conjugate
transpose (also known as the Hermitian), and ${\boldsymbol d}_k$ is
the
$k^{\mbox{\small th}}$ column of ${\boldsymbol D}_K$. 

Using Parseval's relation~\cite{opp}, which is based on the fact
that the DFT is an unitary operator (${\boldsymbol D}_K^H
{\boldsymbol D}_K = {\boldsymbol I}_K)$,  the
minimization~(\ref{align}) is written in the frequency domain~as
\begin{equation}
k^* = \arg\min_{k = 0, 1, \ldots, K-1}
\left\|\widehat{\phi_K({\boldsymbol z})} - \widehat{\phi_K
({\boldsymbol w})\,{\sf mod}\,k} \right\|^2\,.\label{dft1}
\end{equation}
Since the DFT of a $k$-cyclic-shifted version of a signal equals the
DFT of the original signal multiplied by the exponential sequence
$\sqrt{K} {\boldsymbol d}_k$ \cite{opp}, expression (\ref{dft1}) is
equivalent~to
\begin{equation}
k^* =\arg\min_{k = 0, 1, \ldots, K-1} \left\|
\widehat{\phi_K({\boldsymbol z})} - \widehat{\phi_K ({\boldsymbol
w})} \odot \sqrt{K}\, {\boldsymbol d}_k \right\|^2\,, \label{dft2}
\end{equation}
where $\odot$ denotes the elementwise (also known as Schur or
Hadammard) product. Removing from the norm in~(\ref{dft2}) the terms
that do not depend on~$k$, we finally get
\begin{eqnarray}
k^* =\arg\max_{k = 0, 1, \ldots, K-1} {\sf Re} \left\{{\boldsymbol
d}_k^H \left[ {\widehat{\phi_K({\boldsymbol z})}} \odot
\overline{\widehat{\phi_K({\boldsymbol w})}} \right]\right\}\,,
\label{dft3}
\end{eqnarray}
where~$\overline{{\boldsymbol v}}$ denotes the conjugate of
${\boldsymbol v}$ and~${\sf Re}\left\{{\boldsymbol v}\right\}$ its
real part.

Expression~(\ref{dft3}) provides a computational simple scheme to
compute the shift~$k^*$ that best ``aligns" the ANSIGs
$\phi_K({\boldsymbol z})$ and $\phi_K({\boldsymbol w})$ of two shape
vectors ${\boldsymbol z}$ and ${\boldsymbol w}$: i) compute the DFTs
$\widehat{\phi_K({\boldsymbol z})}$ and
$\widehat{\phi_K({\boldsymbol w})}$; ii) compute the elementwise
product $\widehat{\phi_K({\boldsymbol z})} \odot
\overline{\widehat{\phi_K({\boldsymbol w})}}$; iii) compute the DFT
of this product and locate the entry with largest real part. Since
each DFT is computed by using the FFT algorithm, which has
computational complexity ${\mathcal O}(K \log K)$, the overall
complexity of our comparison scheme is ${\mathcal O}(K + 3 K\log
K)$.

Finally, to measure the similarity between the shapes in
${\boldsymbol z}$ and ${\boldsymbol w}$, {\it i.e.}, the similarity
between their ``aligned'' ANSIGs $\phi_K({\boldsymbol z})$ and
$\phi_K({\boldsymbol w})\,{\mbox{{\sf mod}}}\,k^*$, we use the
(cosine of the) angle between the corresponding vectors:
\begin{equation}
\psi(\bm{z},\bm{w})=\frac{\left|{\phi_K({\boldsymbol z})}^T
\phi_K({\boldsymbol w})\,{\mbox{{\sf mod}}}\,k^*\right|}
{\left\|{\phi_K({\boldsymbol
z})}\right\|\,\left\|\phi_K({\boldsymbol w})\right\|}\,.
\label{eq:comparison}
\end{equation}
Thus, the similarity $\psi(\bm{z},\bm{w})$ is such that
$0\leq\psi(\bm{z},\bm{w})\leq 1$ and the larger is
$\psi(\bm{z},\bm{w})$, the more similar are the shapes in
${\boldsymbol z}$ and ${\boldsymbol w}$.

\subsection{Improving robustness when dealing with interior shape detail}
\label{subsec:robustness}

\begin{figure}[hbt]
\centerline{\epsfig{figure=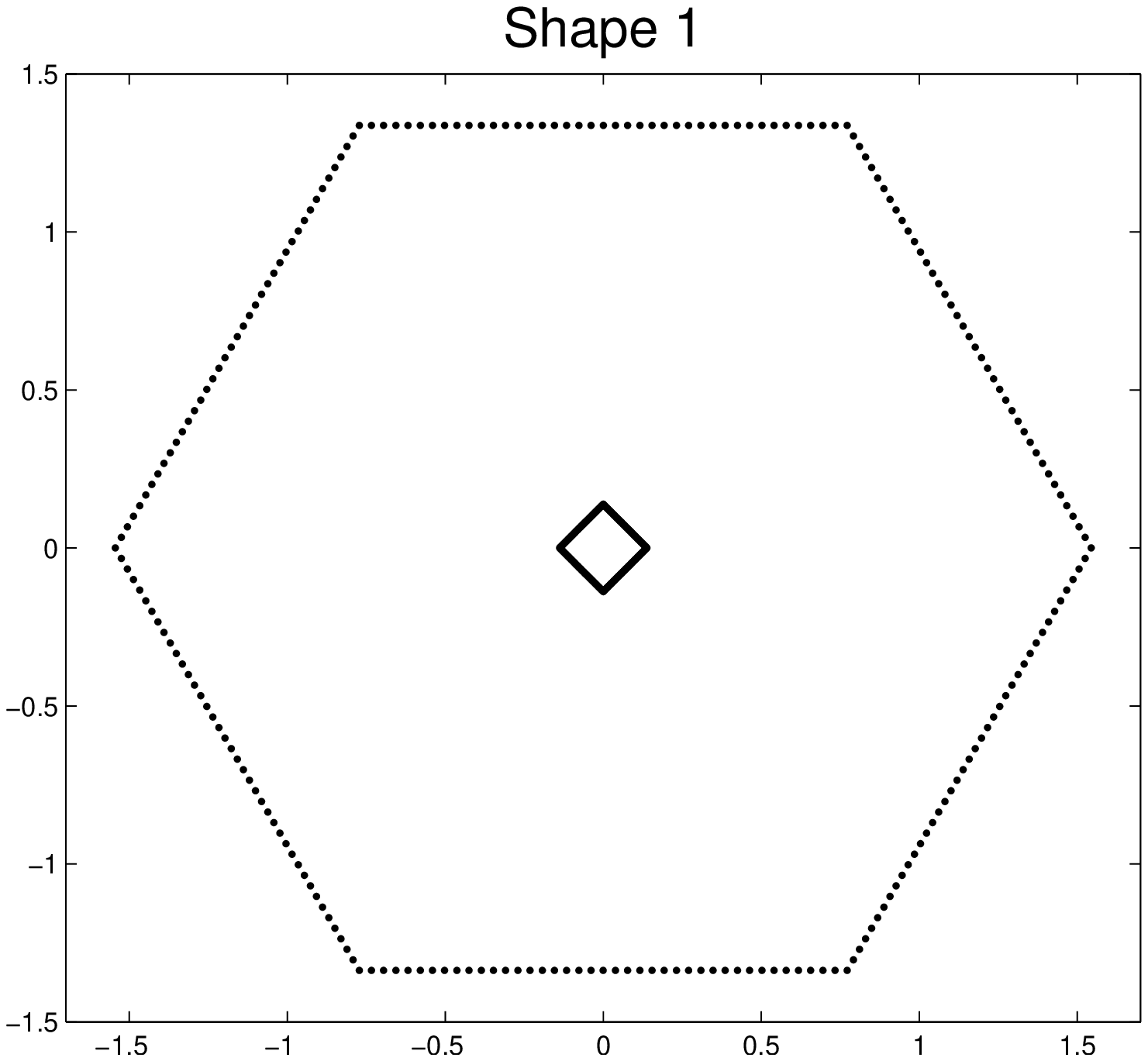,
width=5cm}\hspace*{1cm}
\epsfig{figure=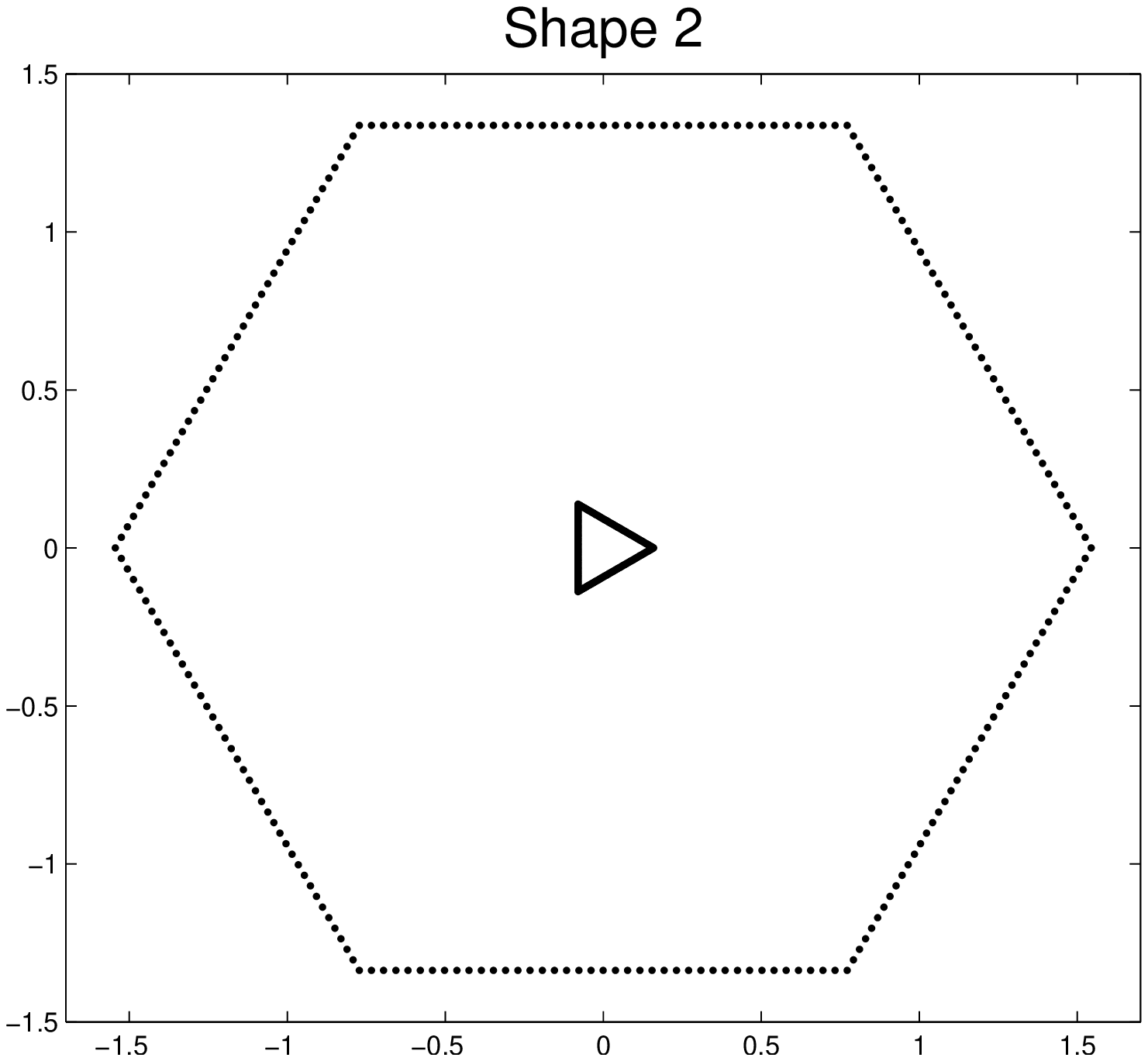,width=5cm}}
\vspace*{-.5cm}\caption{Two shapes that only differ in the points
that are close to their center.\label{fig:illustration4}}
\end{figure}

\begin{figure}[hbt]
\centerline{\epsfig{figure=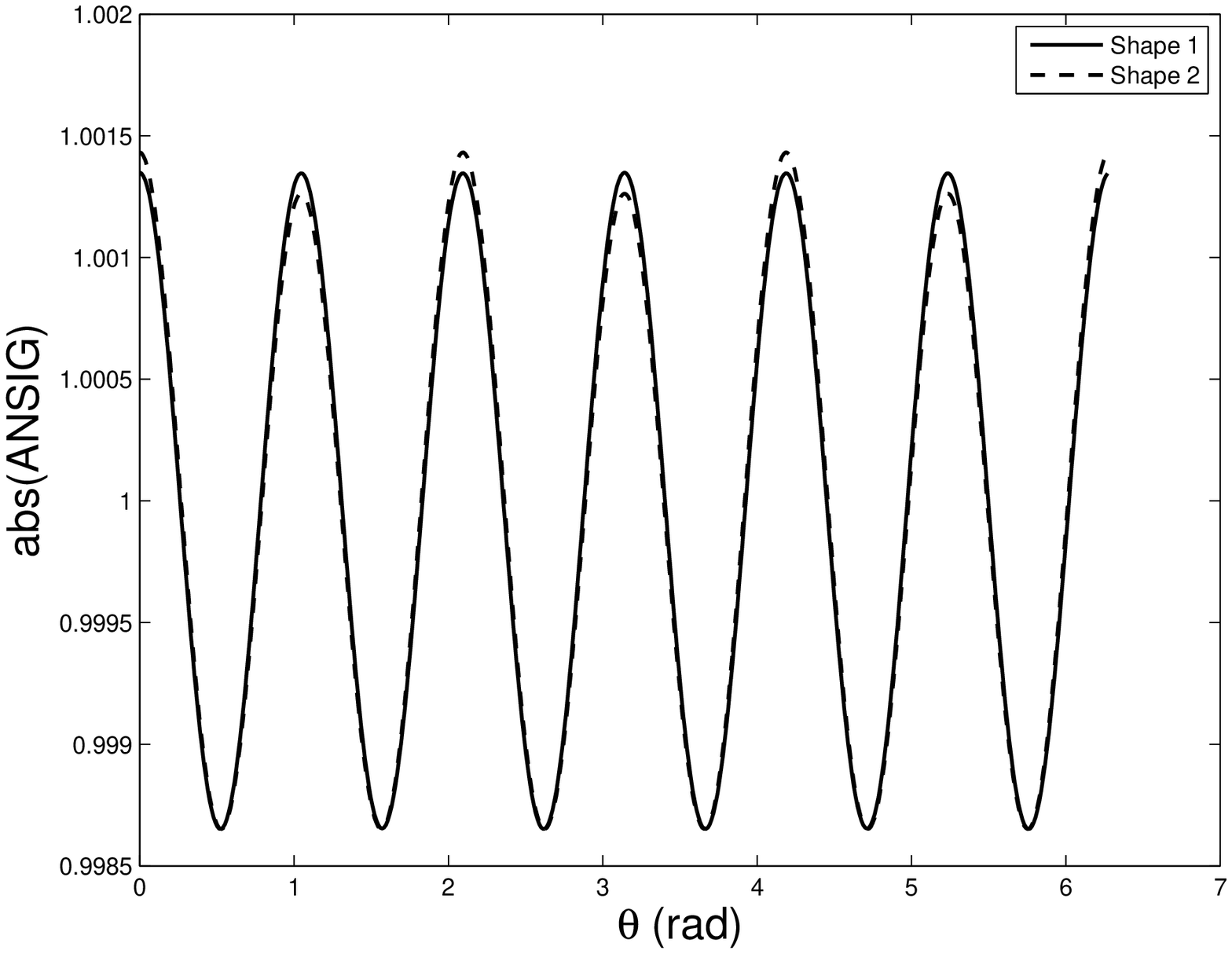,
width=5cm}\hspace*{1cm}
\epsfig{figure=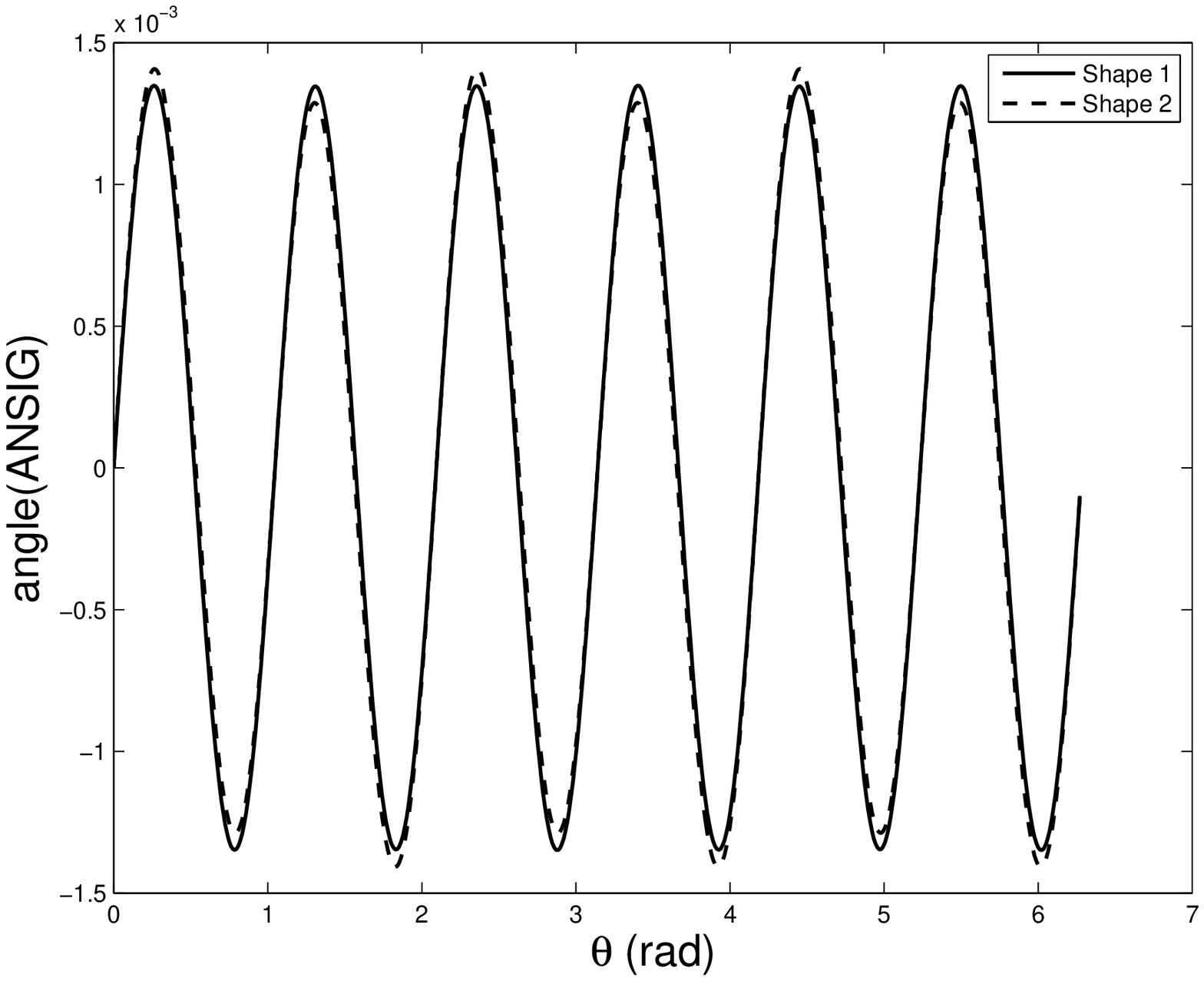,width=5cm}}
\vspace*{-.5cm}\caption{ANSIGs of the two shapes in
Fig.~\ref{fig:illustration4}. The similarity of the exterior
hexagons makes the signatures very
similar.\label{fig:illustration5}}
\end{figure}

Before presenting experiments with shape classification, we
anticipate that when a single ANSIG is used to describe a shape
characterized by having specific details occurring at very different
distances from the geometric center, the representation may lack the
desired robustness. This is illustrated by the example in
Figs.~\ref{fig:illustration4} and~\ref{fig:illustration5}.
Fig.~\ref{fig:illustration4} shows two shapes characterized by a
similar large outer hexagon and distinct small inner polygons. The
ANSIGs of these two shapes are represented in
Fig.~\ref{fig:illustration5}. As easily perceived, these signatures
result very similar. Although one could argue that the shapes in
Fig.~\ref{fig:illustration4} are in fact similar, the argument would
be misleading, since if the small polygons were not in the center
but in the periphery of the image, the ANSIGs would be far more
distinct. We now briefly discuss this issue and ways to deal with
it.

\noindent{\bf Weighted ANSIG.} The similarity of the ANSIGs in
Fig.~\ref{fig:illustration5} is explained by the exponential
weighting of each landmark modulus ({\it i.e.}, distance to the
geometric center), when computing the map $a$ in \eqref{ansig}, in
which the ANSIG is based. In fact, representing the elements of a
shape vector ${\boldsymbol z}$ by their polar coordinates, {\it
i.e.}, $z_m=\rho_m e^{j\theta_m}$, the map $a$ in \eqref{ansig} is
written as
\begin{equation}
a({\boldsymbol z},\xi) =\frac{1}{n} \sum_{m=1}^n e^{\rho_m
e^{j\theta_m} \xi}=\frac{1}{n} \sum_{m=1}^n
e^{\rho_m}e^{e^{j\theta_m} \xi}\,, \label{ansig2}
\end{equation}
which shows that landmarks at large distances from the origin ({\it
i.e.}, with large $\rho_m$), are taken into much larger account
(exponential weighting) than those closer to the center (small
$\rho_m$).

One approach to tackle this problem is precisely to weight the
landmarks in a different way, with the care of maintaining the
maximal invariance property. For example, if the shapes are
preprocessed by an elementwise map~$g$, that maps each entry $\rho_m
e^{j\theta_m}$ to $\log\left(1+\rho_m\right) e^{j\theta_m}$, the
signature obtained corresponds to replacing the map $a$ in
\eqref{ansig2} by a map $a_{g}$, given~by:
\begin{eqnarray}
a_g\left({\boldsymbol z},\xi\right)=a(g({\boldsymbol
z}),\xi)=\frac{1}{n} \sum_{m=1}^n (1+\rho_m)^{e^{j\theta_m} \xi}\,,
\label{ansig4}
\end{eqnarray}
which now takes into account the modulus of the landmarks, {\it
i.e.}, their distance to the origin, in a linear way. It is
straightforward to show that the maximal invariance of the ANSIG is
not affected, because $g$ is an injection, {\it i.e.}, there is a
one-to-one mapping between ${\boldsymbol z}$ and~$g({\boldsymbol
z})$.

%

\begin{figure}[hbt]
\centerline{\epsfig{figure=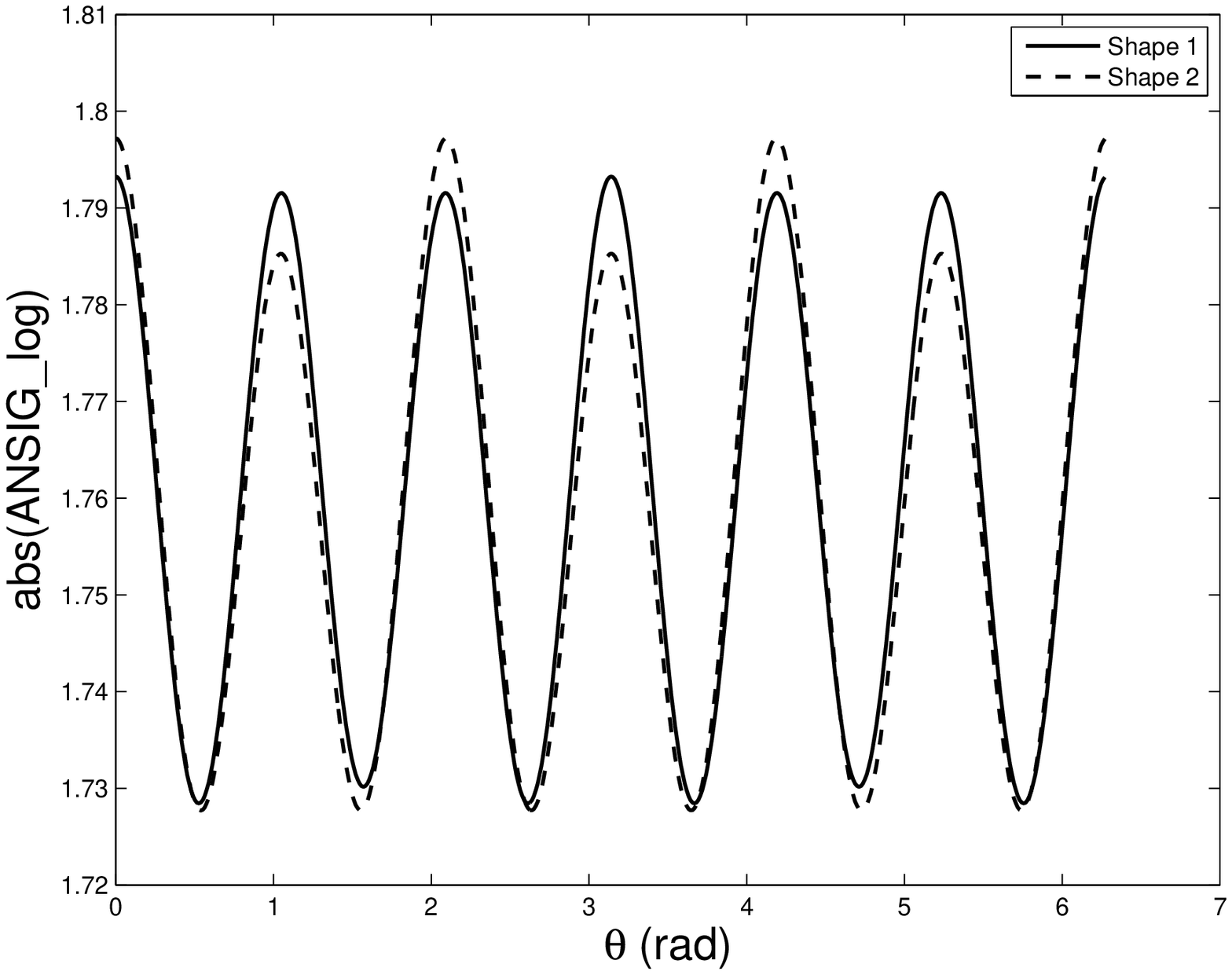,
width=5cm}\hspace*{1cm}
\epsfig{figure=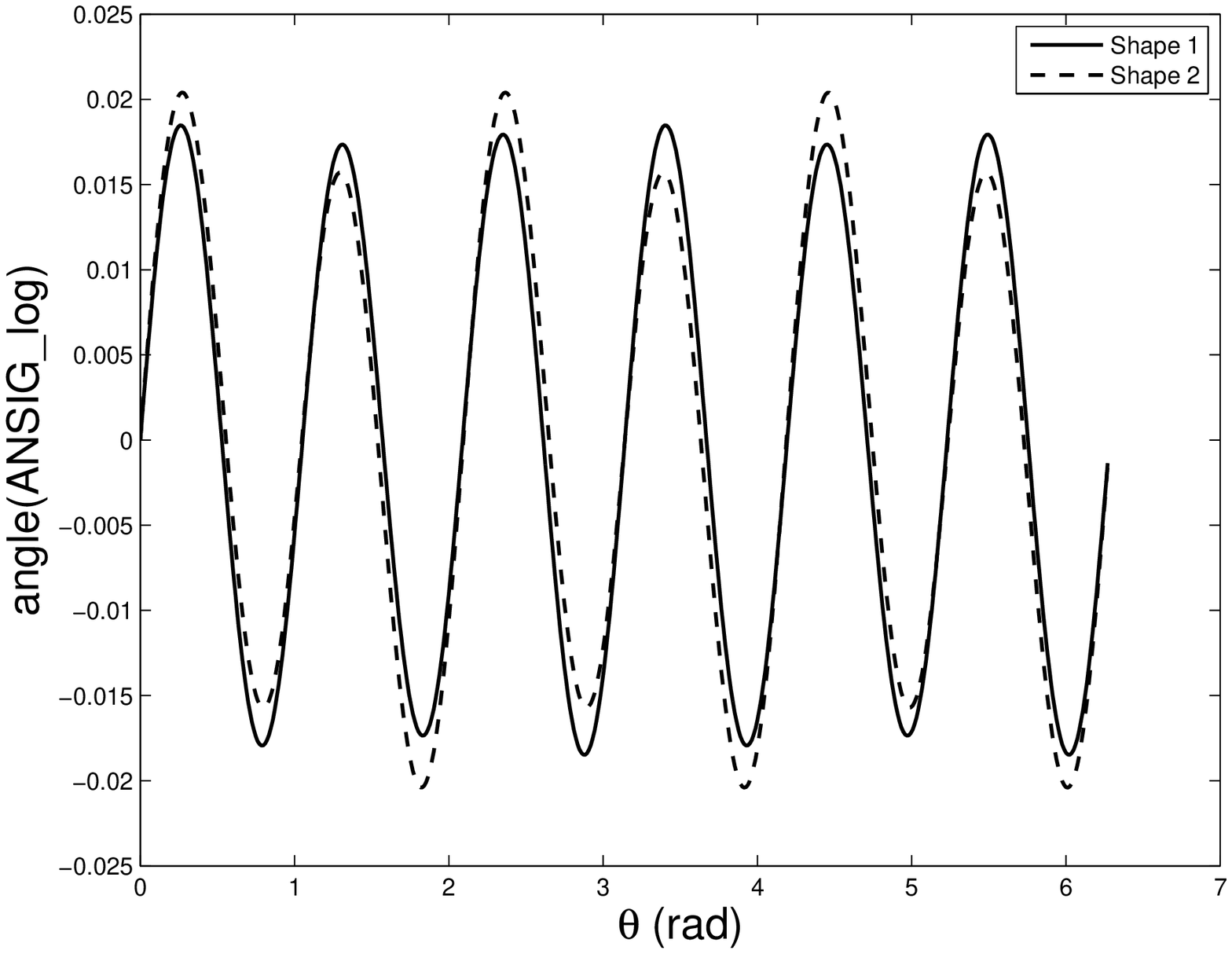,width=5cm}}
\vspace*{-.5cm}\caption{Magnitude and phase of the ANSIGs of the two
shapes in Fig.~\ref{fig:illustration4}, when the map $a$ in
\eqref{ansig} is replaced by the weighted version $a_{g}$
in~(\ref{ansig4}). As desired, the signatures result more distinct
than the ones in Fig.~\ref{fig:illustration5}.
\label{fig:illustration6}}
\end{figure}

In Fig.~\ref{fig:illustration6} we shown the ANSIGs obtained for the
shapes of Fig.~\ref{fig:illustration4}, now using the differently
weighted map $a_{g}$ in~(\ref{ansig4}). Comparing the plots of
Figs.~\ref{fig:illustration5} and~\ref{fig:illustration6}, we see
that the shape signatures are in fact more dissimilar, as desired,
when the weighted version of the ANSIG is used.

\begin{figure}[hbt]
\centerline{\epsfig{figure=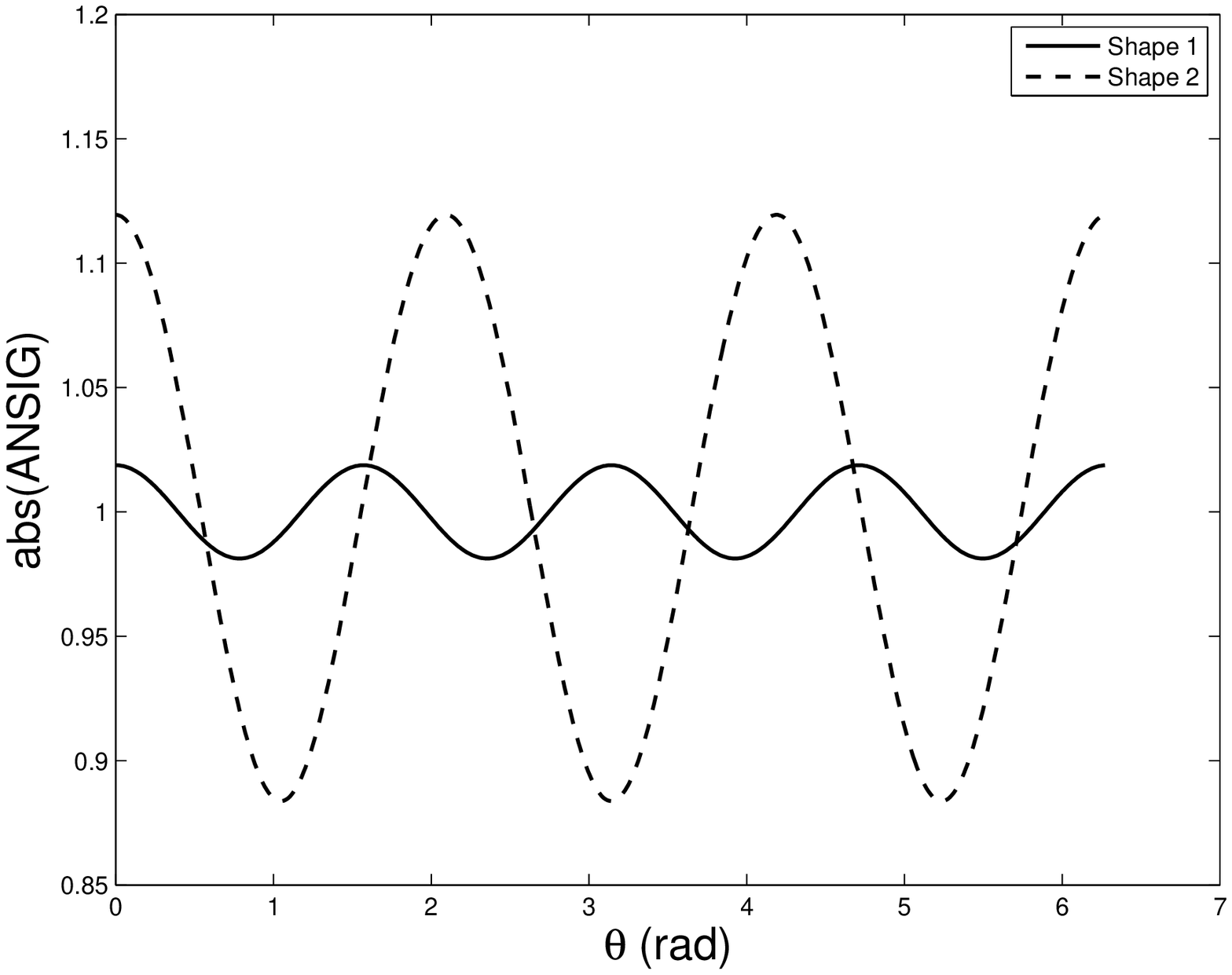,
width=5cm}\hspace*{1cm}
\epsfig{figure=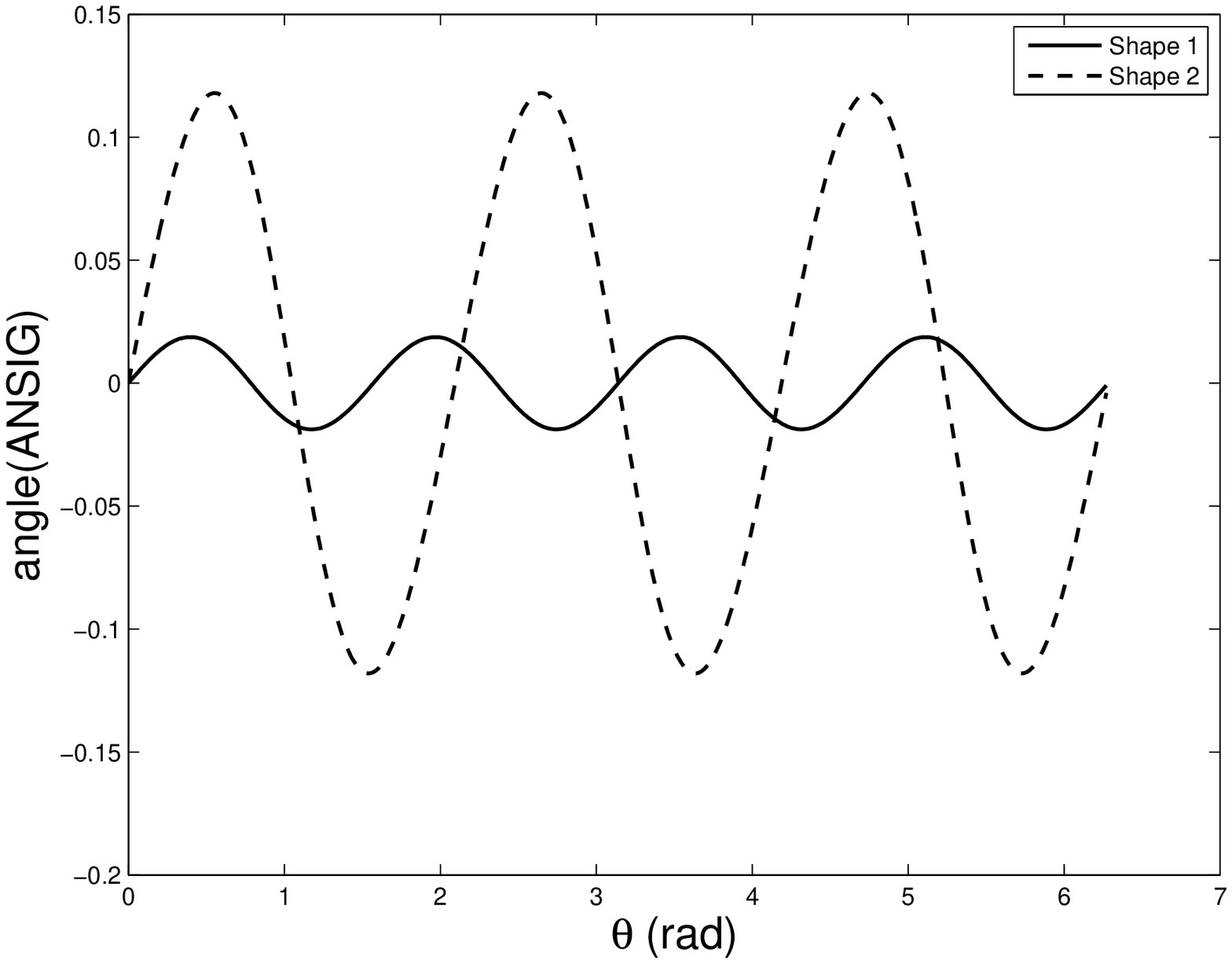,width=5cm}}
\vspace*{-.5cm}\caption{ANSIGs of the small square and triangle in
the shapes of Fig.~\ref{fig:illustration4}. Naturally, when this
inner part is taken into account separately, the signatures result
very distinct. \label{fig:illustration7}}
\end{figure}

\noindent{\bf Using more than one ANSIG.} 
%
Rather than trying to describe shapes such as the ones in
Fig.~\ref{fig:illustration4} with a single signature, an approach
that must be considered is to use multiple descriptions. In fact, if
we compute the ANSIGs of just the inner parts of these shapes, {\it
i.e.}, of the smaller inner polygons (the square and the triangle),
we obtain the plots in Fig.~\ref{fig:illustration7}, which are very
distinct. This motivates the use of more than one ANSIG to
distinguish between shapes like the ones in
Fig.~\ref{fig:illustration4}, {\it i.e.}, shapes with similar outer
content.

Although several hypothesis could be considered, we describe a
simple way to implement a shape representation scheme based on two
ANSIGs: the ANSIG of the entire set of landmarks,
$\phi\left({\boldsymbol z}\right)$, and the ANSIG of some subset of
inner landmarks, $\phi\left(\bm{z}_{{in}}\right)$. Naturally,
$\bm{z}_{{in}}$ can be obtained from $\bm{z}$ in several ways, {\it
e.g.}, by selecting the subset of points that have normalized
absolute value smaller than one.
%
%
Note that this joint signature keeps the invariance properties of
the single ANSIG representation: the maximal invariance is obvious,
since we maintain the ANSIG of the complete shape; the rotational
equivariance comes from the fact that the partition is circular.

When comparing shapes ${\boldsymbol z}$ and ${\boldsymbol w}$ under
this framework, the cyclic-shift~$k^*$ that best ``aligns"
simultaneously both pairs of ANSIG vectors,
$\phi_K\left(\bm{z}\right)$ with $\phi_K\left(\bm{w}\right)$ and
$\phi_K\left(\bm{z}_{{in}}\right)$ with
$\phi_K\left(\bm{w}_{{in}}\right)$ is obtained, by following
 steps similar to the ones in \eqref{align} to \eqref{dft3}, as:
\begin{eqnarray}
k^* &=& \arg\min_{k} \left\| \phi_K({\boldsymbol z}) - \phi_K
({\boldsymbol w})\,\mbox{{\sf mod}}\,k \right\|^2 + \left\|
\phi_K({\boldsymbol z}_{in}) - \phi_K ({\boldsymbol w}_{in})\,
\mbox{{\sf mod}}\,k \right\|^2\,. \\
&=&\arg\max_{k}\, {\sf Re} \left\{{\boldsymbol d}_k^H \left[
{\widehat{\phi_K({\boldsymbol z})}} \odot
\overline{\widehat{\phi_K({\boldsymbol w})}}  +
{\widehat{\phi_K({\boldsymbol z_{in}})}} \odot
\overline{\widehat{\phi_K({\boldsymbol w_{in}})}} \right]\right\}\,.
\end{eqnarray}
To measure the two-ANSIG similarity, we can use an weighted average
of the similarities $\psi\left(\bm{z},\bm{w}\right)$
and~$\psi\left(\bm{z}_{in},\bm{w}_{in}\right)$, defined
in~(\ref{eq:comparison}). A natural option is to weight each partial
similarity by the corresponding number of shape points that are
taken into account.

\section{Experiments}
\label{sec:exp}

We now present experiments that demonstrate the usefulness of the
ANSIG in shape-based classification. We start by evaluating the
robustness to noise, using synthetic data. Then, we illustrate an
application to automatic trademark retrieval, using real images.
Finally, we discuss the behavior of the ANSIG representation when in
presence of model violations.

\subsection{Robustness to noise}
\label{subsec:RN}

\begin{figure}[hbt]\vspace*{-.5cm}
\centerline{\epsfig{figure=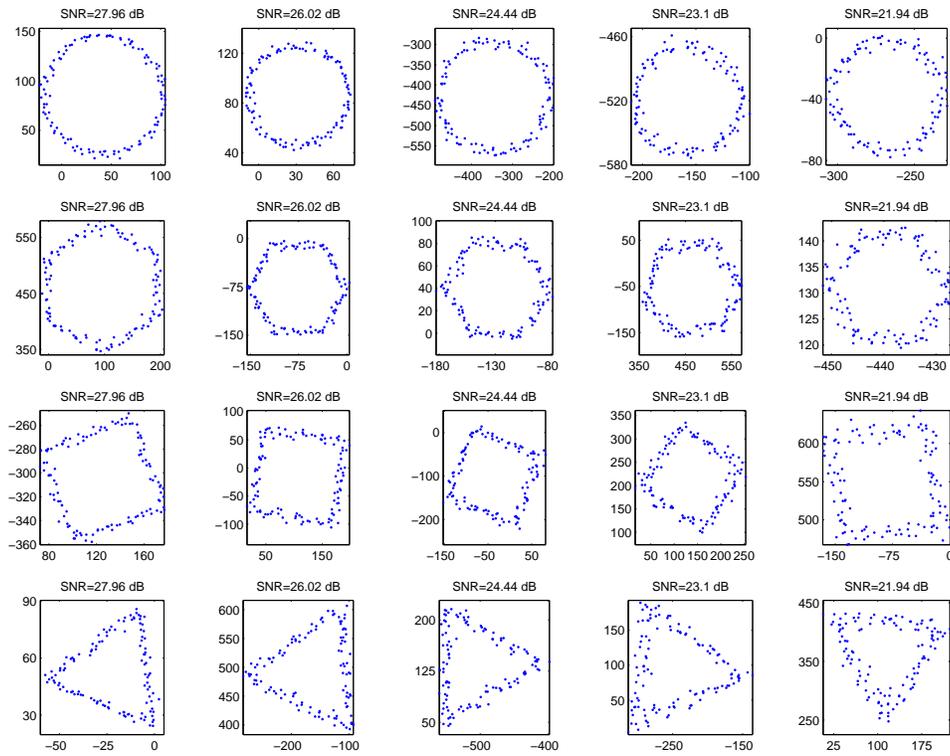,
width=16cm}}\vspace*{-1cm} \vspace*{-.5cm}\caption{Sample test
shapes. From top to bottom, the circle, the hexagon, the square, and
the triangle, with varying noise level, translation, rotation, and
scale factors, and point labeling. Noise increases from left to
right. Note that, for high levels of noise, the shapes become
difficult to distinguish, particularly the circumference and the
hexagon become very similar.\label{fig:noise_levels}}
\end{figure}

To evaluate the robustness of the ANSIG representation to the noise
affecting the landmark positions, we build a particularly
challenging scenario with a database of four geometric shapes
--- a circumference, an hexagon, a square, and a triangle --- that
are difficult to distinguish when in presence of noise. In
Fig.~\ref{fig:noise_levels}, we show sample test shapes. They are
noisy (and translated, rotated, scaled, and re-ordered) versions of
the four geometric shapes. Note how the circle and the hexagon
become similar for high levels of noise.

We then used the classification strategy described in the previous
section: our database was composed just by the ANSIGs of the four
noiseless polygons and the ANSIG of each noisy test shape was
classified by selecting the most similar entry in the database. We
tested the classifier by performing 2000 tests for each of the four
shapes, at each noisy level, obtaining 100\% correct classifications
for in all tests with SNR above around 28dB. Note that shapes with
this level of noise are far from being visually ``clean", see the
illustrations in the first column (the leftmost plots) of
Fig.~\ref{fig:noise_levels}. The results are summarized in
Tab.~\ref{class_results}, where we see that a few classifications
errors happen for higher levels of noise, when dealing with circles
or hexagons, which as noted before, for such levels of noise, are in
fact difficult to distinguish, even for humans, see the rightmost
plots of the first two lines of Fig.~\ref{fig:noise_levels}.

    \begin{table}[hbt]
        \caption{Percentages of correct classifications for the noisy shapes in Fig.~\ref{fig:noise_levels}.}\vspace*{-.5cm}
        \renewcommand{\arraystretch}{1.3}
        \centering
        \begin{tabular}{|c|c|c|c|c|c|}
        \hline
        &\multicolumn{5}{c|}{SNR [dB]}\\
        \cline{2-6}
        & $\geq$ 27.96&26.02&24.44&23.1&21.94\\
        \hline
        \epsfig{figure=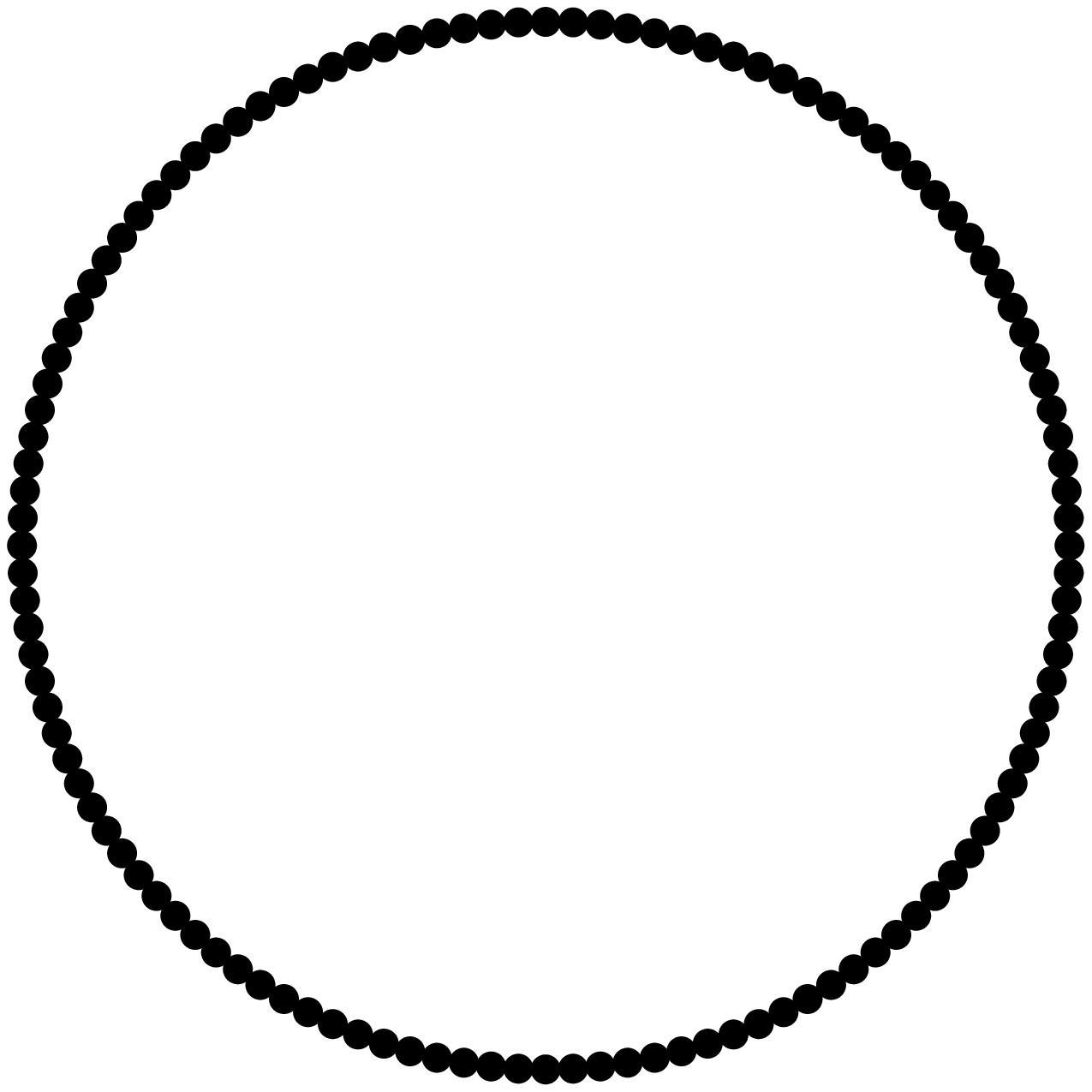, height=.5cm}&100\% &99.95\% &99.9\% &99.45\% &97.3\% \\
        \epsfig{figure=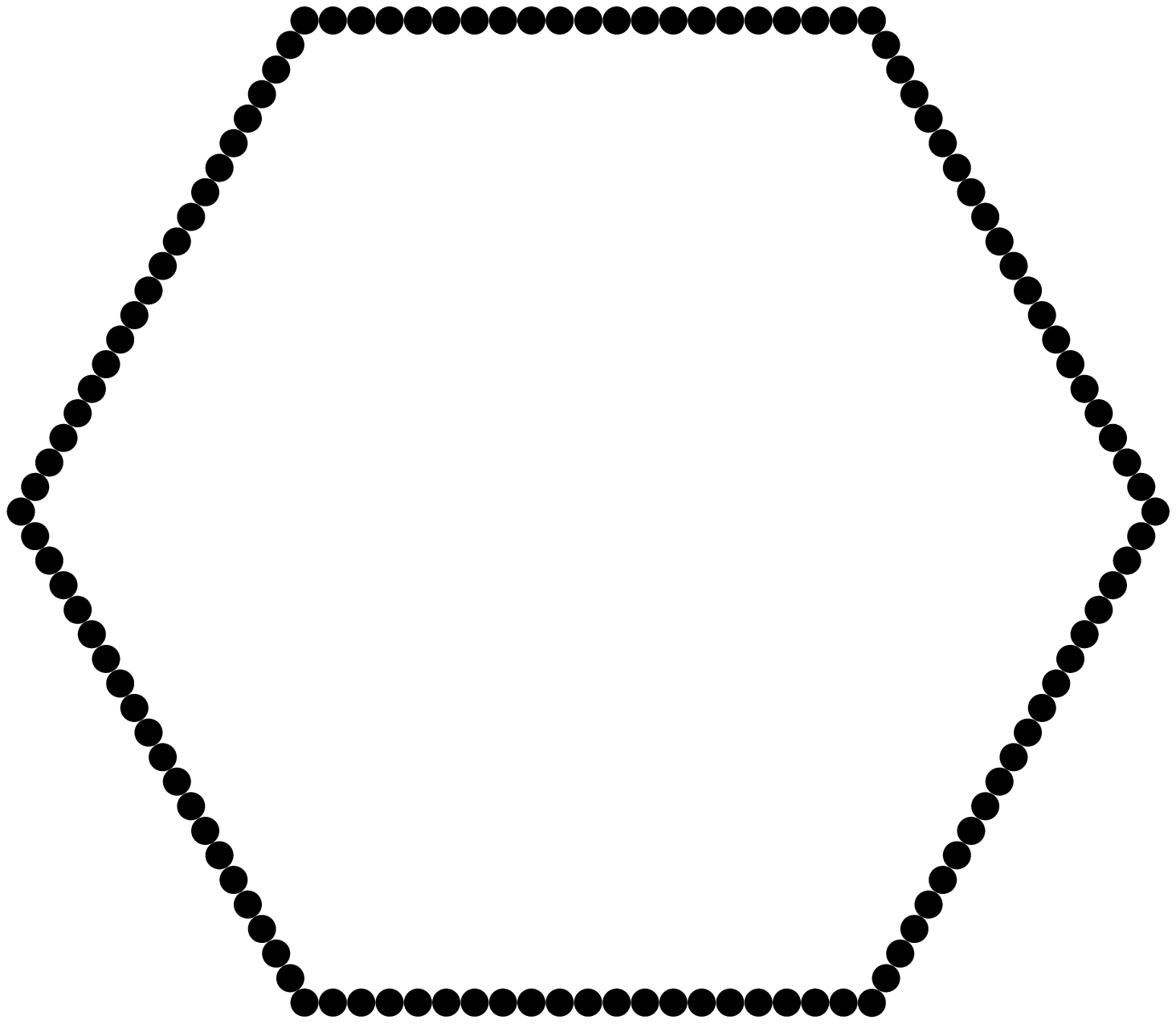, height=.5cm}&100\% &100\% &99.9\% &99.85\% &99.25\% \\
        \epsfig{figure=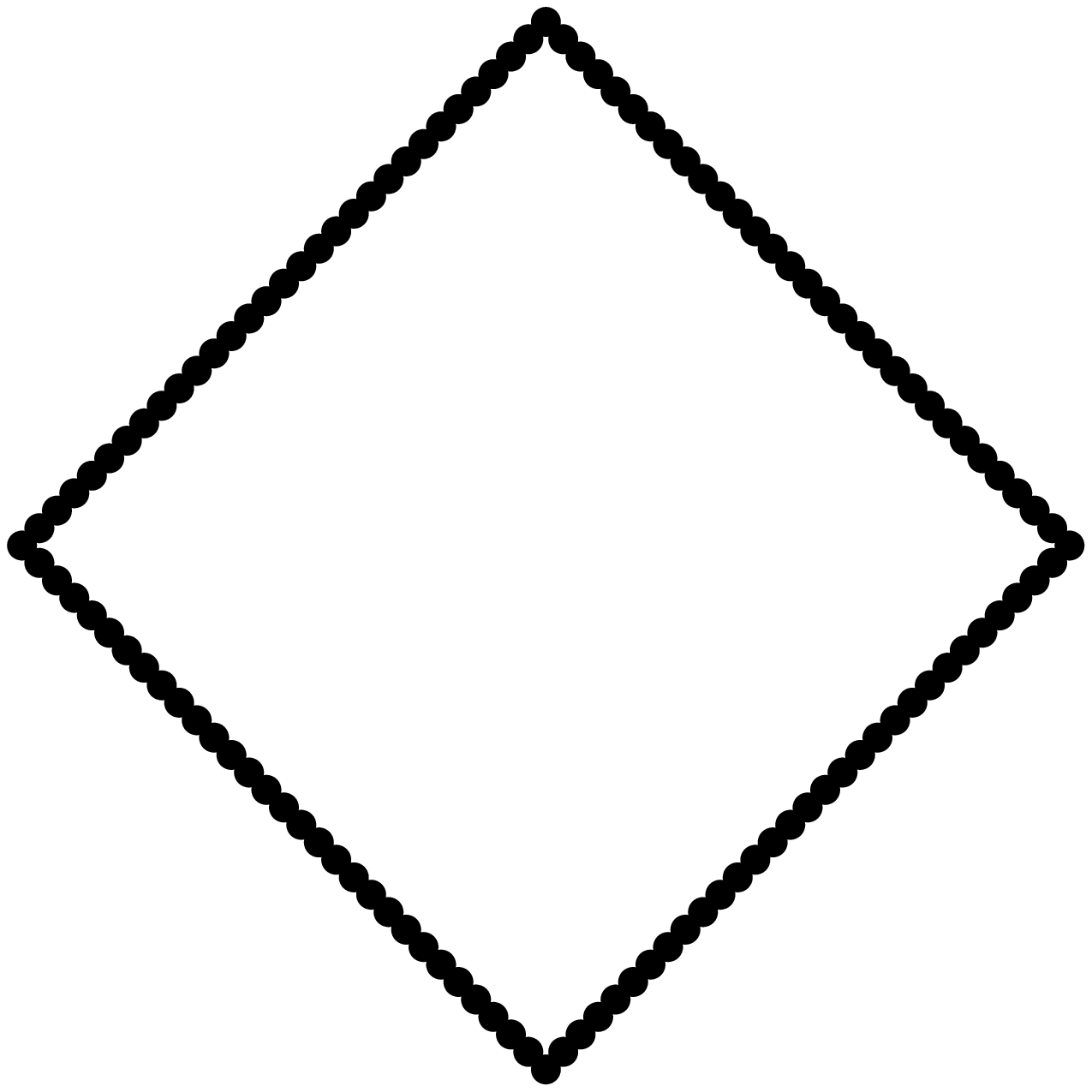, height=.5cm}&100\% &100\% &100\% &100\% &100\% \\
        \epsfig{figure=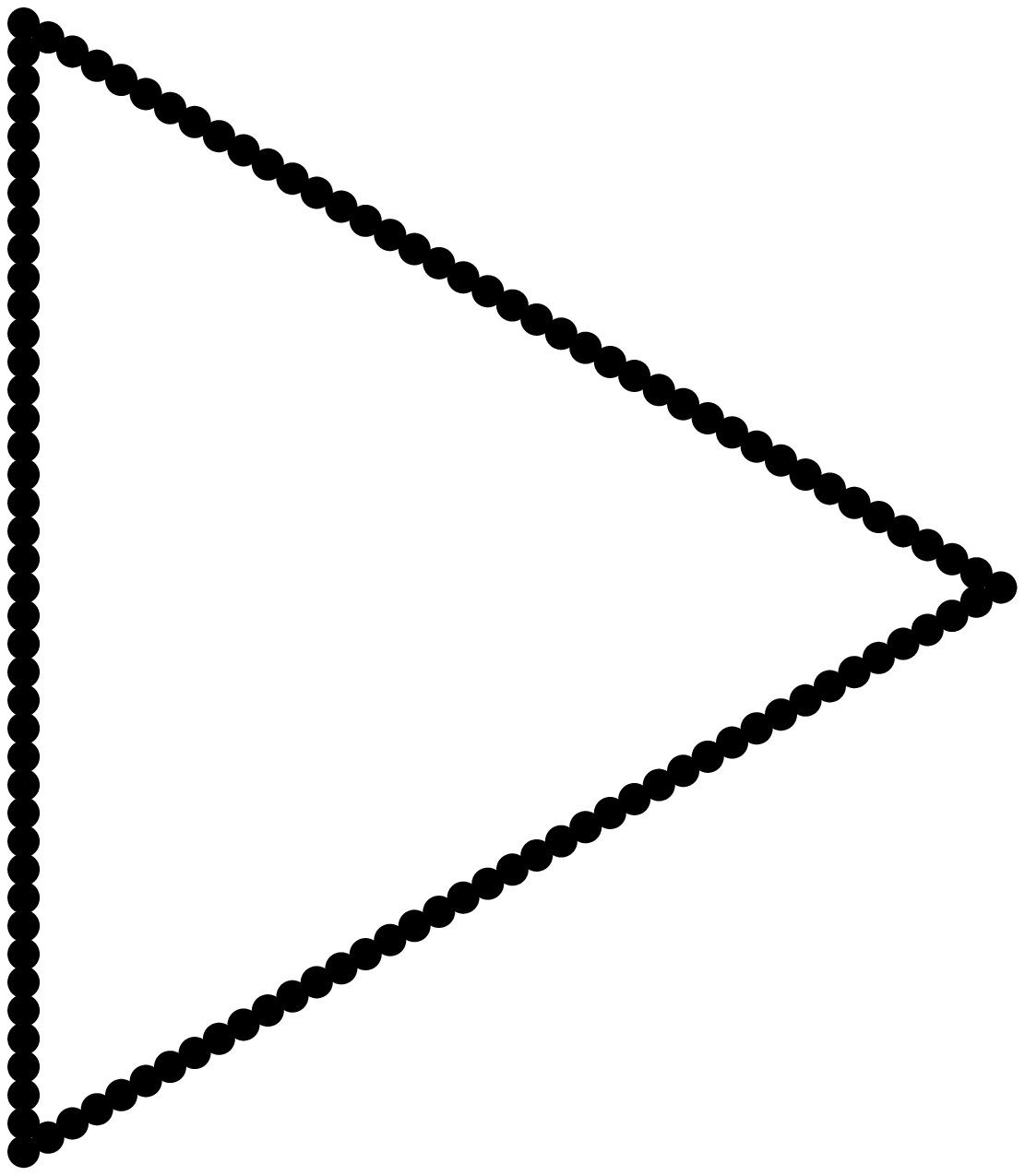, height=.5cm}&100\% &100\% &100\% &100\% &100\% \\
        \hline
        \end{tabular}  \label{class_results}
    \end{table}

\begin{figure}[hbt]
  \centerline{
    \epsfig{figure=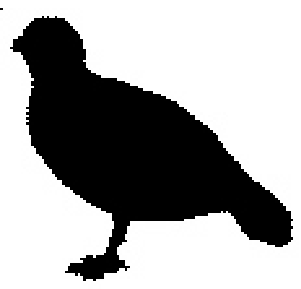, width=.55cm}
        \epsfig{figure=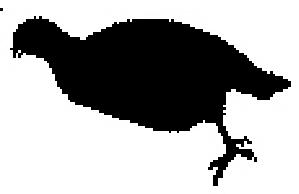, width=.55cm}
        \epsfig{figure=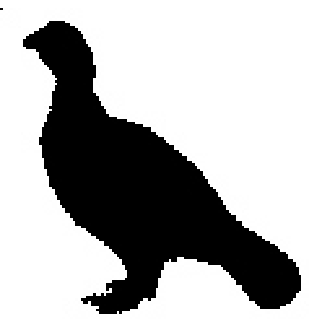, width=.55cm}
        \epsfig{figure=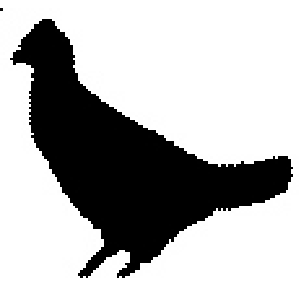, width=.55cm}
        \epsfig{figure=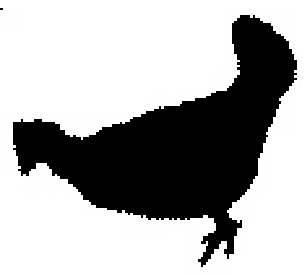, width=.55cm}
        \epsfig{figure=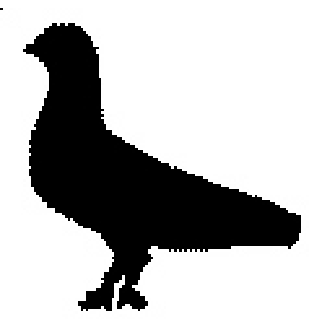, width=.55cm}
        \epsfig{figure=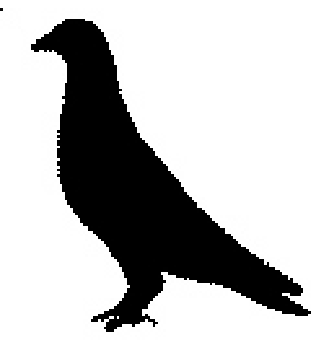, width=.55cm}
        \epsfig{figure=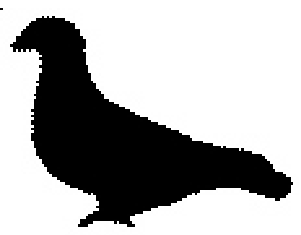, width=.55cm}
        \epsfig{figure=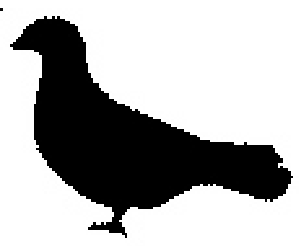, width=.55cm}
        \epsfig{figure=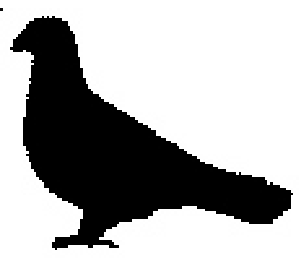, width=.55cm}
        \epsfig{figure=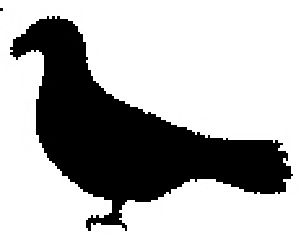, width=.55cm}
        \epsfig{figure=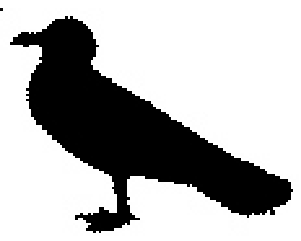,
        width=.55cm}
        \epsfig{figure=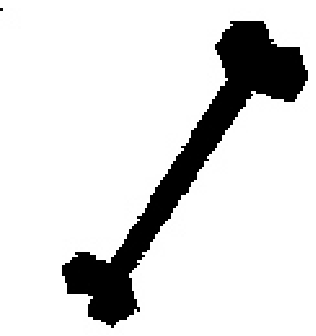, width=.55cm}
        \epsfig{figure=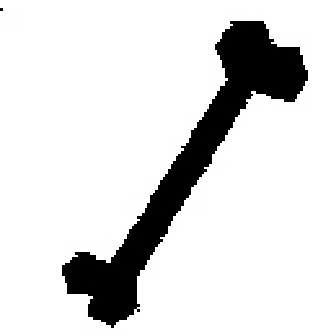, width=.55cm}
        \epsfig{figure=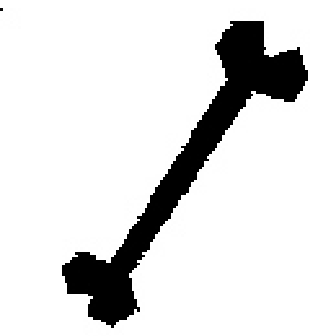, width=.55cm}
        \epsfig{figure=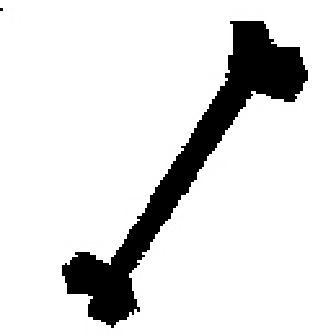, width=.55cm}
        \epsfig{figure=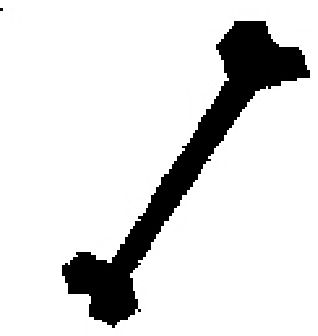, width=.55cm}
        \epsfig{figure=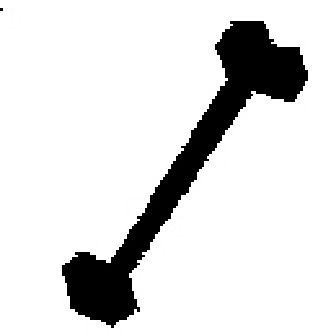, width=.55cm}
        \epsfig{figure=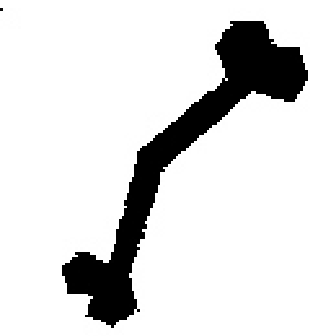, width=.55cm}
        \epsfig{figure=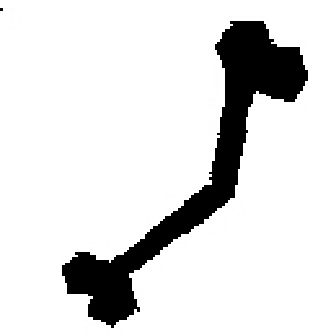, width=.55cm}
        \epsfig{figure=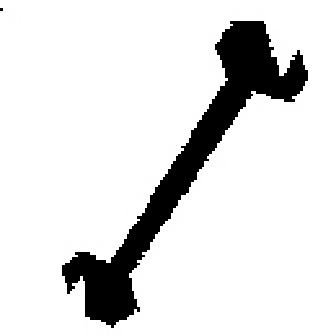, width=.55cm}
        \epsfig{figure=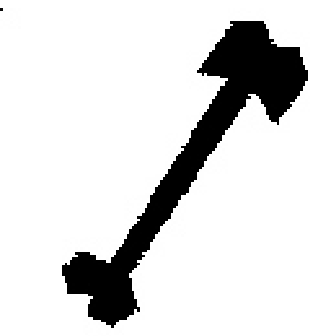, width=.55cm}
        \epsfig{figure=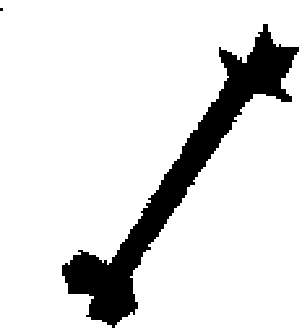, width=.55cm}
        \epsfig{figure=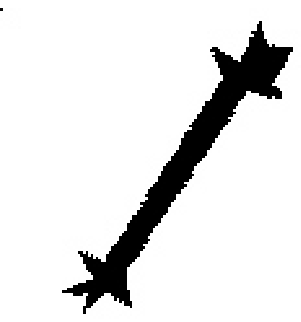, width=.55cm}}
    \centerline{
        \epsfig{figure=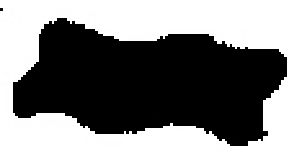, width=.55cm}
        \epsfig{figure=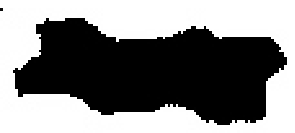, width=.55cm}
        \epsfig{figure=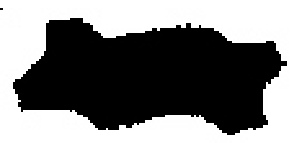, width=.55cm}
        \epsfig{figure=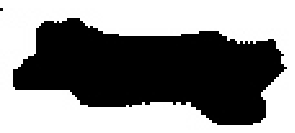, width=.55cm}
        \epsfig{figure=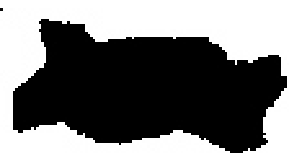, width=.55cm}
        \epsfig{figure=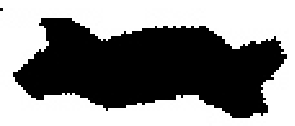, width=.55cm}
        \epsfig{figure=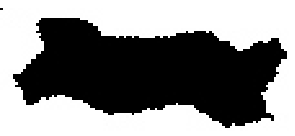, width=.55cm}
        \epsfig{figure=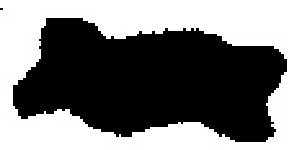, width=.55cm}
        \epsfig{figure=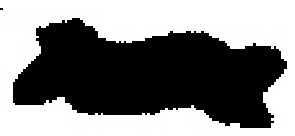, width=.55cm}
        \epsfig{figure=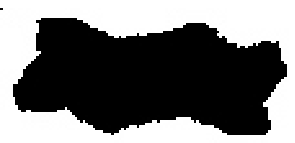, width=.55cm}
        \epsfig{figure=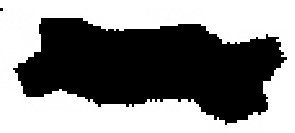, width=.55cm}
        \epsfig{figure=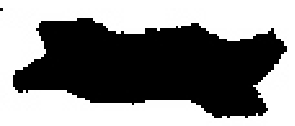, width=.55cm}
        \epsfig{figure=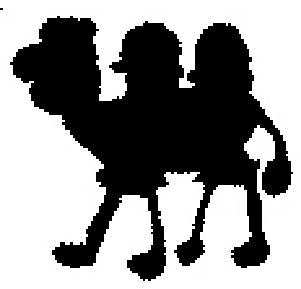, width=.55cm}
        \epsfig{figure=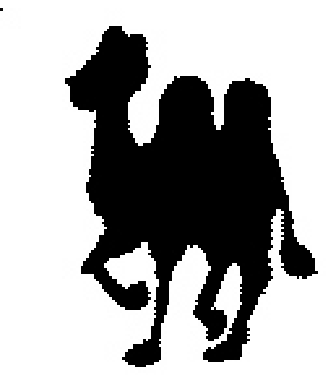, width=.55cm}
        \epsfig{figure=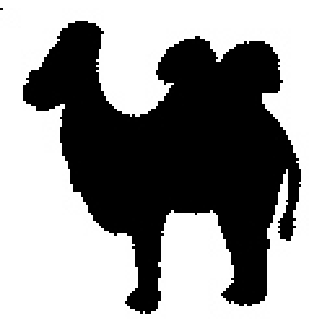, width=.55cm}
        \epsfig{figure=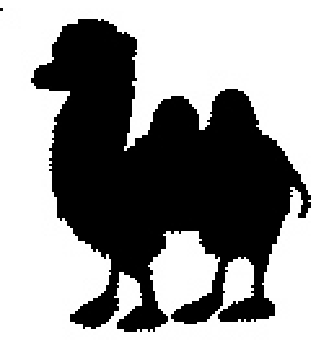, width=.55cm}
        \epsfig{figure=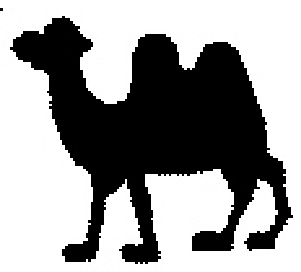, width=.55cm}
        \epsfig{figure=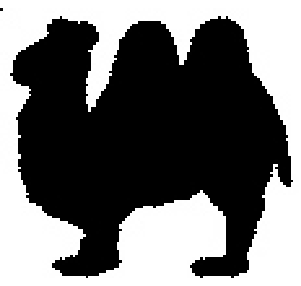, width=.55cm}
        \epsfig{figure=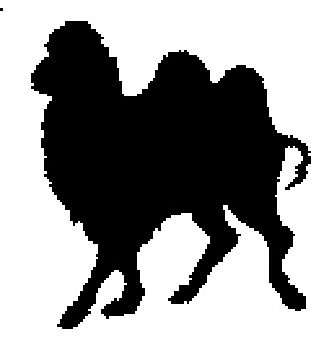, width=.55cm}
        \epsfig{figure=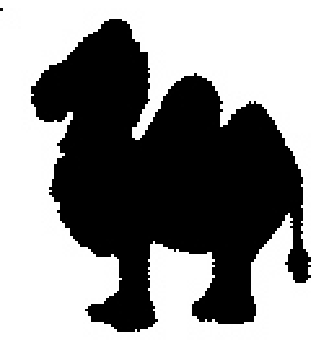, width=.55cm}
        \epsfig{figure=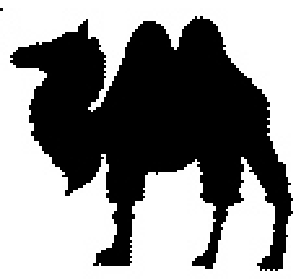, width=.55cm}
        \epsfig{figure=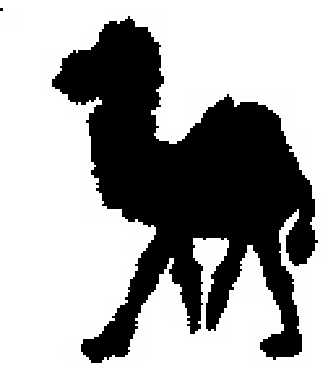, width=.55cm}
        \epsfig{figure=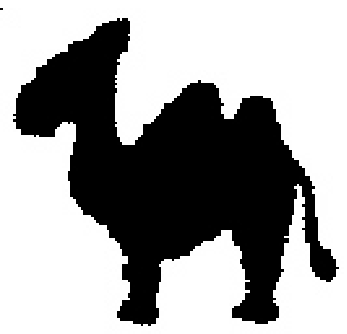, width=.55cm}
        \epsfig{figure=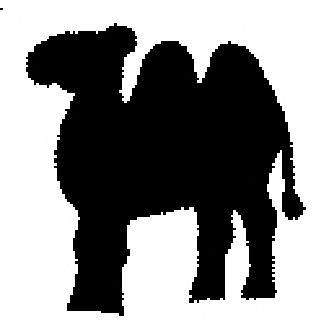, width=.55cm}}\centerline{
        \epsfig{figure=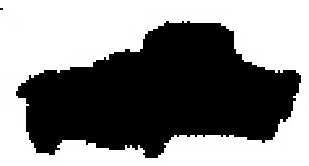, width=.55cm}
        \epsfig{figure=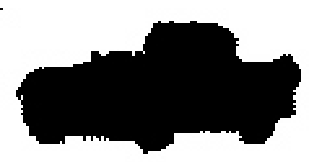, width=.55cm}
        \epsfig{figure=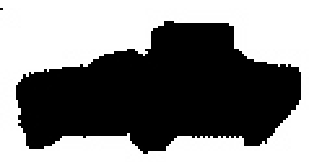, width=.55cm}
        \epsfig{figure=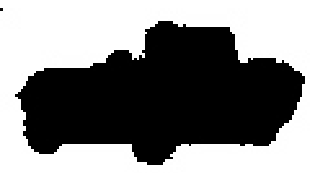, width=.55cm}
        \epsfig{figure=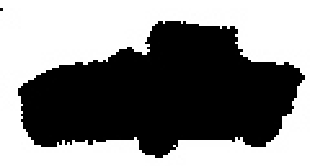, width=.55cm}
        \epsfig{figure=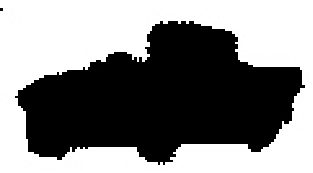, width=.55cm}
        \epsfig{figure=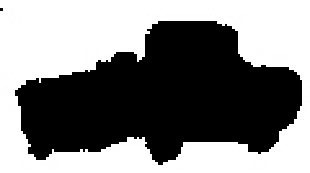, width=.55cm}
        \epsfig{figure=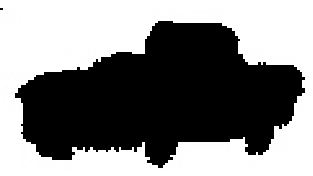, width=.55cm}
        \epsfig{figure=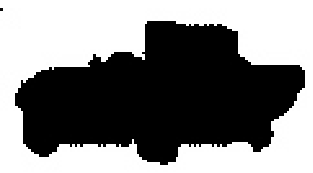, width=.55cm}
        \epsfig{figure=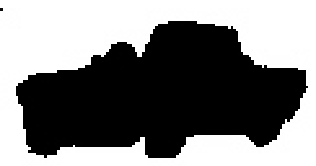, width=.55cm}
        \epsfig{figure=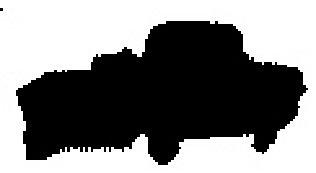, width=.55cm}
        \epsfig{figure=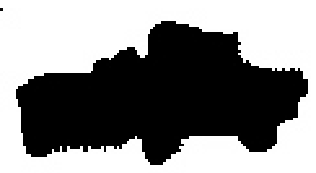, width=.55cm}
        \epsfig{figure=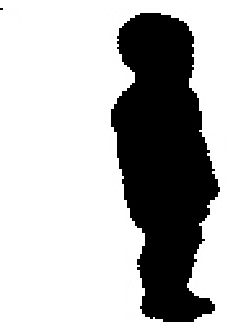, width=.55cm}
        \epsfig{figure=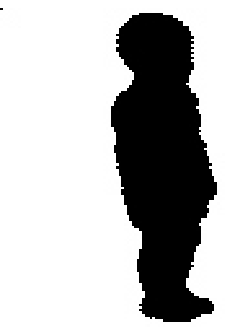, width=.55cm}
        \epsfig{figure=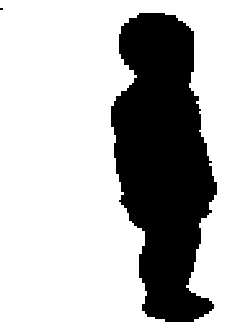, width=.55cm}
        \epsfig{figure=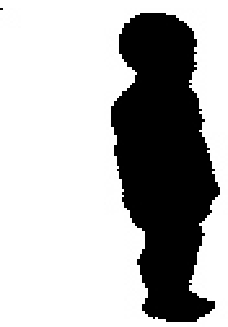, width=.55cm}
        \epsfig{figure=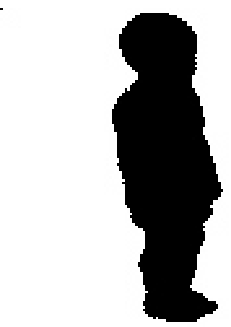, width=.55cm}
        \epsfig{figure=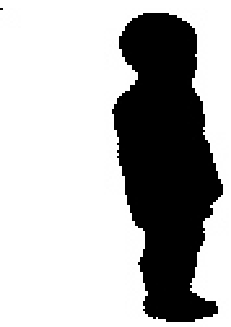, width=.55cm}
        \epsfig{figure=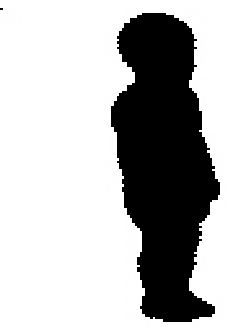, width=.55cm}
        \epsfig{figure=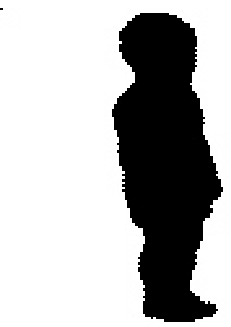, width=.55cm}
        \epsfig{figure=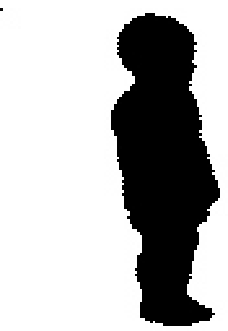, width=.55cm}
        \epsfig{figure=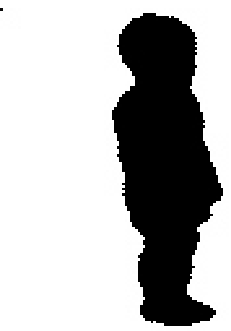, width=.55cm}
        \epsfig{figure=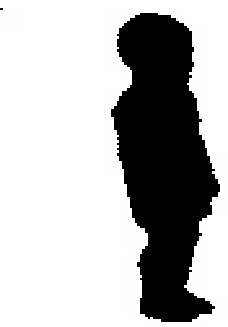, width=.55cm}
        \epsfig{figure=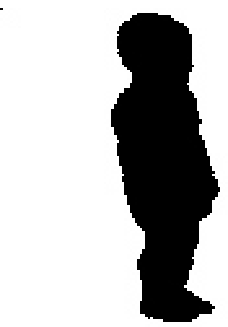, width=.55cm}}\centerline{
        \epsfig{figure=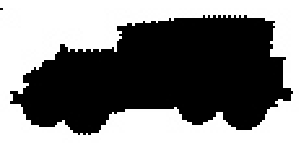, width=.55cm}
        \epsfig{figure=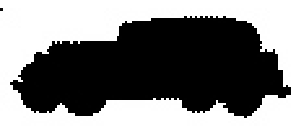, width=.55cm}
        \epsfig{figure=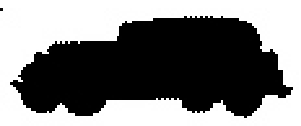, width=.55cm}
        \epsfig{figure=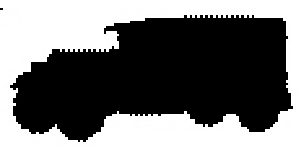, width=.55cm}
        \epsfig{figure=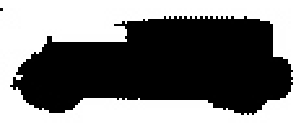, width=.55cm}
        \epsfig{figure=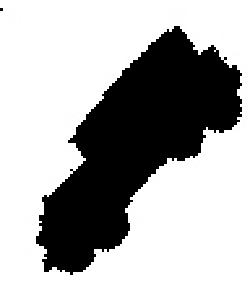, width=.55cm}
        \epsfig{figure=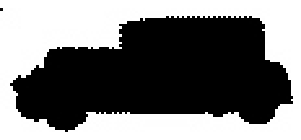, width=.55cm}
                \epsfig{figure=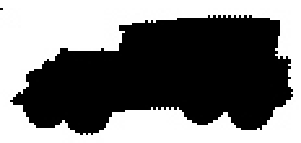, width=.55cm}
        \epsfig{figure=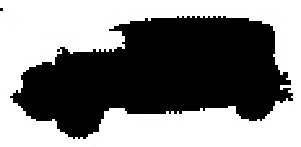, width=.55cm}
        \epsfig{figure=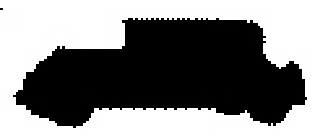, width=.55cm}
        \epsfig{figure=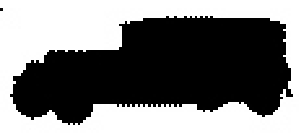, width=.55cm}
        \epsfig{figure=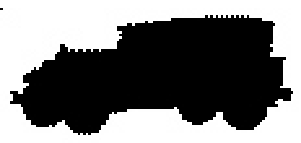, width=.55cm}
        \epsfig{figure=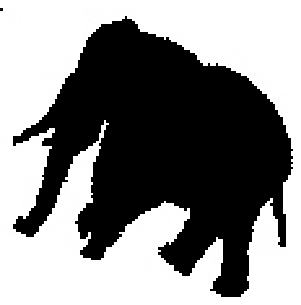, width=.55cm}
        \epsfig{figure=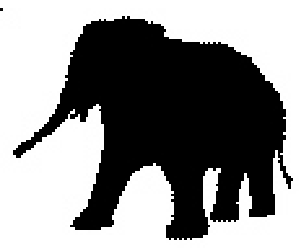, width=.55cm}
        \epsfig{figure=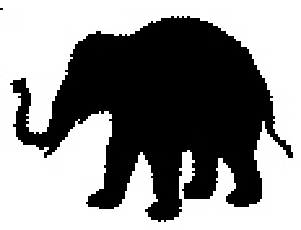, width=.55cm}
        \epsfig{figure=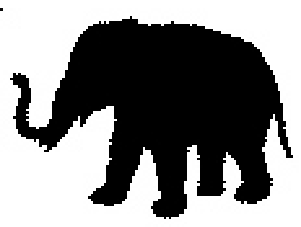, width=.55cm}
        \epsfig{figure=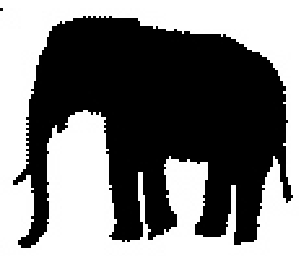, width=.55cm}
        \epsfig{figure=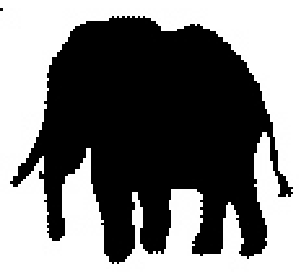, width=.55cm}
        \epsfig{figure=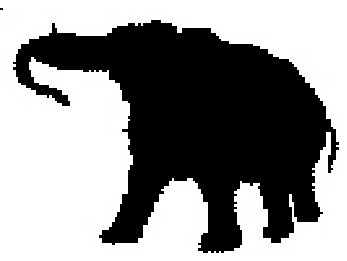, width=.55cm}
        \epsfig{figure=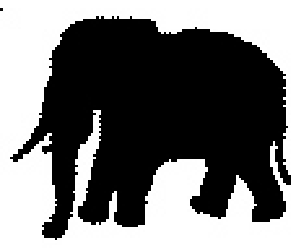, width=.55cm}
        \epsfig{figure=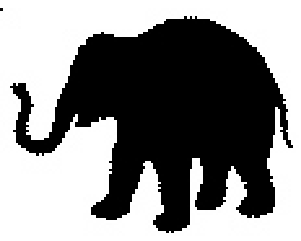, width=.55cm}
        \epsfig{figure=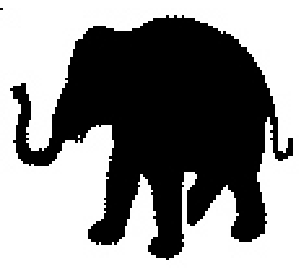, width=.55cm}
        \epsfig{figure=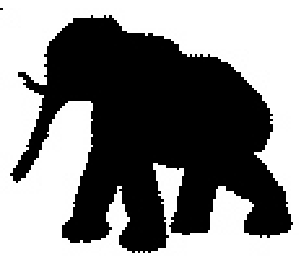, width=.55cm}
        \epsfig{figure=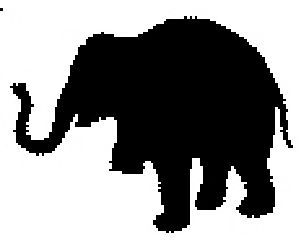, width=.55cm}}\centerline{
        \epsfig{figure=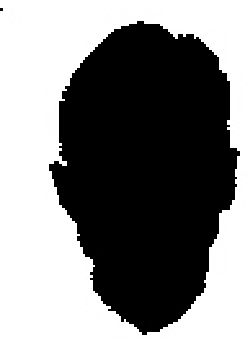, width=.55cm}
        \epsfig{figure=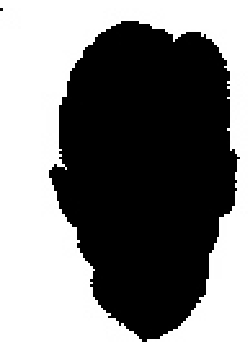, width=.55cm}
        \epsfig{figure=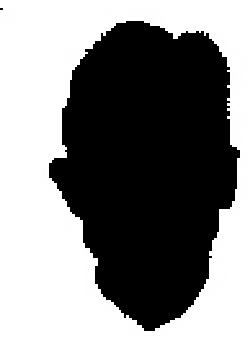, width=.55cm}
        \epsfig{figure=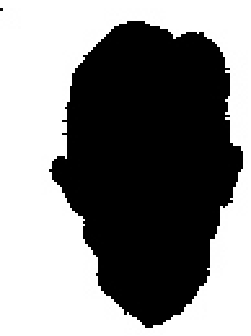, width=.55cm}
        \epsfig{figure=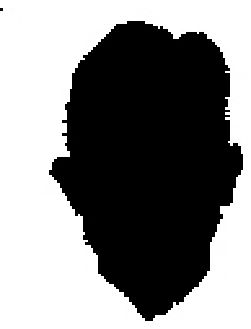, width=.55cm}
        \epsfig{figure=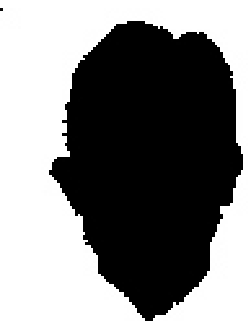, width=.55cm}
        \epsfig{figure=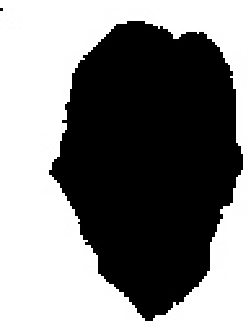, width=.55cm}
        \epsfig{figure=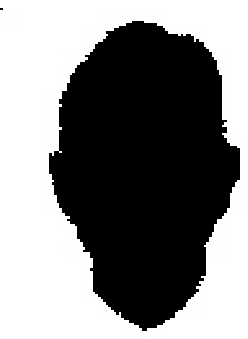, width=.55cm}
        \epsfig{figure=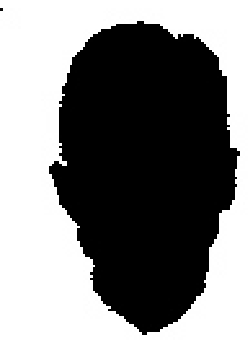, width=.55cm}
        \epsfig{figure=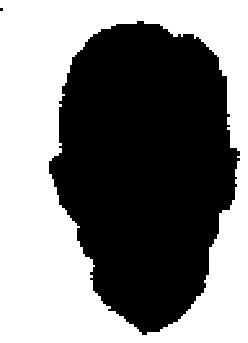, width=.55cm}
        \epsfig{figure=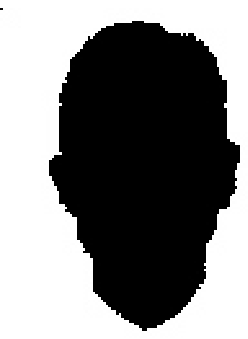, width=.55cm}
        \epsfig{figure=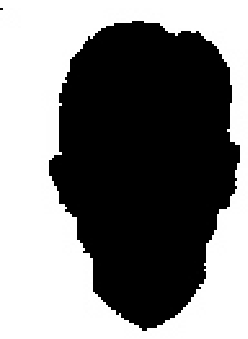, width=.55cm}
        \epsfig{figure=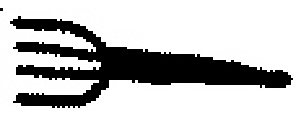, width=.55cm}
        \epsfig{figure=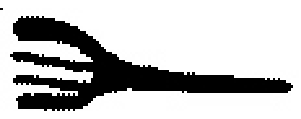, width=.55cm}
        \epsfig{figure=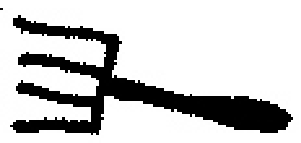, width=.55cm}
        \epsfig{figure=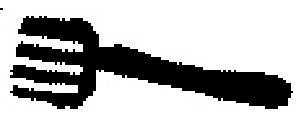, width=.55cm}
        \epsfig{figure=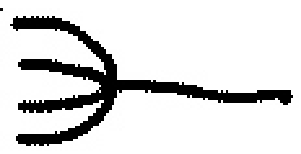, width=.55cm}
        \epsfig{figure=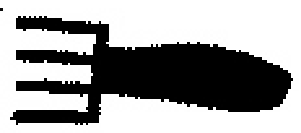, width=.55cm}
        \epsfig{figure=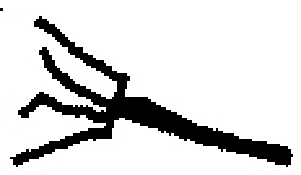, width=.55cm}
        \epsfig{figure=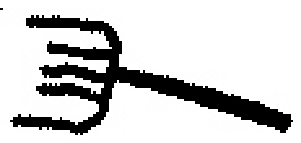, width=.55cm}
        \epsfig{figure=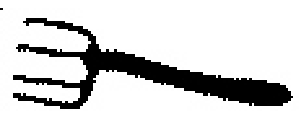, width=.55cm}
        \epsfig{figure=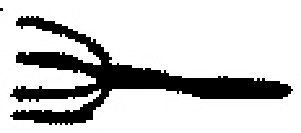, width=.55cm}
        \epsfig{figure=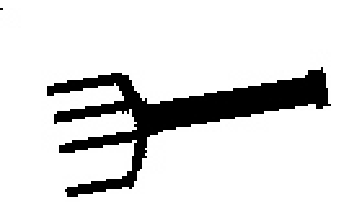, width=.55cm}
        \epsfig{figure=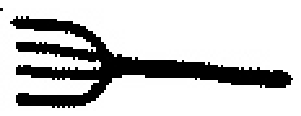, width=.55cm}}\centerline{
        \epsfig{figure=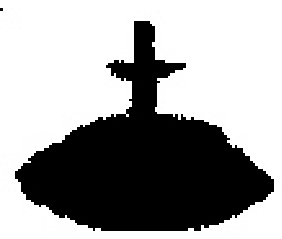, width=.55cm}
        \epsfig{figure=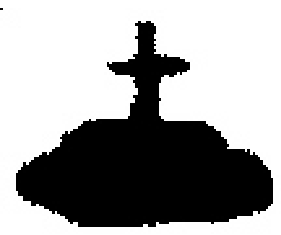, width=.55cm}
        \epsfig{figure=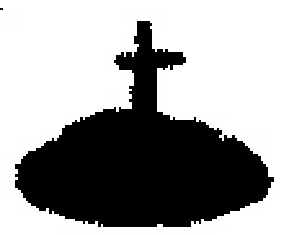, width=.55cm}
        \epsfig{figure=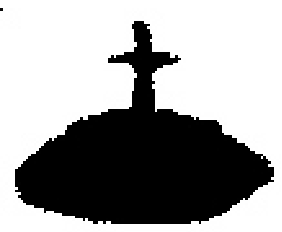, width=.55cm}
        \epsfig{figure=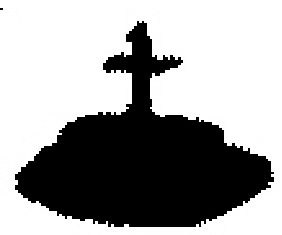, width=.55cm}
        \epsfig{figure=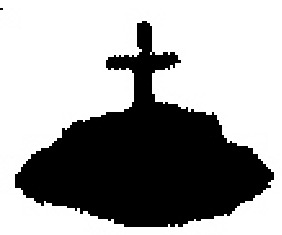, width=.55cm}
        \epsfig{figure=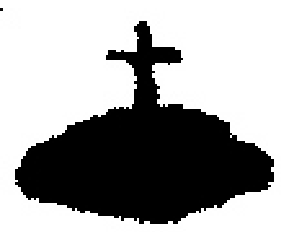, width=.55cm}
        \epsfig{figure=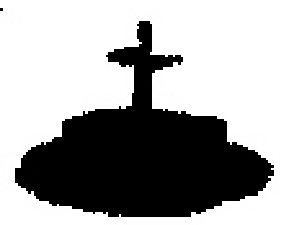, width=.55cm}
        \epsfig{figure=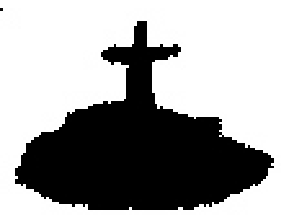, width=.55cm}
        \epsfig{figure=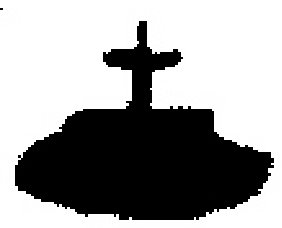, width=.55cm}
        \epsfig{figure=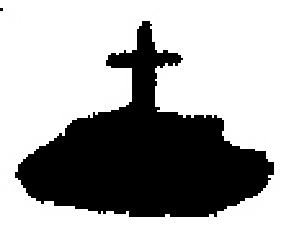, width=.55cm}
        \epsfig{figure=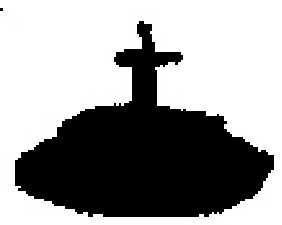, width=.55cm}
        \epsfig{figure=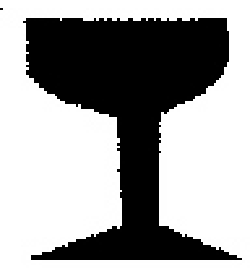, width=.55cm}
        \epsfig{figure=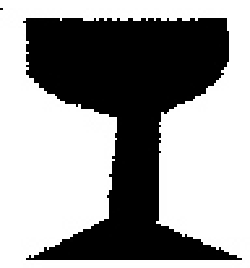, width=.55cm}
        \epsfig{figure=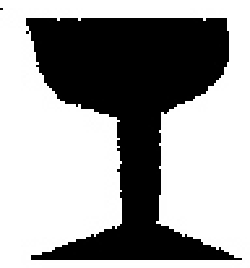, width=.55cm}
        \epsfig{figure=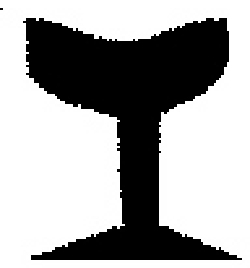, width=.55cm}
        \epsfig{figure=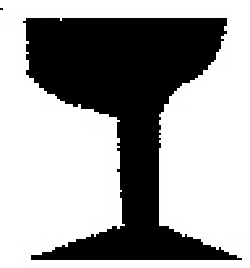, width=.55cm}
        \epsfig{figure=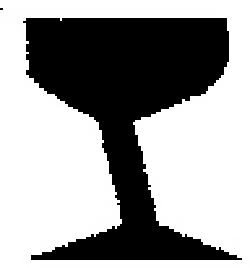, width=.55cm}
        \epsfig{figure=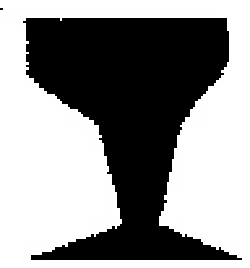, width=.55cm}
        \epsfig{figure=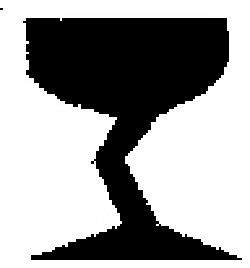, width=.55cm}
        \epsfig{figure=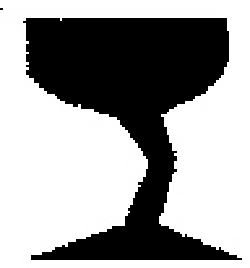, width=.55cm}
        \epsfig{figure=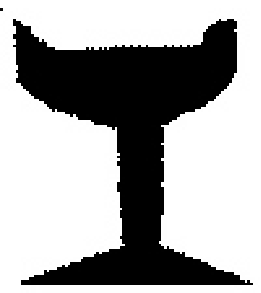, width=.55cm}
        \epsfig{figure=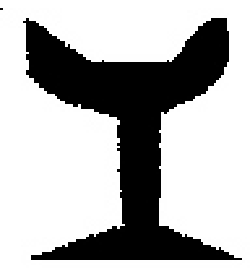, width=.55cm}
        \epsfig{figure=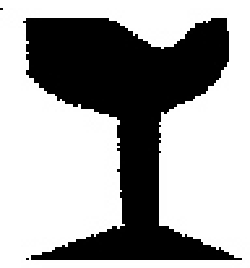, width=.55cm}}\centerline{
        \epsfig{figure=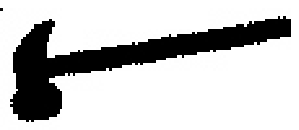, width=.55cm}
        \epsfig{figure=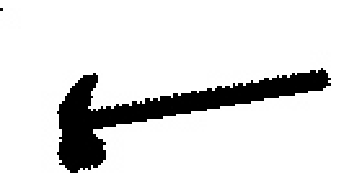, width=.55cm}
        \epsfig{figure=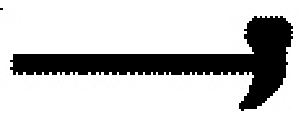, width=.55cm}
        \epsfig{figure=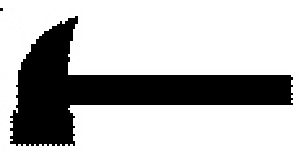, width=.55cm}
        \epsfig{figure=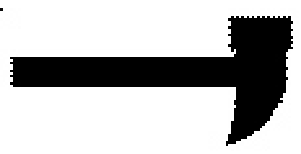, width=.55cm}
        \epsfig{figure=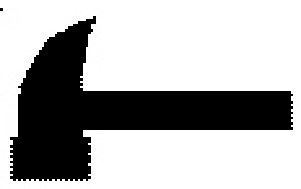, width=.55cm}
        \epsfig{figure=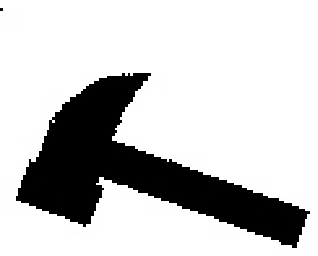, width=.55cm}
        \epsfig{figure=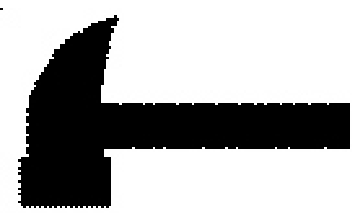, width=.55cm}
        \epsfig{figure=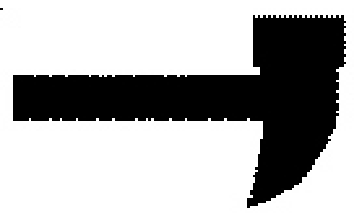, width=.55cm}
        \epsfig{figure=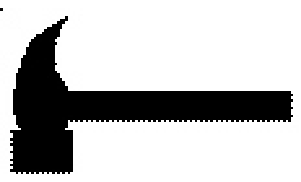, width=.55cm}
        \epsfig{figure=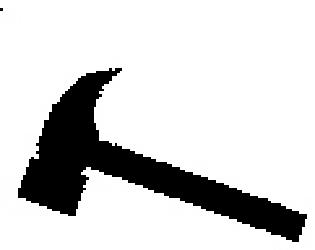, width=.55cm}
        \epsfig{figure=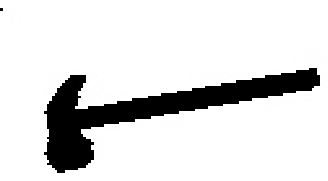, width=.55cm}
        \epsfig{figure=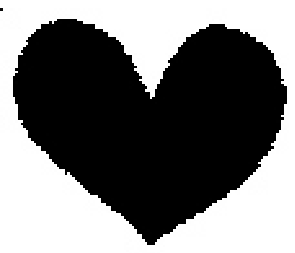, width=.55cm}
        \epsfig{figure=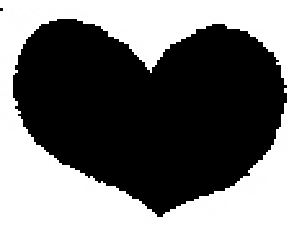, width=.55cm}
        \epsfig{figure=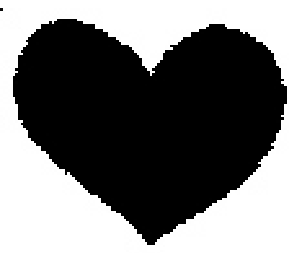, width=.55cm}
        \epsfig{figure=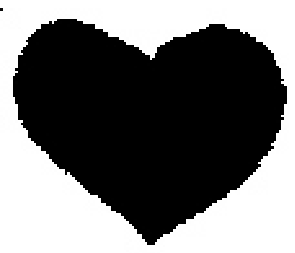, width=.55cm}
        \epsfig{figure=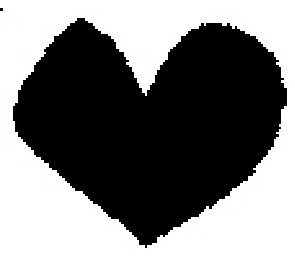, width=.55cm}
        \epsfig{figure=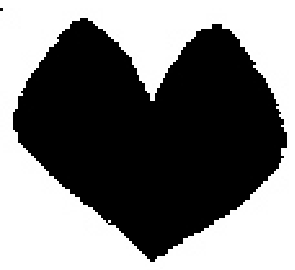, width=.55cm}
        \epsfig{figure=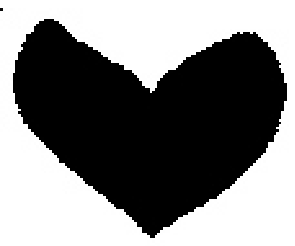, width=.55cm}
        \epsfig{figure=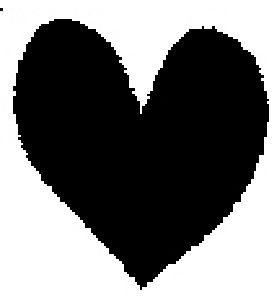, width=.55cm}
        \epsfig{figure=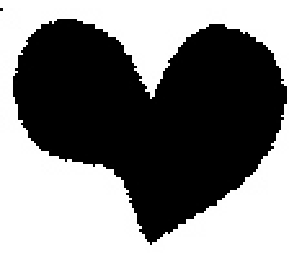, width=.55cm}
        \epsfig{figure=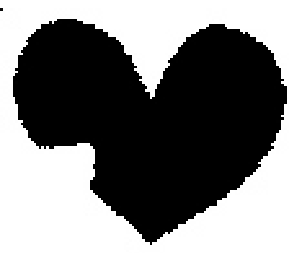, width=.55cm}
        \epsfig{figure=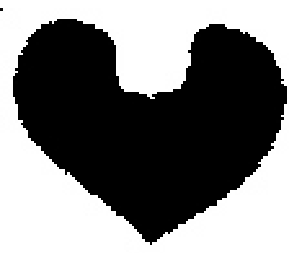, width=.55cm}
        \epsfig{figure=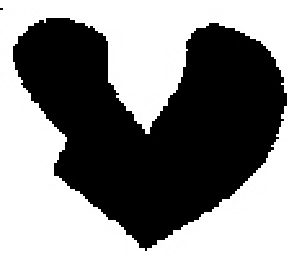, width=.55cm}}\centerline{
        \epsfig{figure=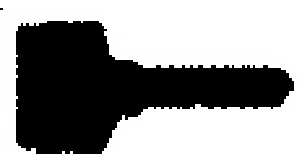, width=.55cm}
        \epsfig{figure=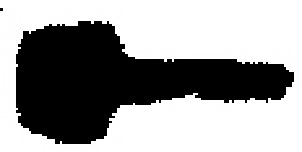, width=.55cm}
        \epsfig{figure=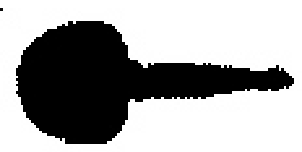, width=.55cm}
        \epsfig{figure=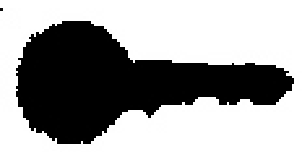, width=.55cm}
        \epsfig{figure=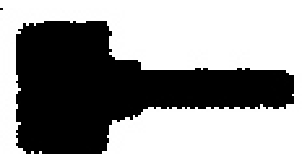, width=.55cm}
        \epsfig{figure=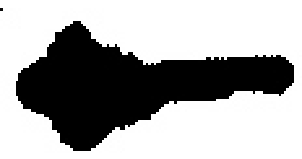, width=.55cm}
        \epsfig{figure=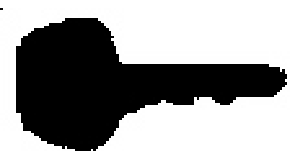, width=.55cm}
        \epsfig{figure=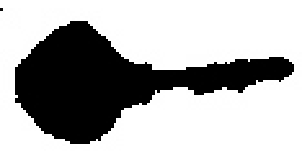, width=.55cm}
        \epsfig{figure=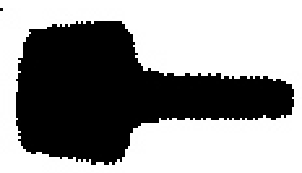, width=.55cm}
        \epsfig{figure=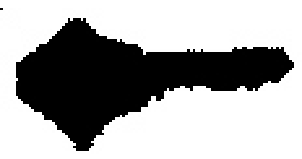, width=.55cm}
        \epsfig{figure=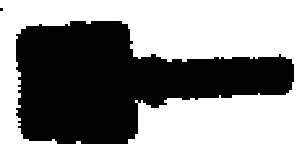, width=.55cm}
        \epsfig{figure=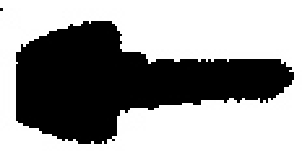, width=.55cm}
        \epsfig{figure=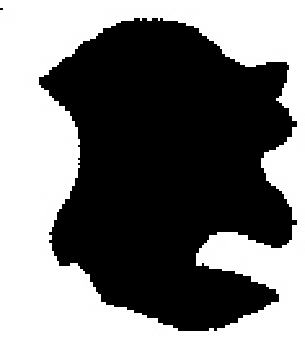, width=.55cm}
        \epsfig{figure=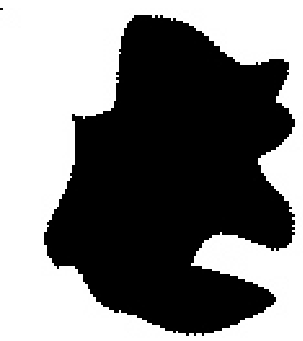, width=.55cm}
        \epsfig{figure=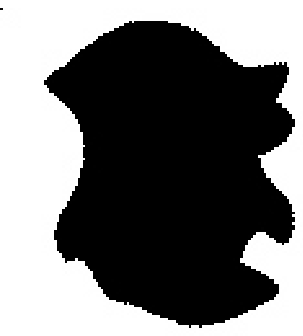, width=.55cm}
        \epsfig{figure=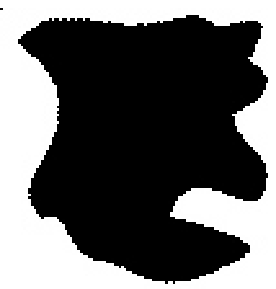, width=.55cm}
        \epsfig{figure=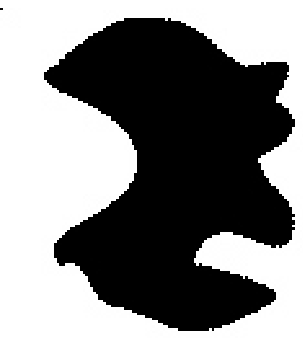, width=.55cm}
        \epsfig{figure=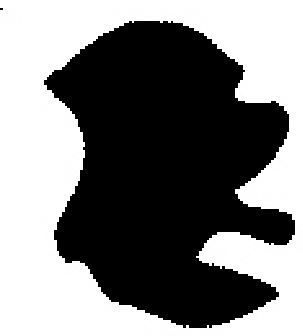, width=.55cm}
        \epsfig{figure=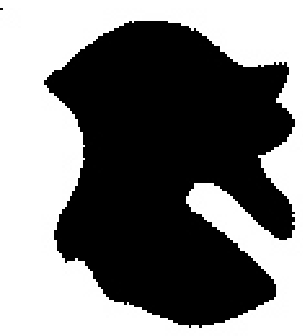, width=.55cm}
        \epsfig{figure=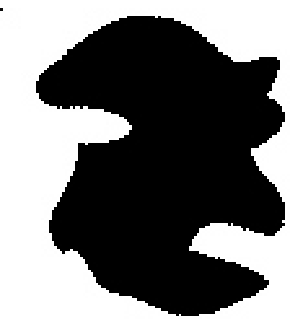, width=.55cm}
        \epsfig{figure=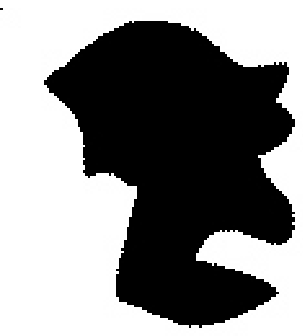, width=.55cm}
        \epsfig{figure=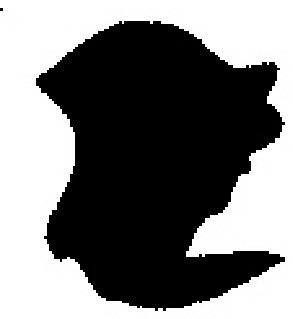, width=.55cm}
        \epsfig{figure=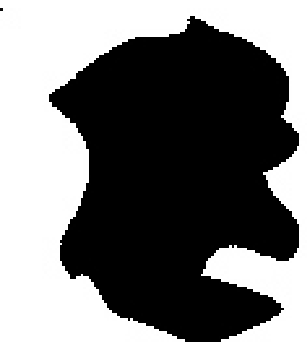, width=.55cm}
        \epsfig{figure=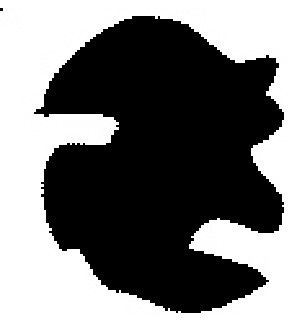, width=.55cm}}\centerline{
        \epsfig{figure=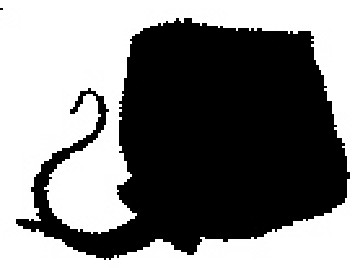, width=.55cm}
        \epsfig{figure=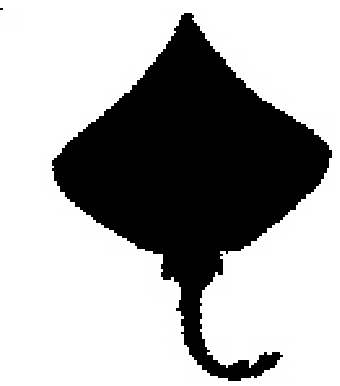, width=.55cm}
        \epsfig{figure=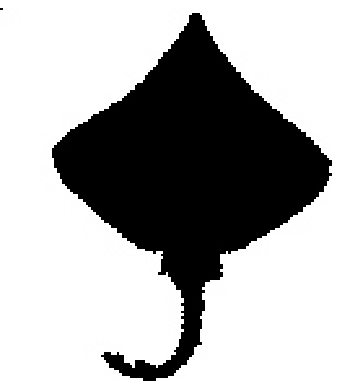, width=.55cm}
        \epsfig{figure=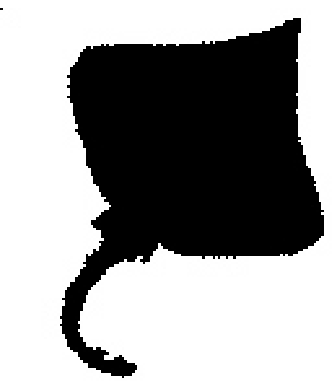, width=.55cm}
        \epsfig{figure=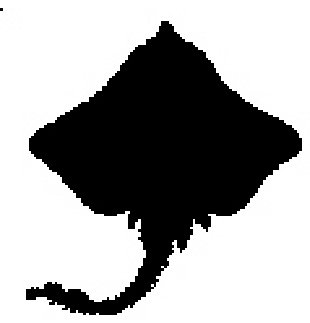, width=.55cm}
        \epsfig{figure=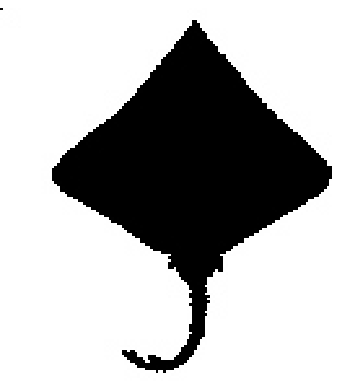, width=.55cm}
        \epsfig{figure=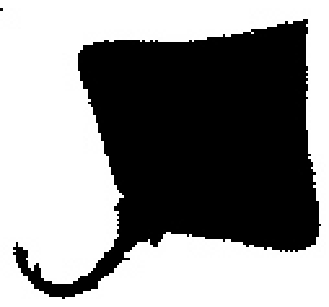, width=.55cm}
        \epsfig{figure=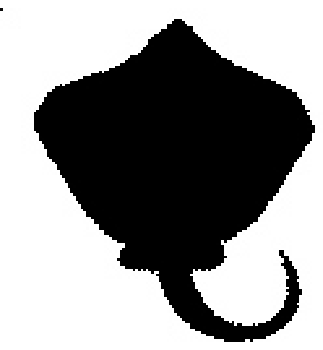, width=.55cm}
        \epsfig{figure=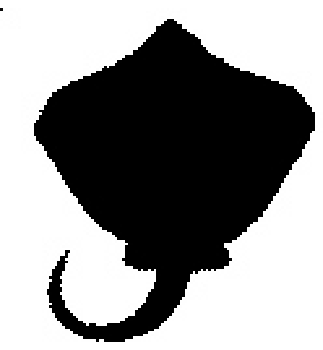, width=.55cm}
        \epsfig{figure=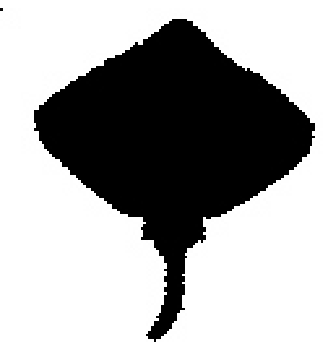, width=.55cm}
        \epsfig{figure=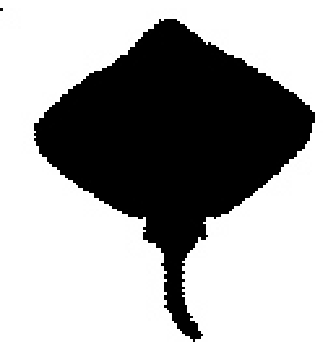, width=.55cm}
        \epsfig{figure=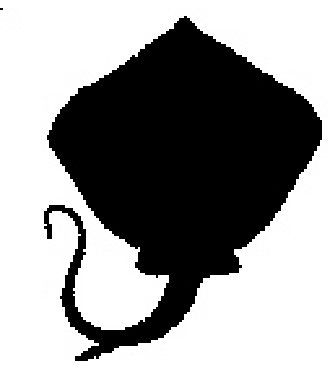, width=.55cm}
        \epsfig{figure=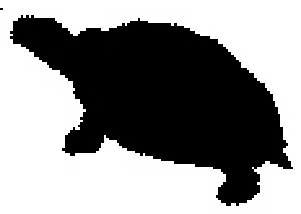, width=.55cm}
        \epsfig{figure=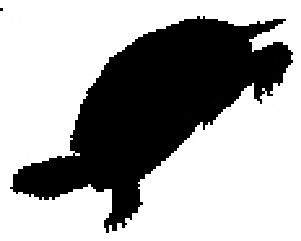, width=.55cm}
        \epsfig{figure=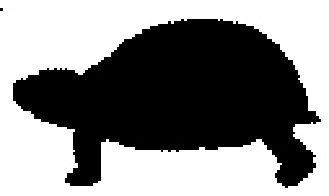, width=.55cm}
        \epsfig{figure=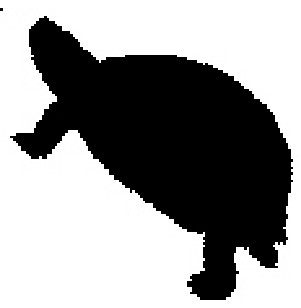, width=.55cm}
        \epsfig{figure=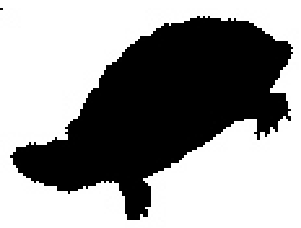, width=.55cm}
        \epsfig{figure=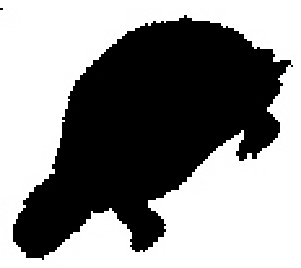, width=.55cm}
        \epsfig{figure=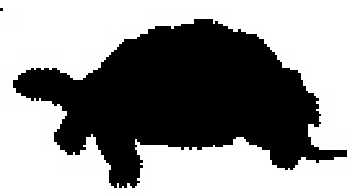, width=.55cm}
        \epsfig{figure=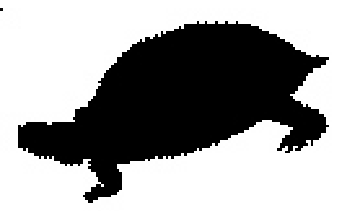, width=.55cm}
        \epsfig{figure=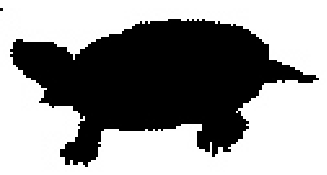, width=.55cm}
        \epsfig{figure=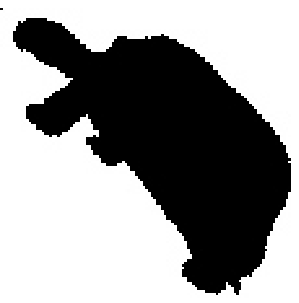, width=.55cm}
        \epsfig{figure=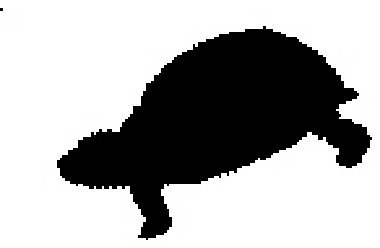, width=.55cm}
        \epsfig{figure=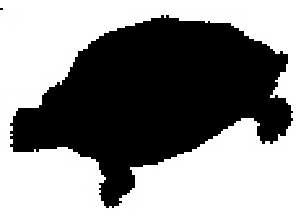, width=.55cm}}
    \vspace*{-.5cm}\caption{The $216$-shape subset~\cite{sebastian04} of the MPEG-7 shape
    database~\cite{latecki00}, containing $12$ shape classes, each with $18$
    shapes.\label{fig:216Shapes}}
    \end{figure}

\begin{figure}[hbt]
  \centerline{\epsfig{figure=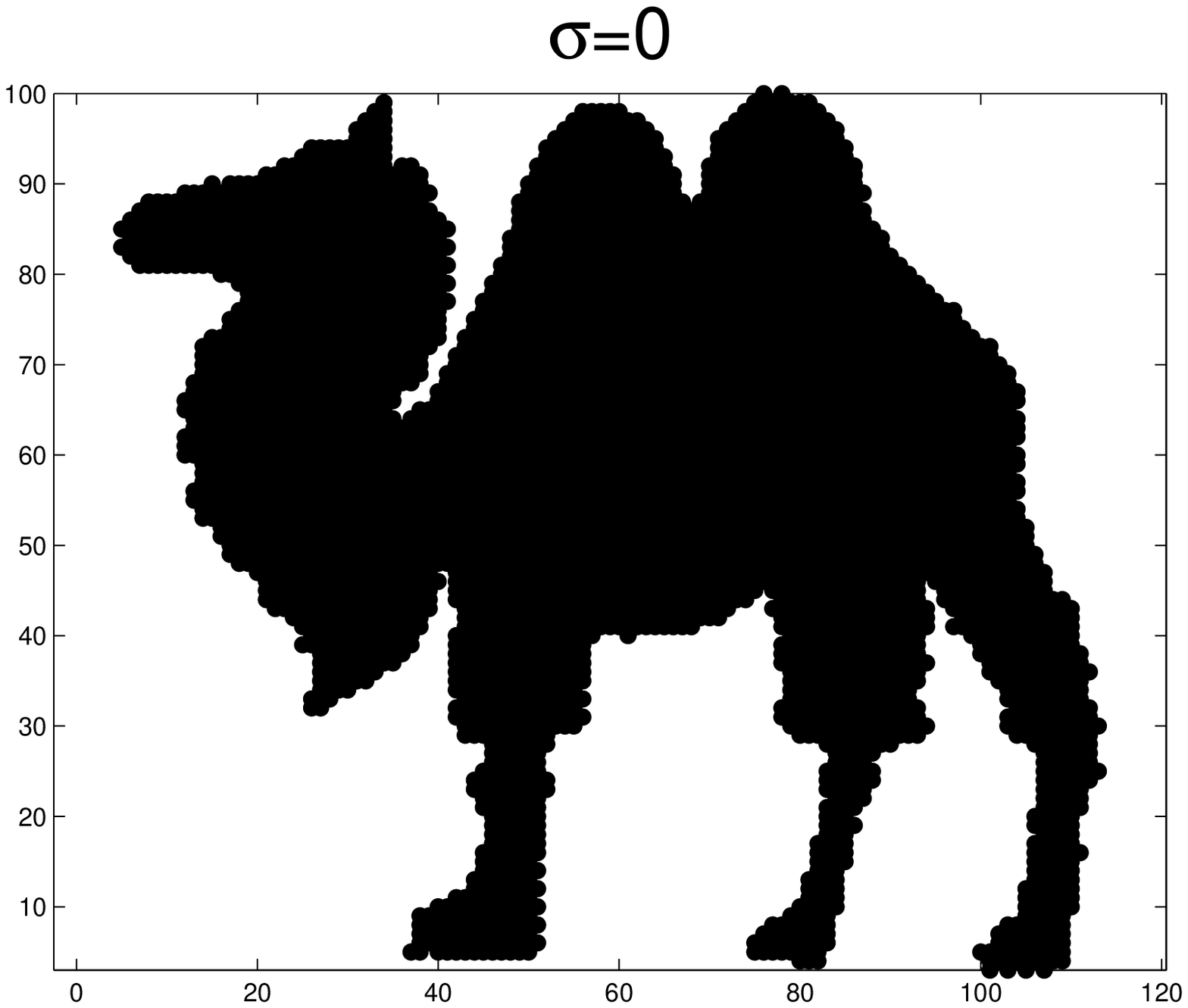,width=4cm}
  \hspace*{.25cm} \epsfig{figure=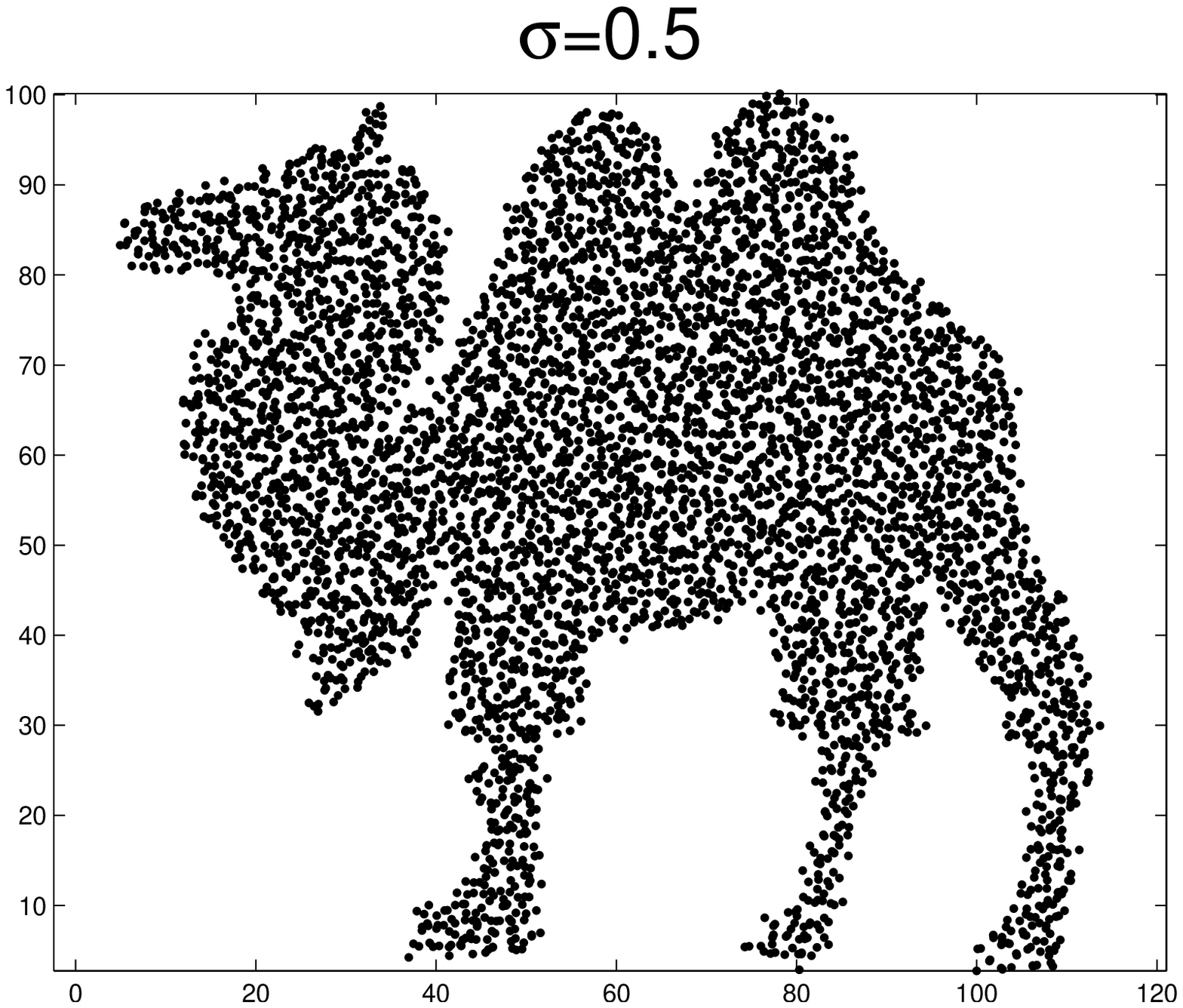,width=4cm}\hspace*{.25cm}
   \epsfig{figure=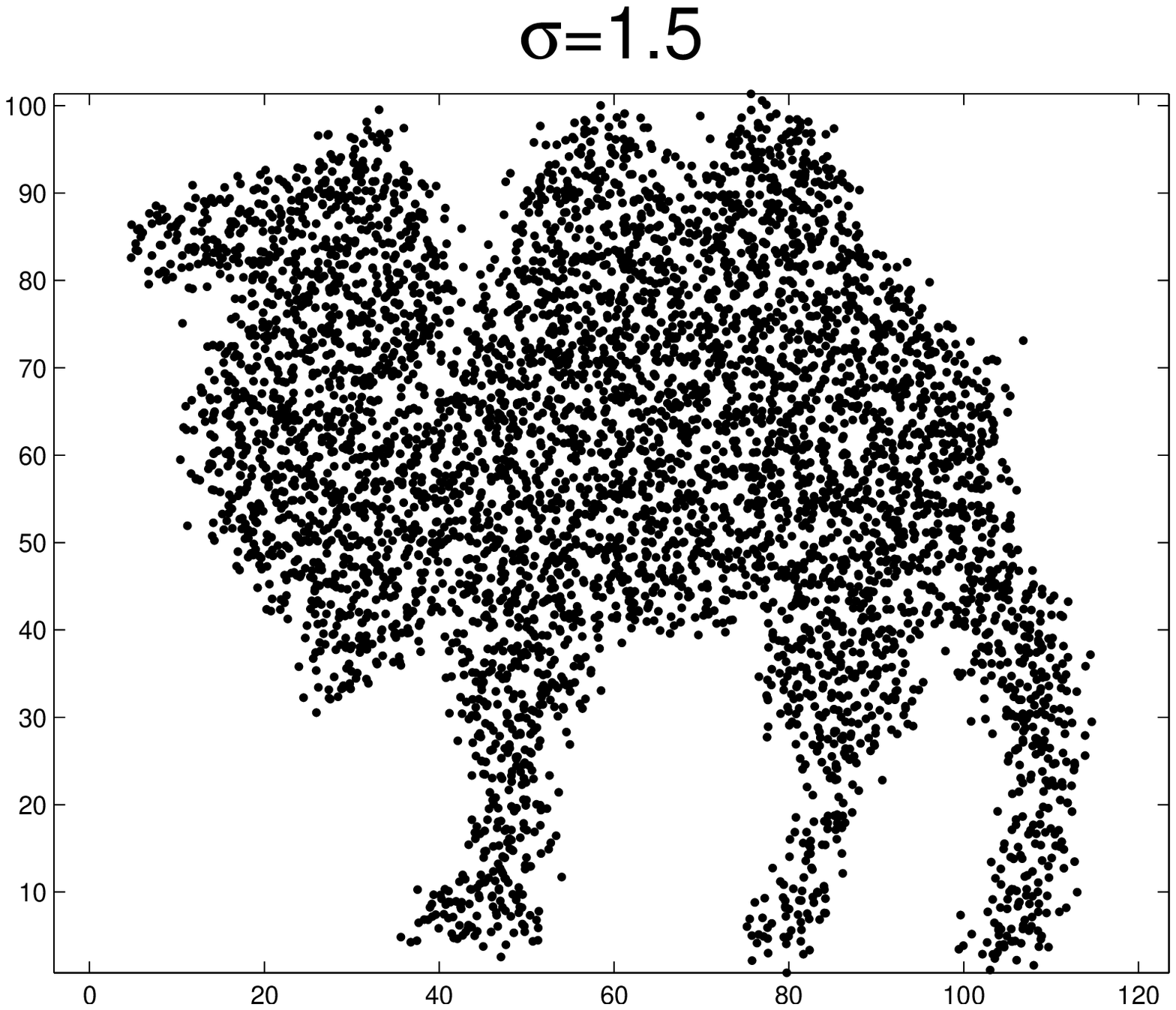,width=4cm}}
\vspace*{-.5cm}\caption{Illustration of the noise levels used in the
MPEG-7 shape database illustrated in Fig.~\ref{fig:216Shapes}.
    \label{fig:216Shapes_noise_levels}}
    \end{figure}

Although not the focus of this paper, we now illustrate that the
ANSIG representation can also be used when dealing with non-rigid
distortions, if more than one prototype per class is included in the
database. We use the subset of the MPEG-7 shape
database~\cite{latecki00,sebastian04} show in
Fig.~\ref{fig:216Shapes}. It has 18 classes, each containing 12
shapes that, although perceptually similar, are not geometrically
equal. We increase shape variability by creating test shapes that
are noisy versions of the ones in Fig.~\ref{fig:216Shapes} (the
noise levels are illustrated with an example in
Fig.~\ref{fig:216Shapes_noise_levels}).

We stored the ANSIGs of the noiseless shapes and then classified a
set of 43200 test shapes (200 for each sample shape in
Fig.~\ref{fig:216Shapes}), for each of the two noise levels
illustrated in Fig.~\ref{fig:216Shapes_noise_levels}, using a simple
1-NN classifier, {\it i.e.}, selecting the class with most similar
ANSIG. We got 100\% correct classifications for both noise levels.
See~\cite{ANSIG1} for more details on this experiment
and~\cite{ANSIG2} for other experiment that demonstrates robustness
to noise with a clip-art shape database.

\subsection{Real images}
\label{subsec:ATR}

We now describe an experiment with real images of trademarks. We
obtained from the internet a set of images of logos that are well
described by its shape content.
These images range from 80$\times$80 to 500$\times$500 pixels. To
obtain shape vectors from intensity images, {\it i.e.}, sets of
points or landmarks that describe the shape content, there is a
panoply of low- and mid-level processing methods that can be used,
ranging from the simple image intensity threshold to sophisticated
segmentation procedures. In this experiment, we just processed the
images with the Canny edge detector~\cite{canny86} and then stored
the ANSIGs of the corresponding edge maps in a database.

\begin{figure}[hbt]
  \centerline{\epsfig{figure=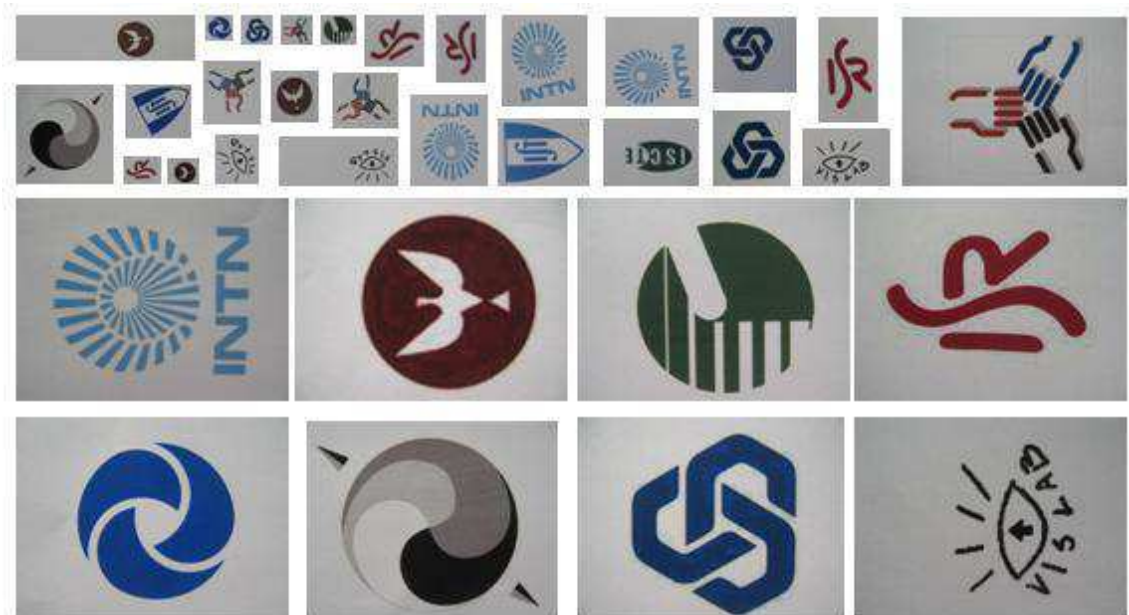}}
\vspace*{-.5cm}\caption{Examples of trademark images to be
classified.}
    \label{fig:real_logos_fotos}
    \end{figure}

To test the performance of our shape recognition scheme with
challenging test images, we printed the trademark images and
photographed their paper version with a low quality digital camera,
with different paper-camera positions, orientations, and distances.
This way we obtained a set of 88 test images, where the candidate
logos appear at different sizes and positions, see some examples in
Fig.~\ref{fig:real_logos_fotos}.

After running the Canny edge detector~\cite{canny86} on the test
images, we performed the shape-based classification in the same way
as above, {\it i.e.}, by just selecting the database entry that was
more similar to each test shape ANSIG. The results are summarized in
Table~\ref{tab:class_results_ATR2}, in the form of a confusion
matrix. We see that the generality of the test images were correctly
classified. Exceptions are: one of the photographs of the fifth logo
($12.5\%$); and several photographs of the tenth one ($50\%$,
$37.5\%$). In the sequel, we discuss these miss-classifications.

    \begin{table}[hbt]
        \caption{Automatic trademark retrieval experiment: confusion matrix.}\vspace*{-.5cm}
        \label{tab:class_results_ATR2}
        \centering
        \begin{tabular}{|c|c|c|c|c|c|c|c|c|c|c|c|}
        \hline
        &\multicolumn{11}{c|}{$Database$}\\
        \cline{2-12}
        $Photos$&
        \epsfig{figure=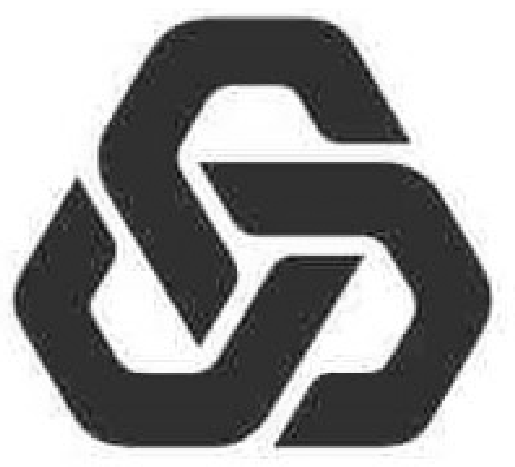, height=.4cm}&
        \epsfig{figure=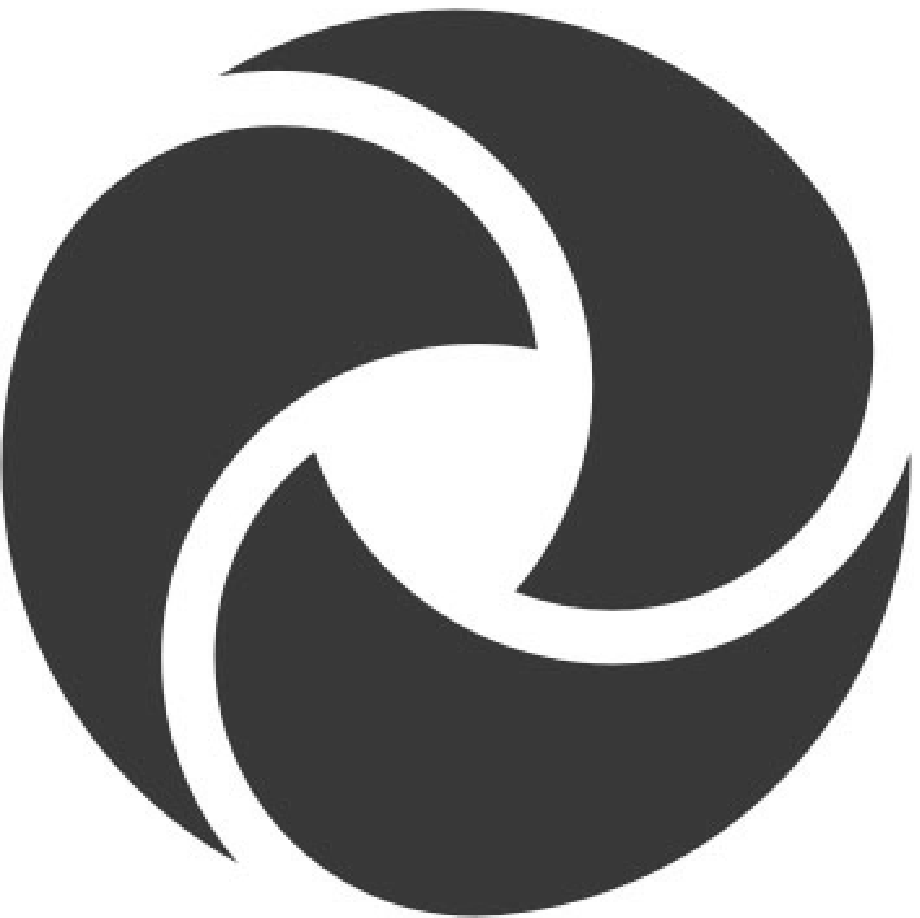, height=.4cm}&
        \epsfig{figure=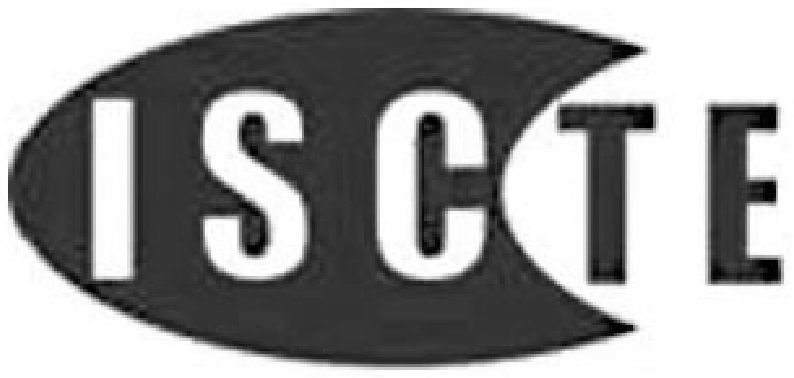, height=.35cm}&
        \epsfig{figure=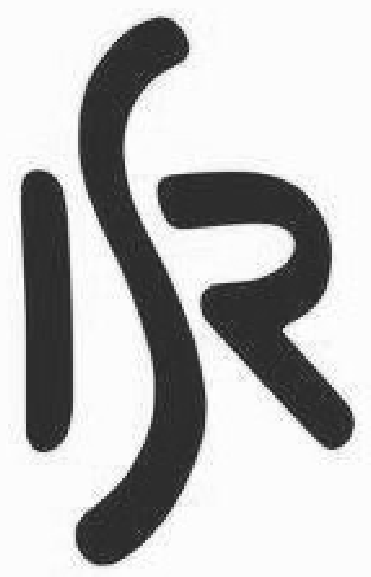, height=.4cm}&
        \epsfig{figure=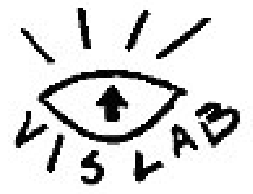, height=.4cm}&
        \epsfig{figure=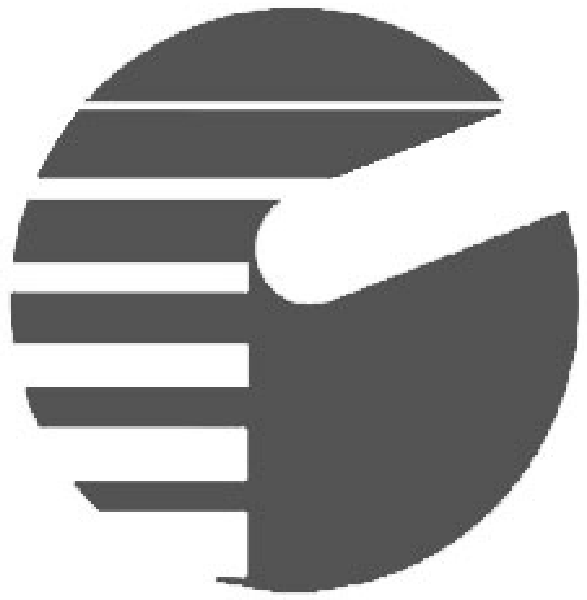, height=.4cm}&
        \epsfig{figure=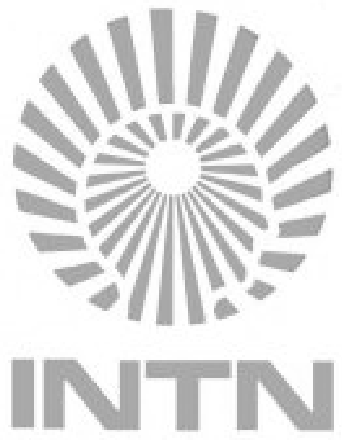, height=.4cm}&
        \epsfig{figure=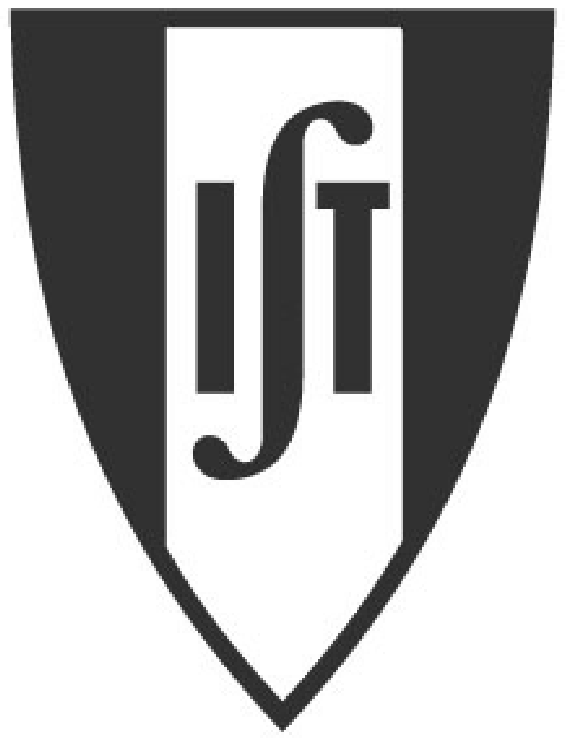, height=.4cm}&
        \epsfig{figure=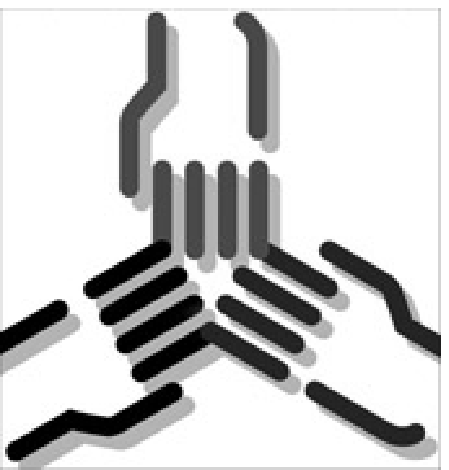, height=.4cm}&
        \epsfig{figure=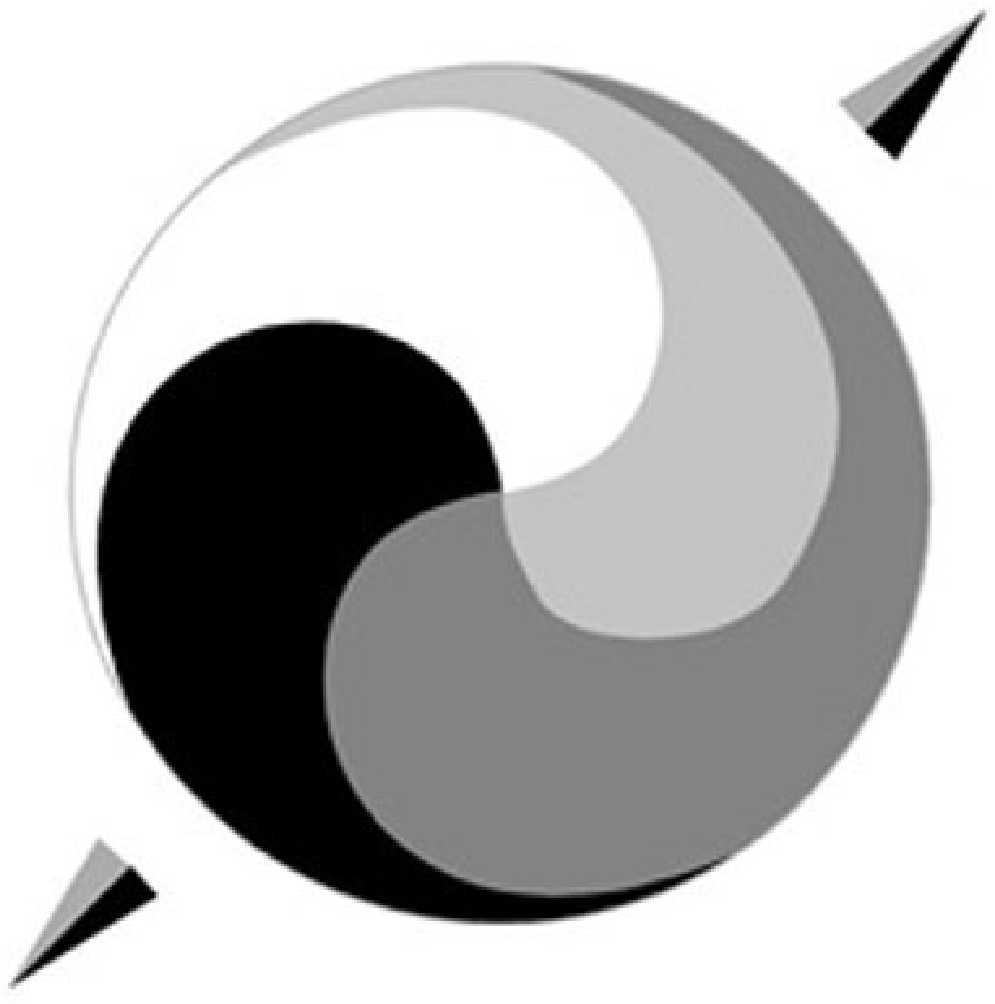, height=.4cm}&
        \epsfig{figure=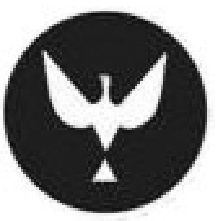, height=.4cm}\\
        \hline
        \epsfig{figure=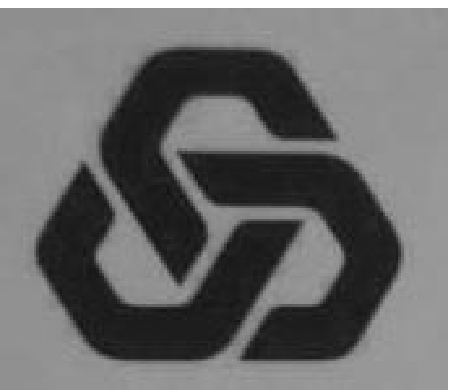, height=.7cm}&100\%&0\%&0\%&0\%&0\%&0\%&0\%&0\%&0\%&0\%&0\%\\
        \epsfig{figure=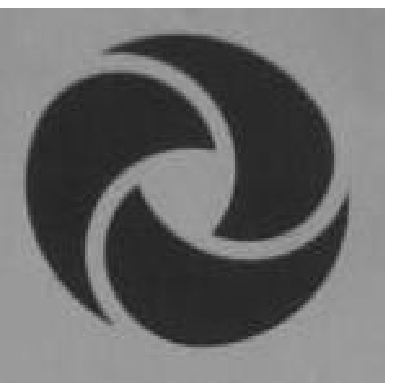, height=.7cm}&0\%&100\%&0\%&0\%&0\%&0\%&0\%&0\%&0\%&0\%&0\%\\
        \epsfig{figure=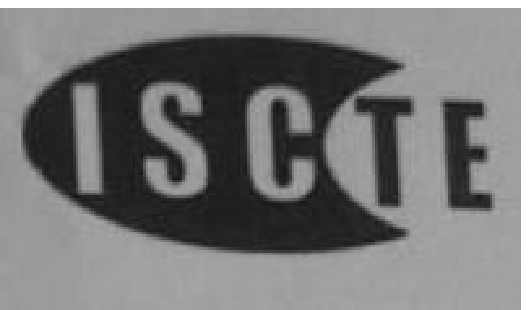, height=.5cm}&0\%&0\%&100\%&0\%&0\%&0\%&0\%&0\%&0\%&0\%&0\%\\
        \epsfig{figure=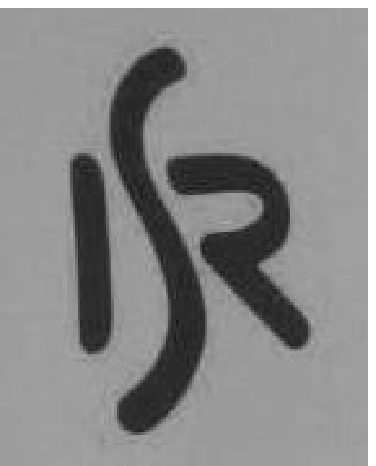, height=.7cm}&0\%&0\%&0\%&100\%&0\%&0\%&0\%&0\%&0\%&0\%&0\%\\
        \epsfig{figure=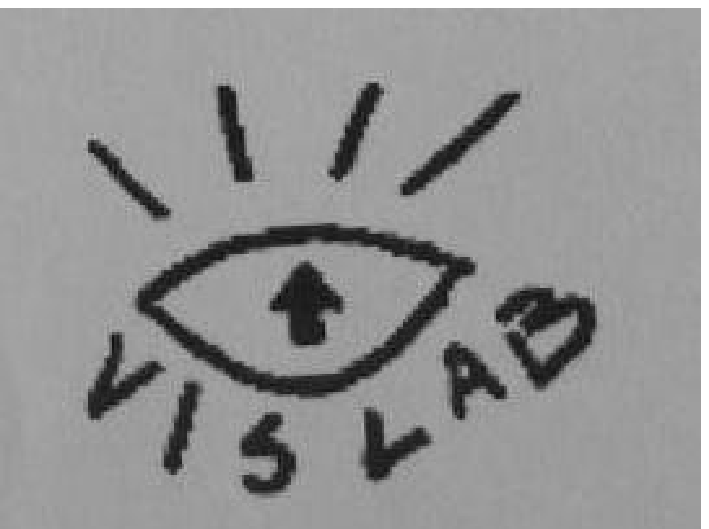, height=.6cm} &0\%&0\%&0\%&0\%&87.5\%&0\%&12.5\%&0\%&0\%&0\%&0\%\\
        \epsfig{figure=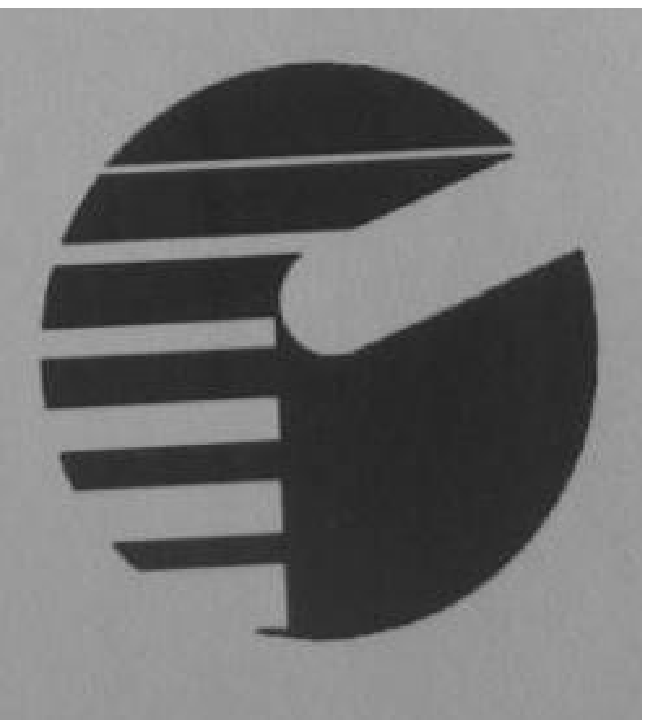, height=.7cm}&0\%&0\%&0\%&0\%&0\%&100\%&0\%&0\%&0\%&0\%&0\%\\
        \epsfig{figure=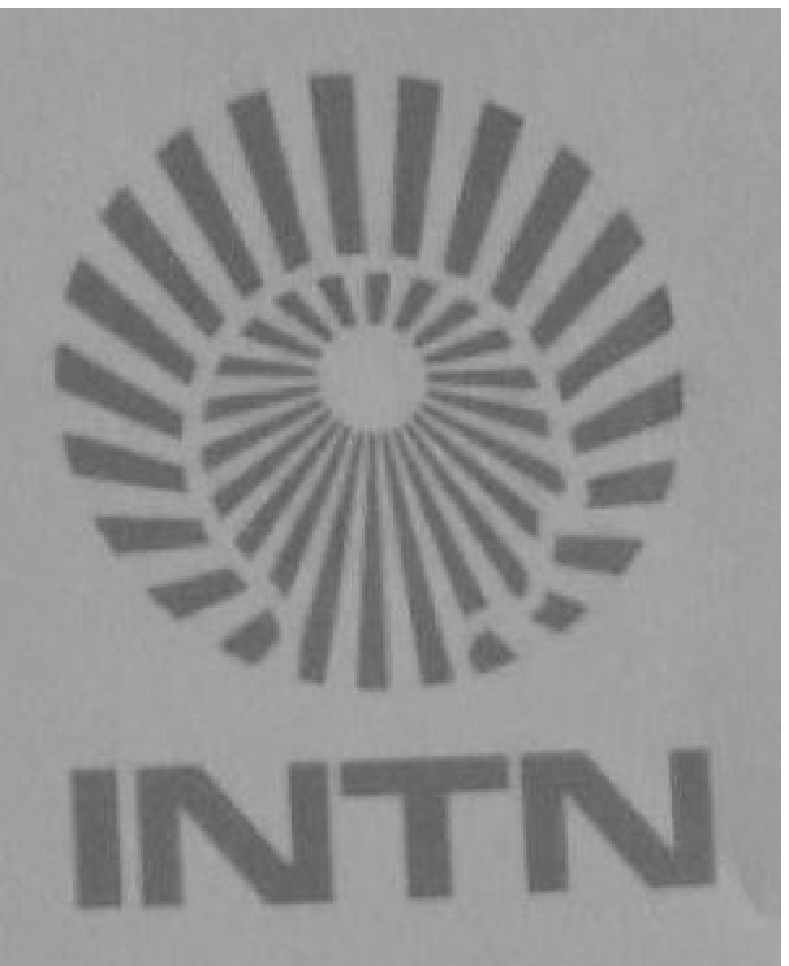, height=.7cm}&0\%&0\%&0\%&0\%&0\%&0\%&100\%&0\%&0\%&0\%&0\%\\
        \epsfig{figure=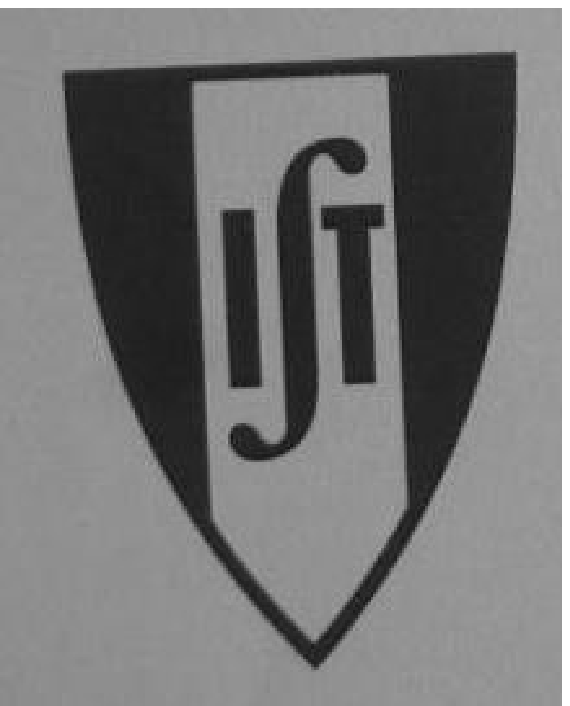, height=.7cm}&0\%&0\%&0\%&0\%&0\%&0\%&0\%&100\%&0\%&0\%&0\%\\
        \epsfig{figure=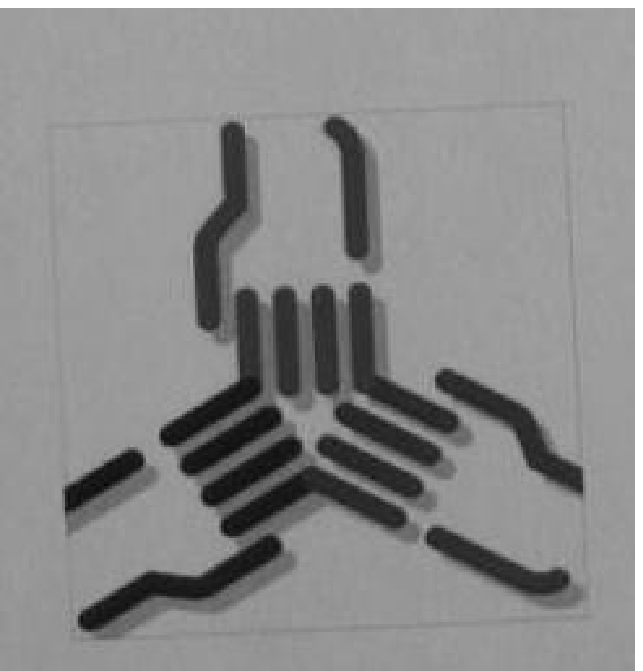, height=.7cm}&0\%&0\%&0\%&0\%&0\%&0\%&0\%&0\%&100\%&0\%&0\%\\
        \epsfig{figure=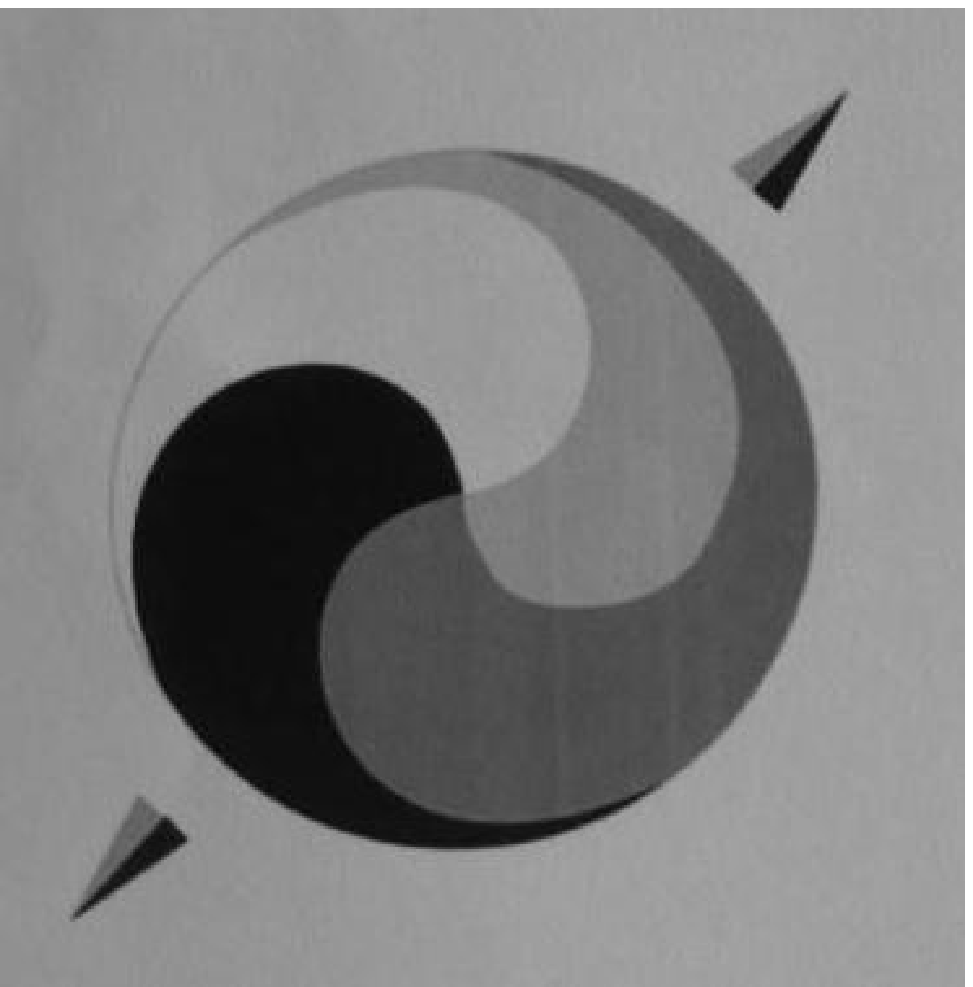, height=.7cm}&0\%&0\%&50\%&37.5\%&0\%&0\%&0\%&0\%&0\%&12.5\%&0\%\\
        \epsfig{figure=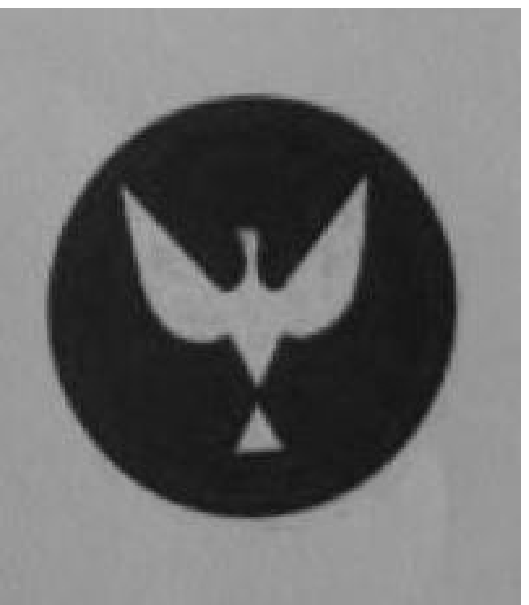, height=.7cm}&0\%&0\%&0\%&0\%&0\%&0\%&0\%&0\%&0\%&0\%&100\%\\
        \hline
        \end{tabular}
    \end{table}

\subsection{Sensitivity to model violations}
\label{subsec:SMM}

We now discuss how the behavior of the ANSIG representation is
affected by model violations that may arise when dealing with shapes
obtained from real images. In particular, we address two kinds of
model violation: failures in edge detection and perspective
distortions.

\begin{figure}[hbt]
\centerline{\epsfig{figure=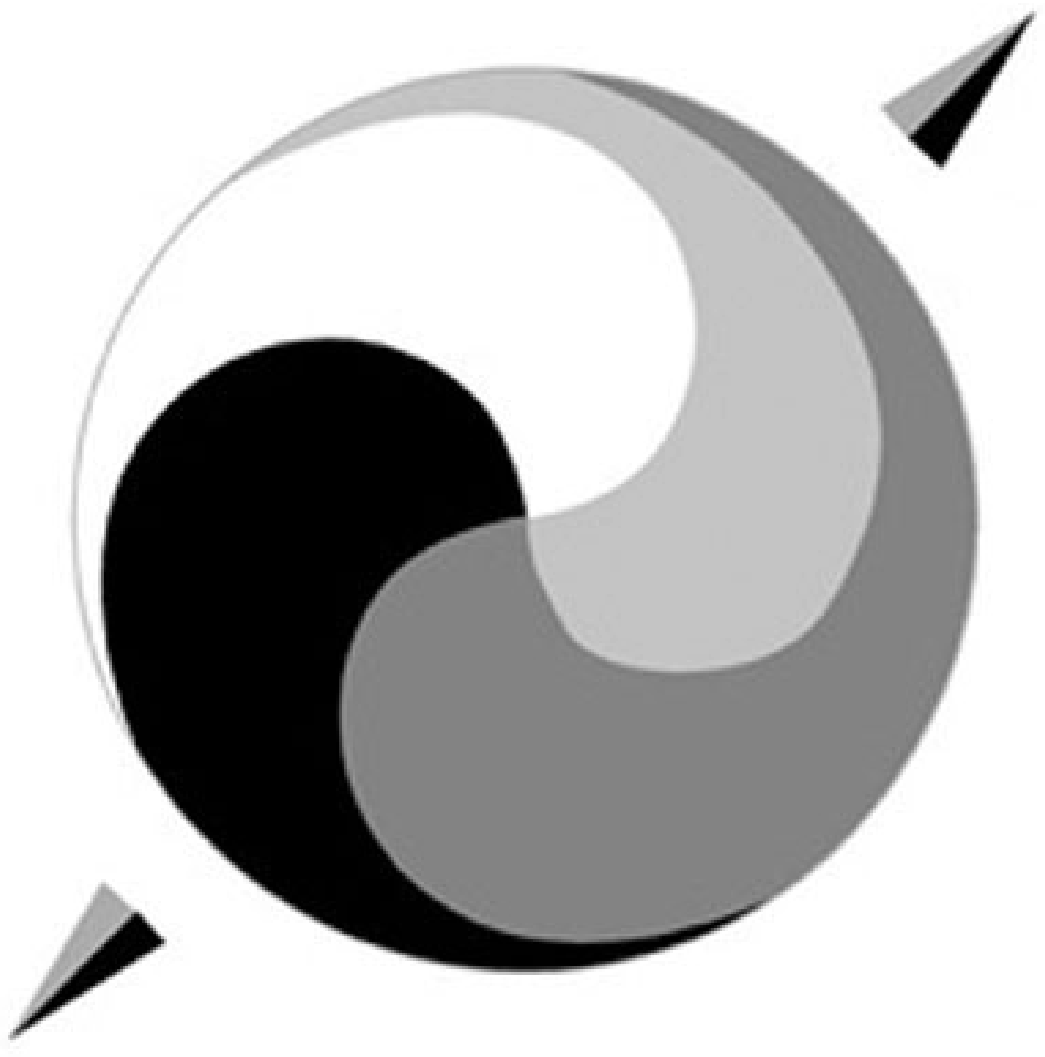,width=4.3cm}
\hspace*{0.5cm}
\epsfig{figure=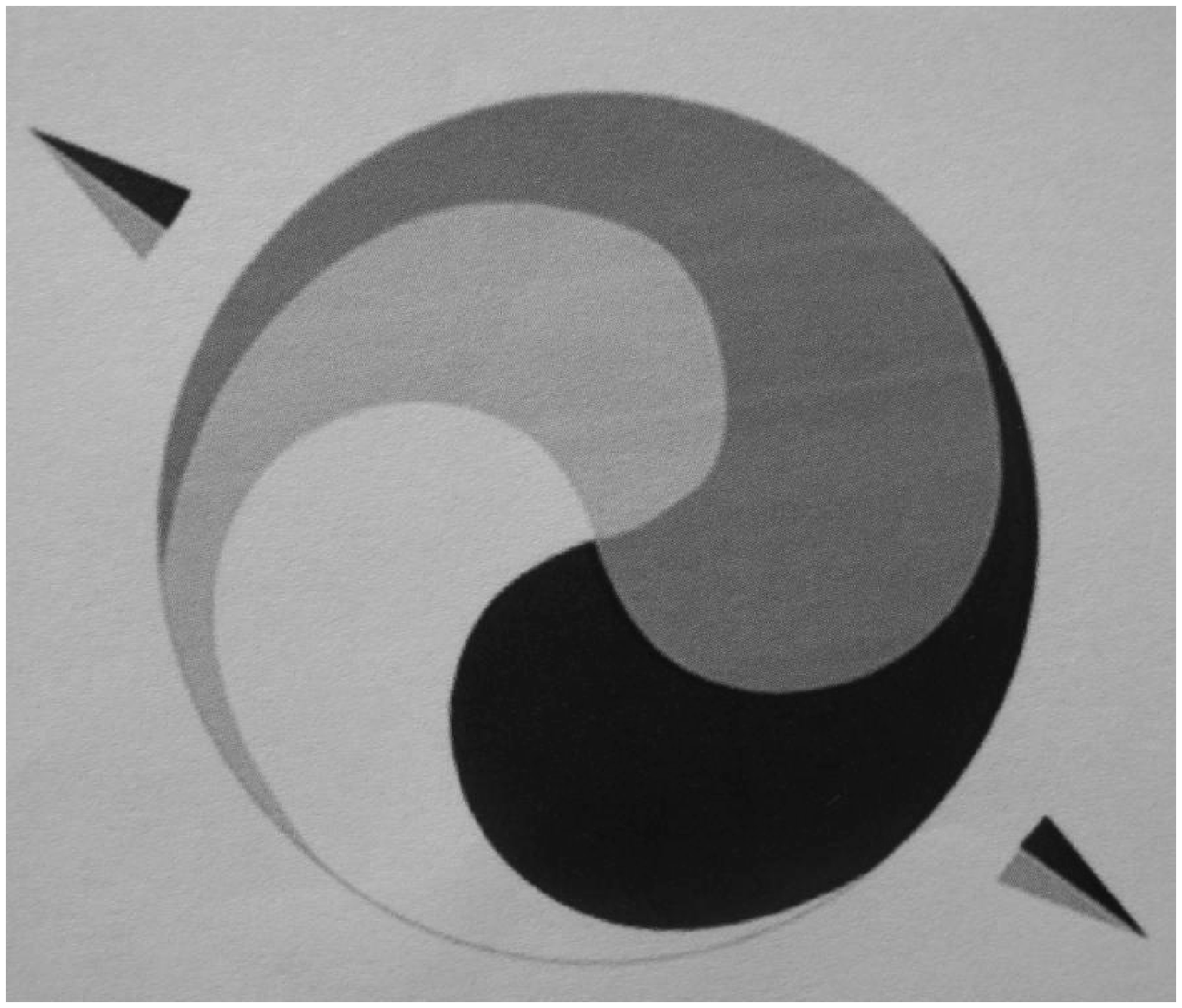,width=4.2cm}
\hspace*{0.5cm}
\epsfig{figure=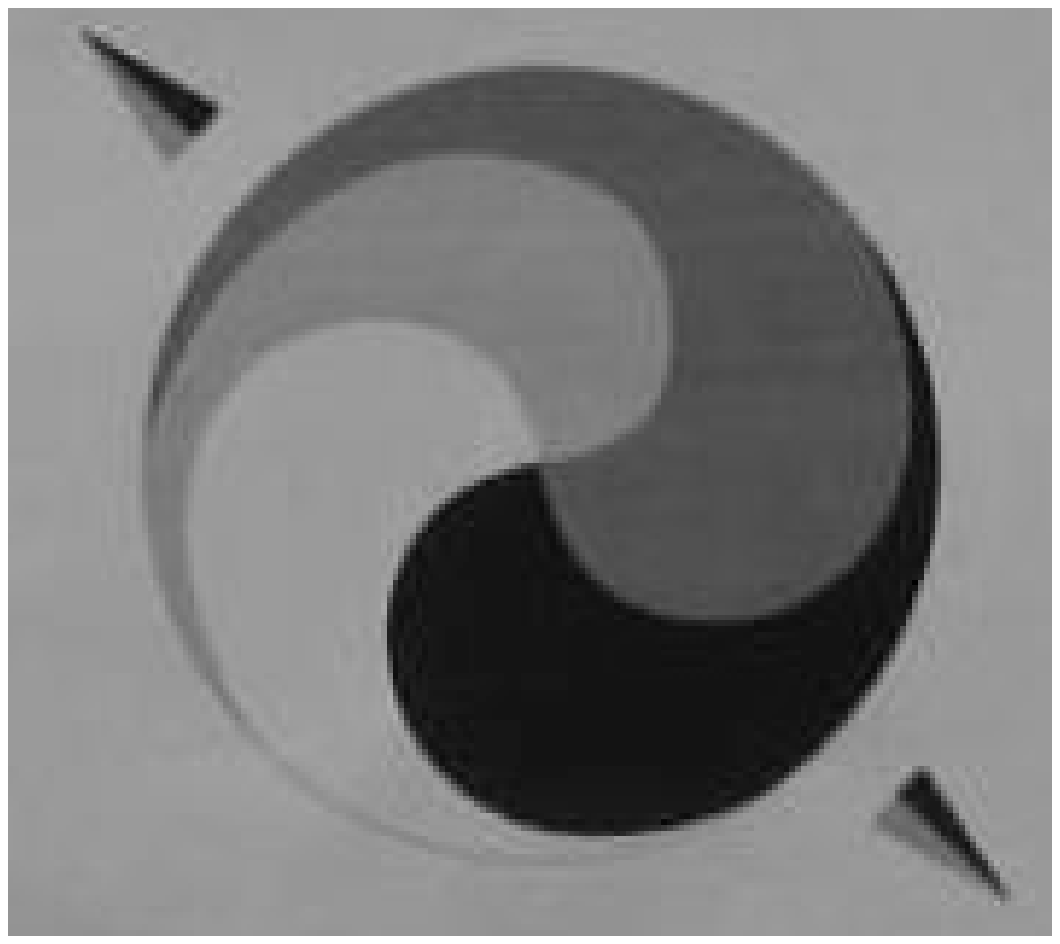,width=4.2cm}}
\centerline{\epsfig{figure=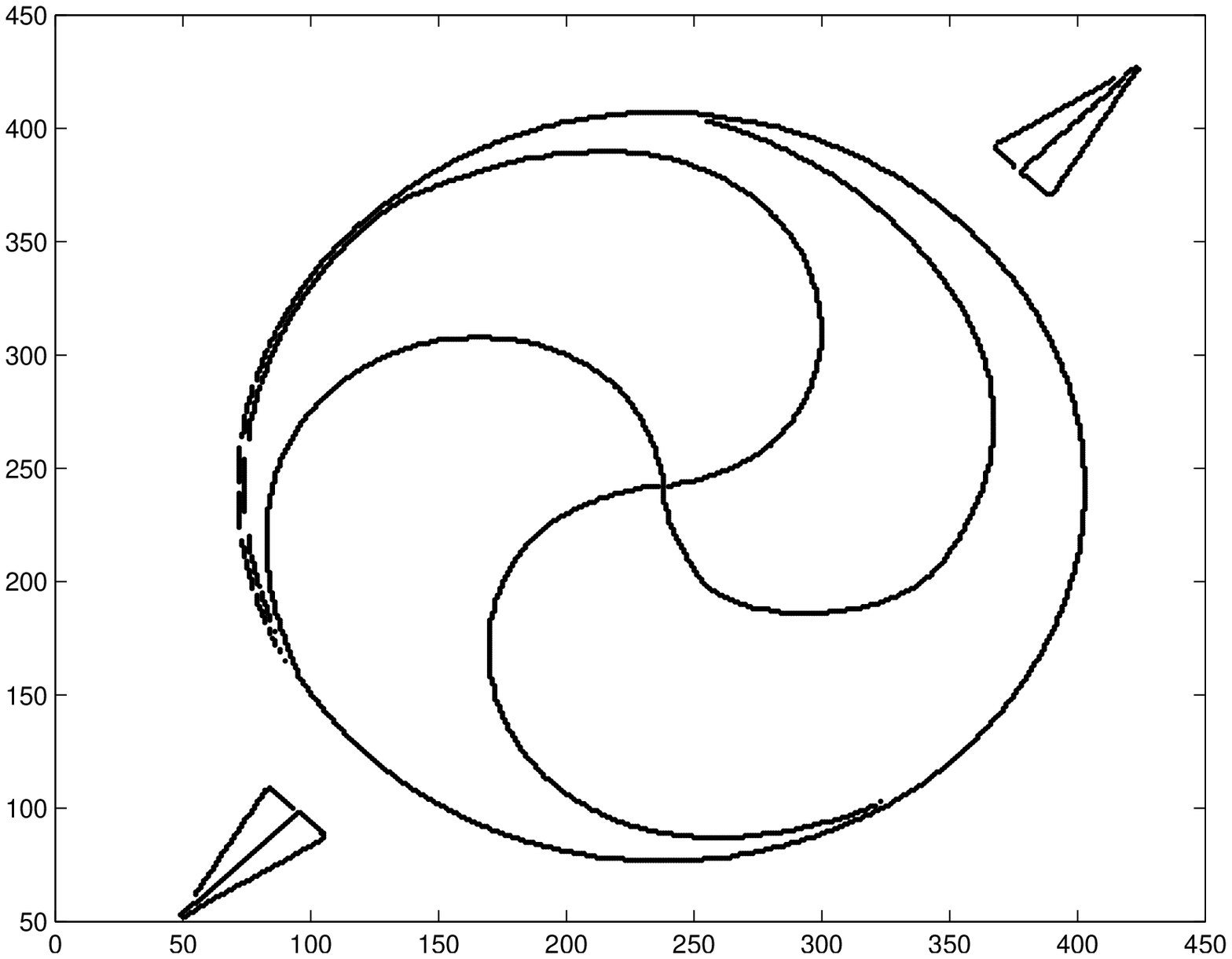,width=4.7cm}
\hspace*{0.2cm}
\epsfig{figure=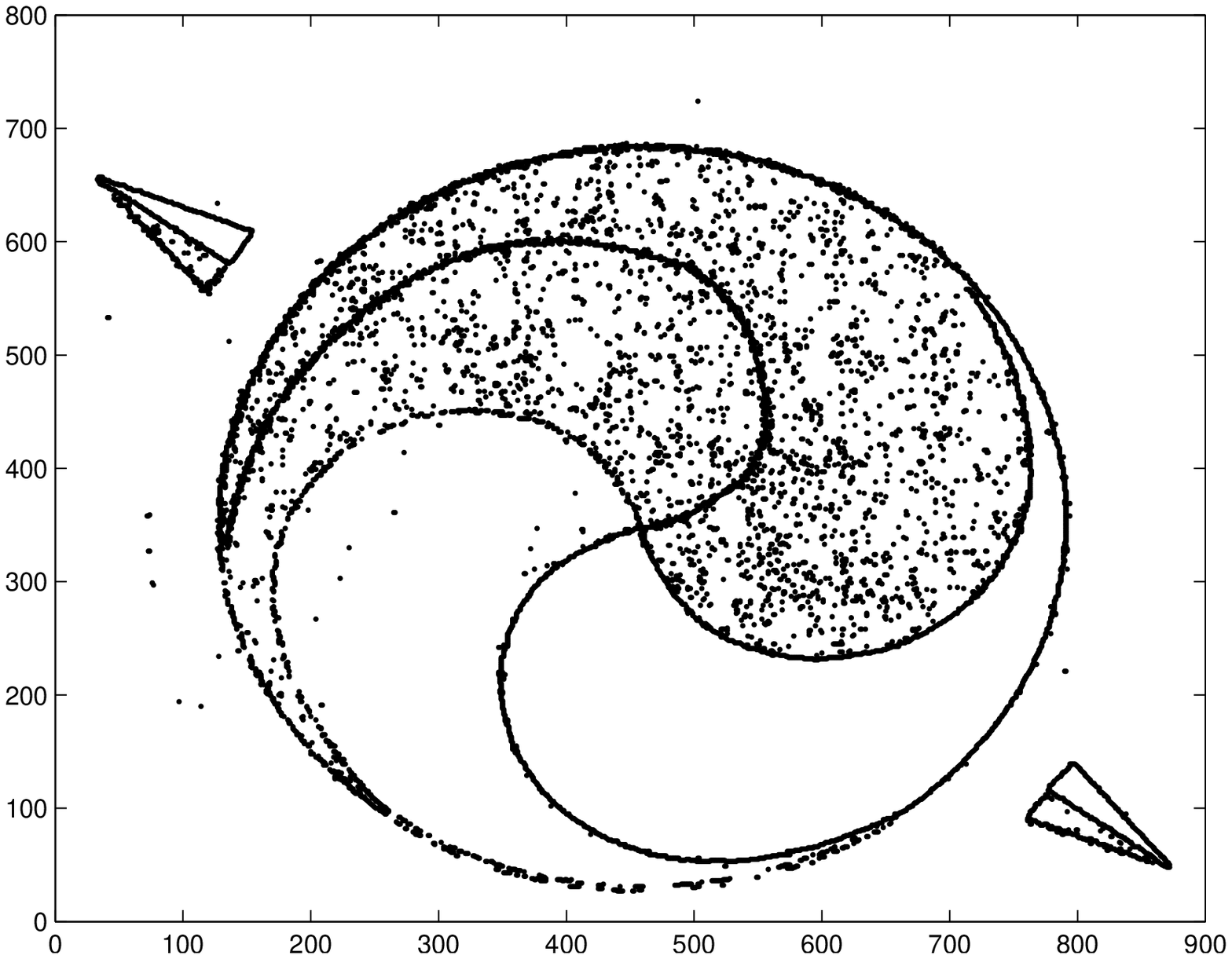,width=4.7cm}
\hspace*{0.2cm}
\epsfig{figure=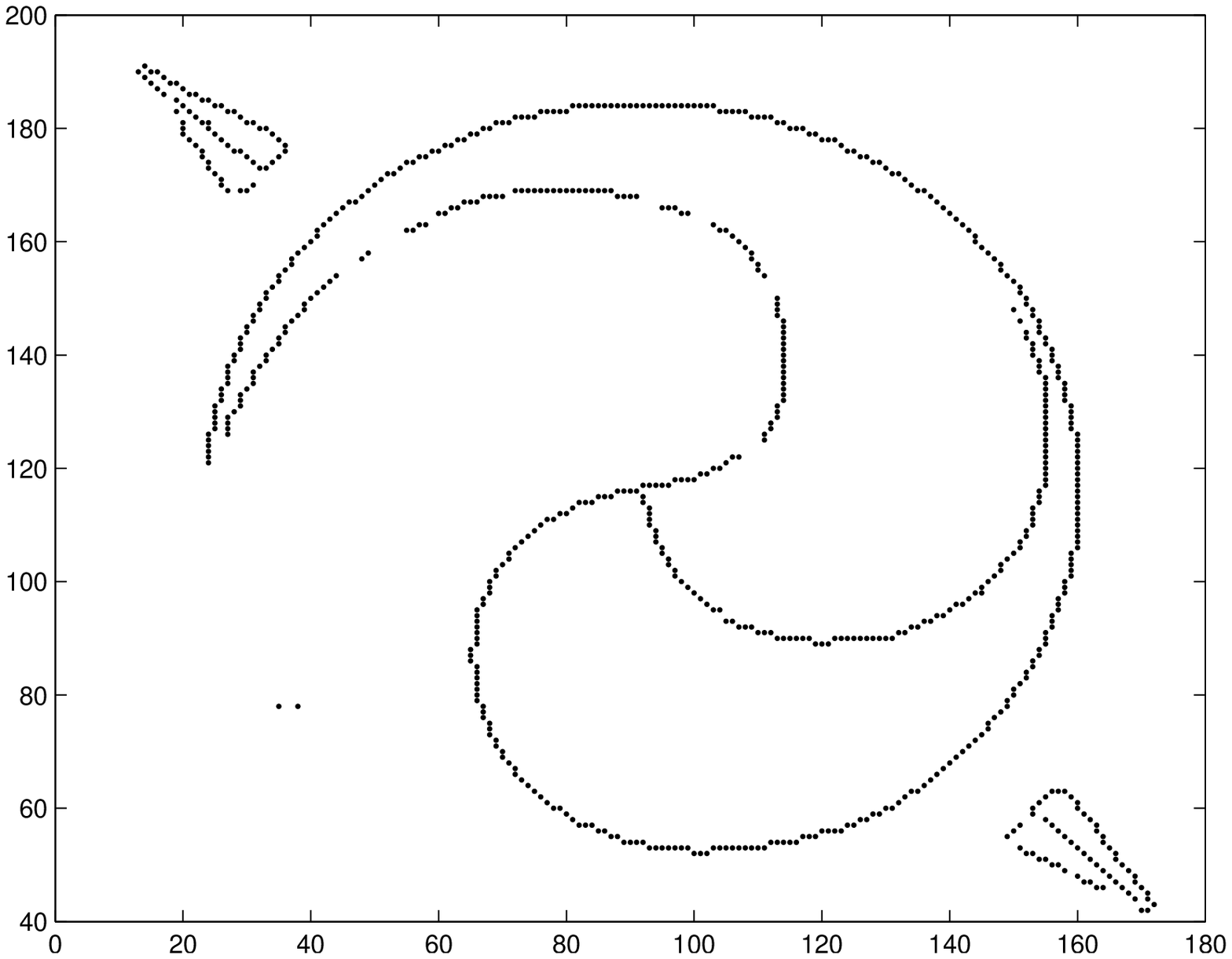,width=4.7cm}}
    \vspace*{-.5cm}\caption{Failures in edge detection. Top: images; bottom: corresponding Canny edge maps~\cite{canny86}.
    Left: image in database;
    middle: photograph with large zoom, for which the perceived texture of the paper
    originates spurious edge points; bottom: photograph at low
    light, for which several edge points are not detected, due to less abrupt
    changes in image intensity.}.
    \label{fig:exedges}
    \end{figure}

\noindent{\bf Edge-detection failures.} Many reasons may motivate
gross failures in edge detection. For example, too many edges may be
detected, due to image noise or, in opposition, too few edges may be
detected, due to the absence of abrupt enough changes in the image
intensity. The examples in Fig.~\ref{fig:exedges} illustrate these
situations. In the top left, there is one of the images in the
trademark database. In the top middle and right, two test images
that are photographs of the same logo. The photograph in the middle
was obtained with a very large zoom (around $20\times$) and the
right one was obtained in low light. Both lead to edge maps that are
very distinct to the one in the database, see the plots in the
bottom row of Fig.~\ref{fig:exedges}. The trademark images in
Fig.~\ref{fig:exedges} are precisely the ones that originated the
miss-classifications in the tenth row of
Table~\ref{tab:class_results_ATR2}. This is explained by the fact
that our ANSIG representation, although invariant to permutation,
translation, rotation, and scale, and robust to sampling density,
can not cope, without further developments, with severe degradations
such as the ones in the edge maps of Fig.~\ref{fig:exedges}.

\begin{figure}[hbt]
\centerline{\epsfig{figure=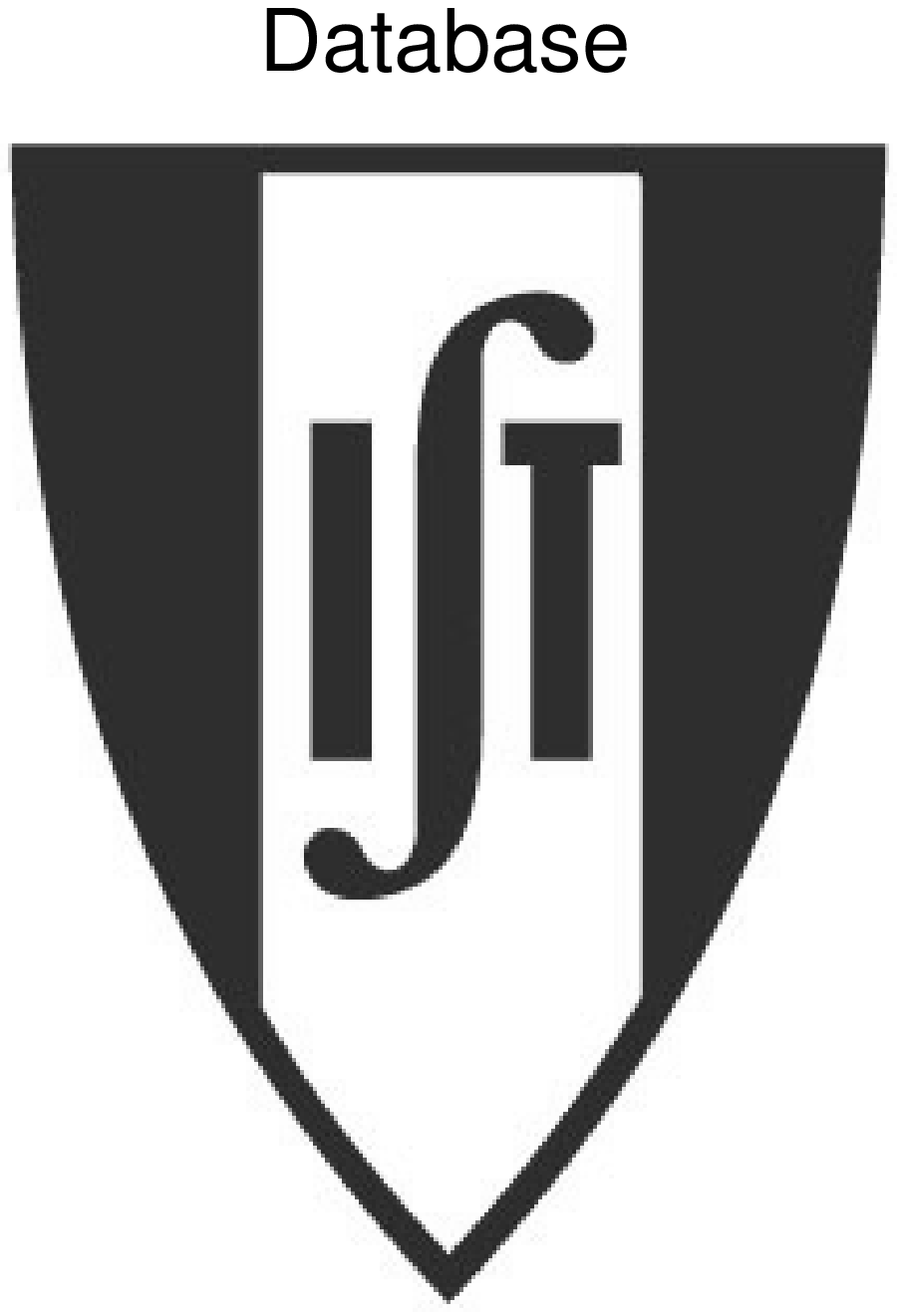,height=3.5cm}
\hspace*{.5cm}\epsfig{figure=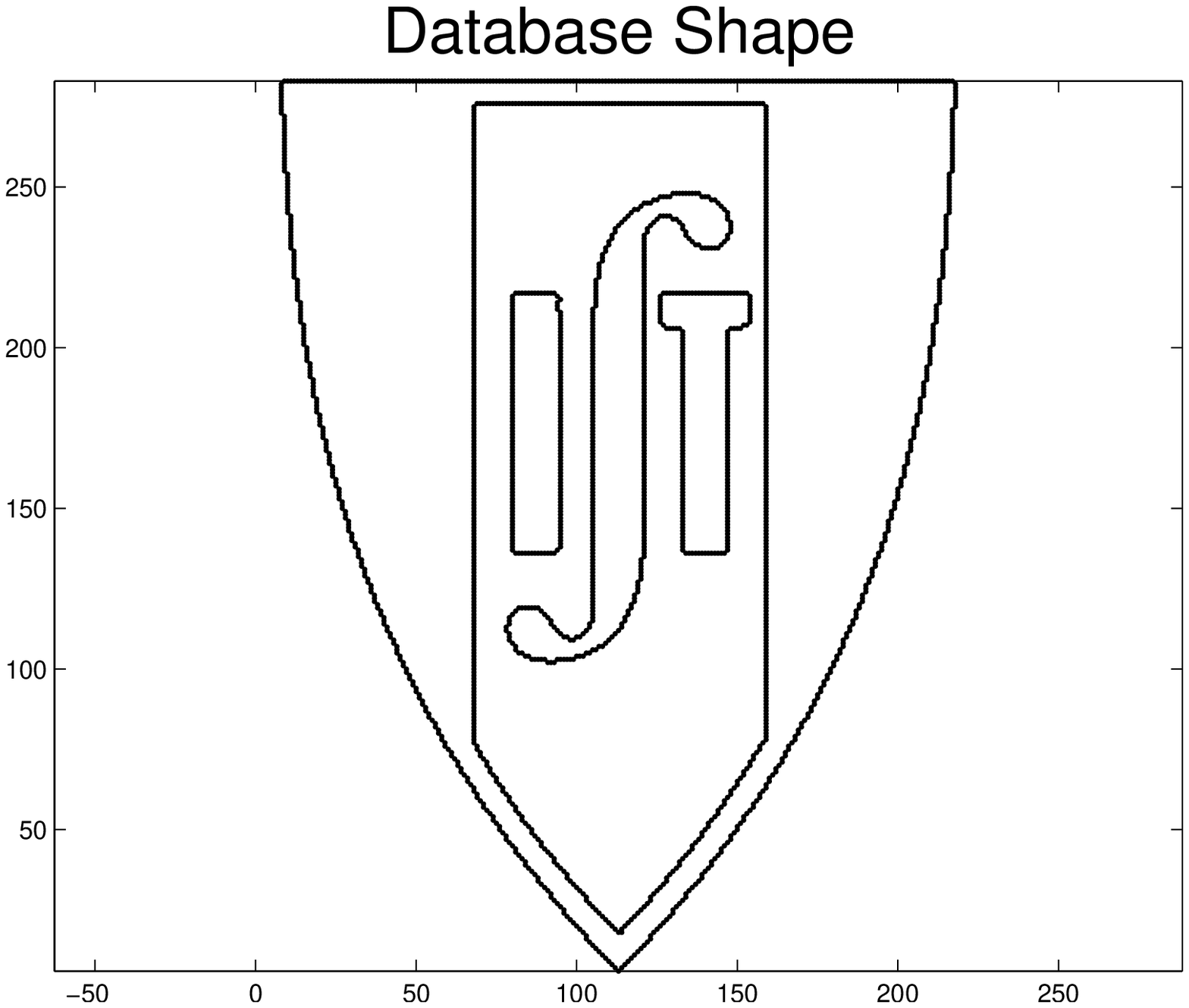,height=3.5cm}
\epsfig{figure=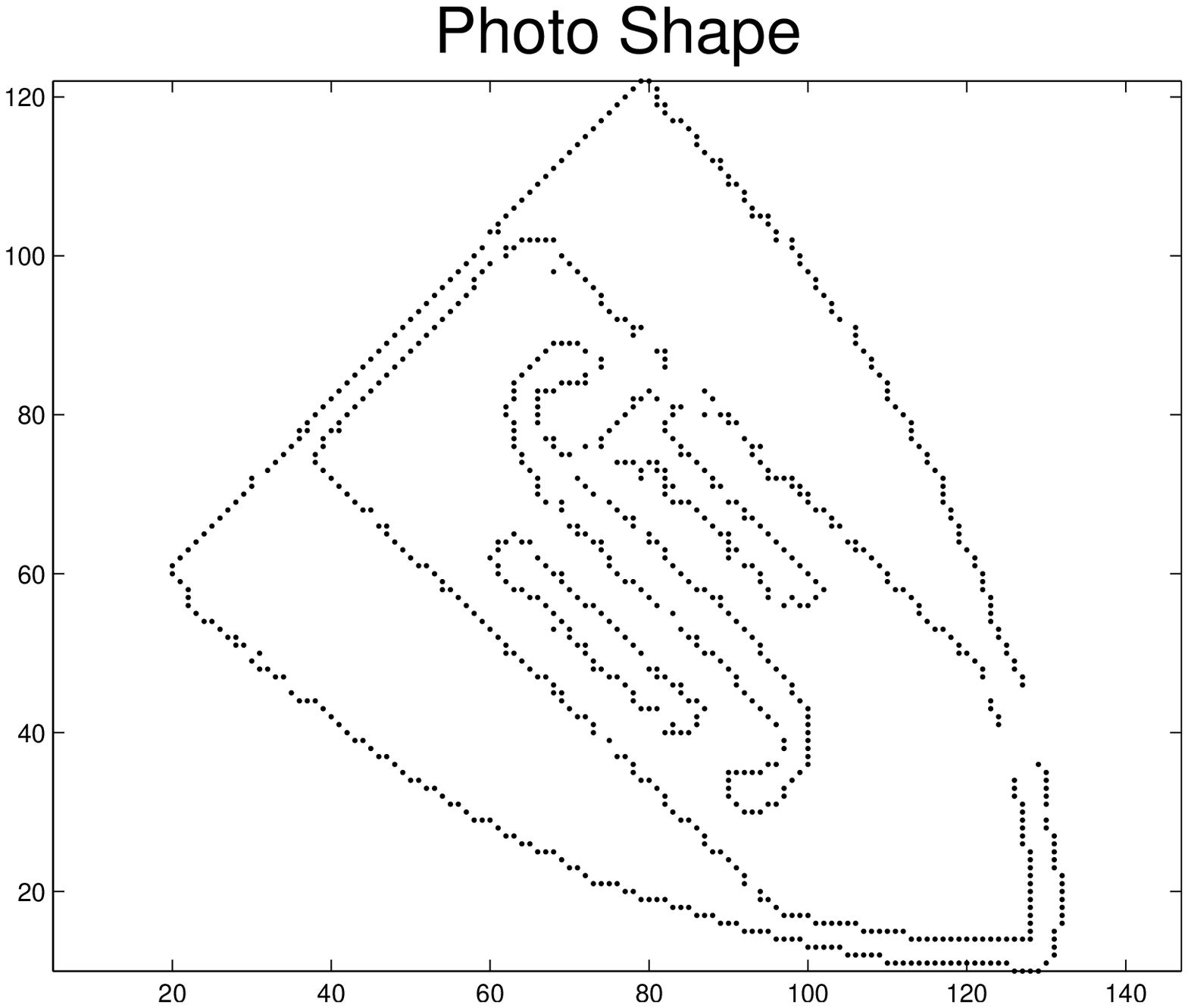,height=3.5cm}
\hspace*{.5cm}\epsfig{figure=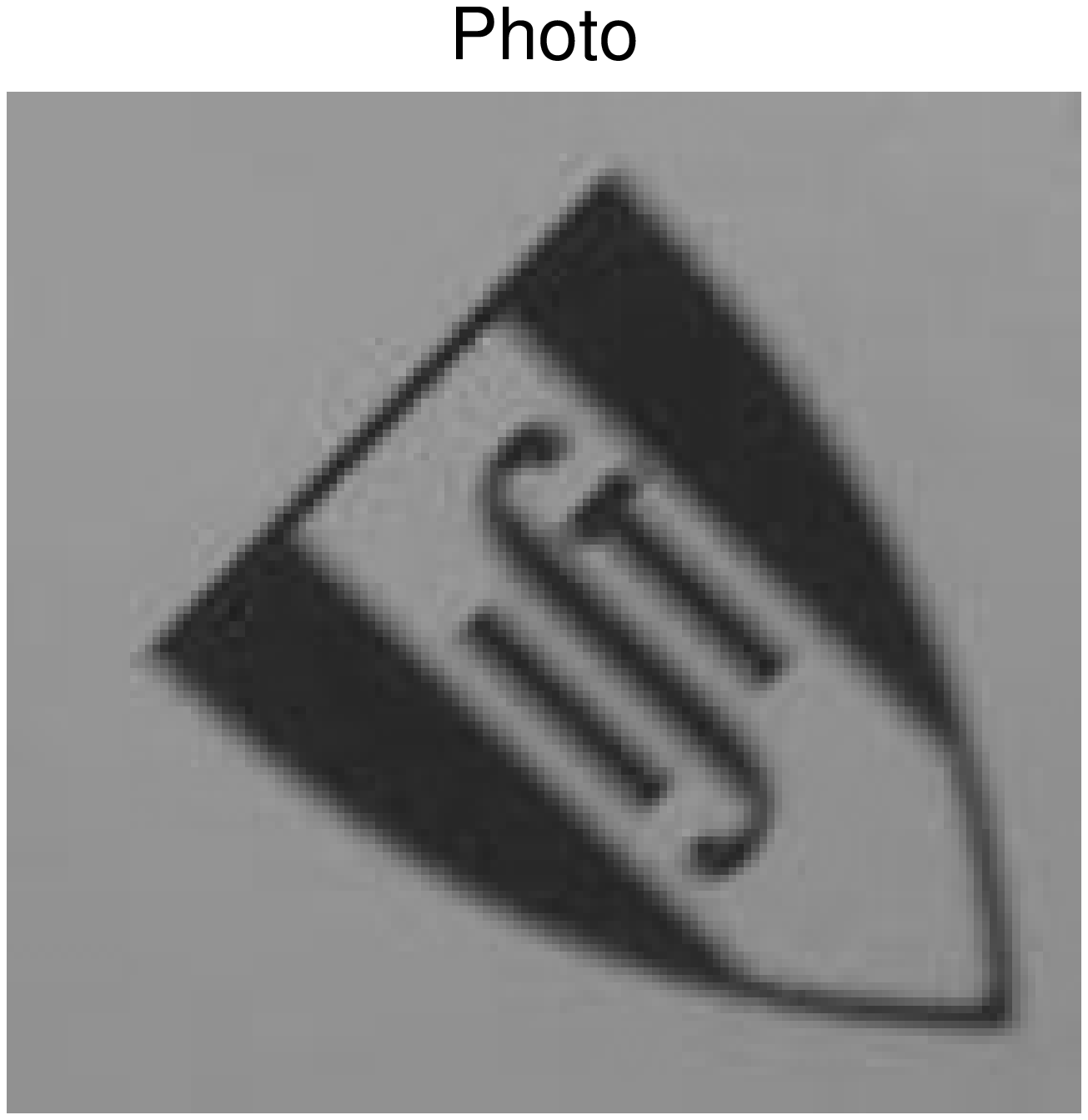,height=3.5cm}
\hspace*{.5cm}} \vspace*{-.5cm}\caption{Recognition of an
out-of-focus image. From left to right: database image,
corresponding edge map, edge map of out-of-focus image, out-of-focus
image. See the ANSIGs of the middle plots in
Fig.~\ref{fig:goodunfocus}.}
    \label{fig:real_algorithm}
    \end{figure}

\begin{figure}[hbt]
\centerline{\epsfig{figure=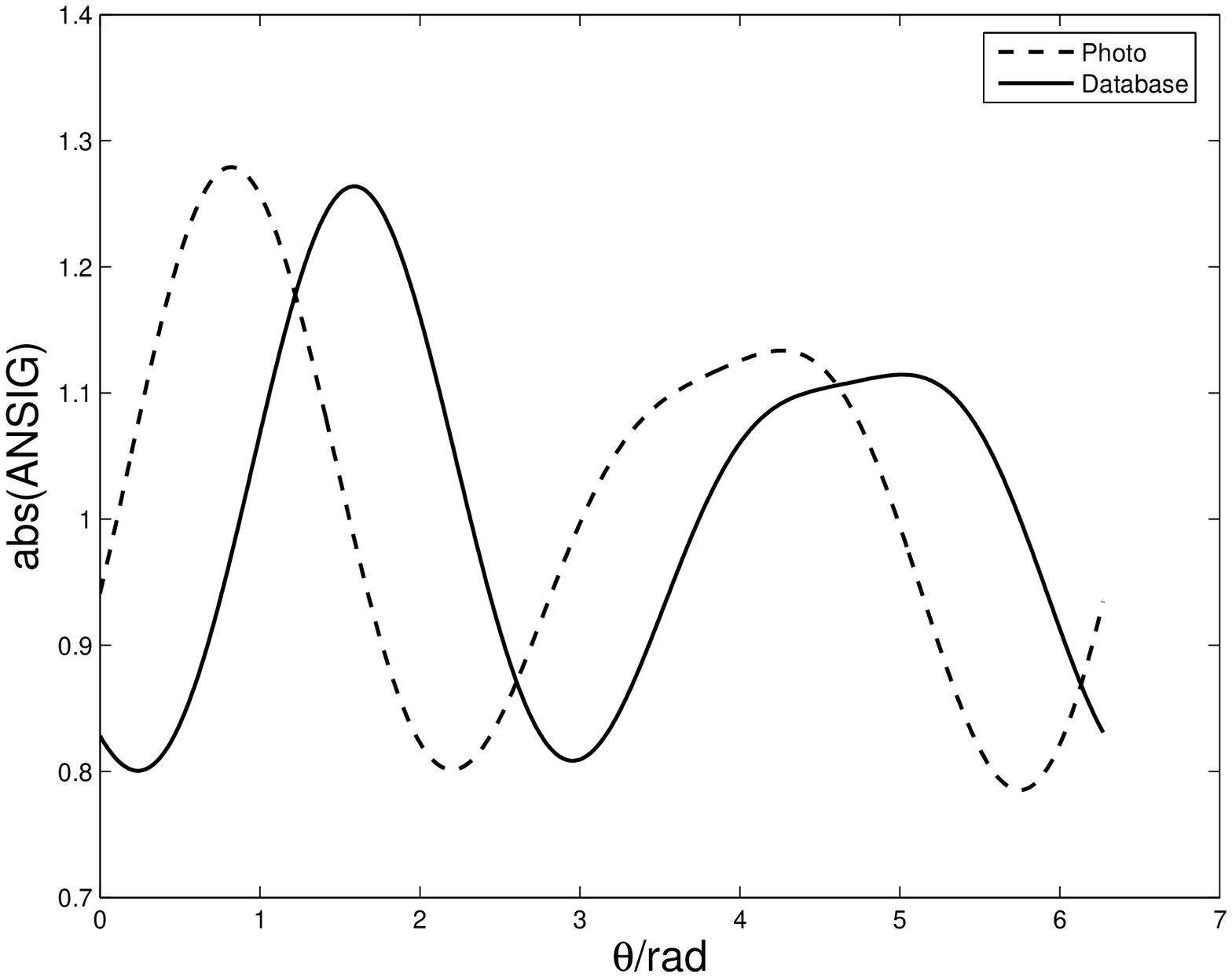,width=5cm}\hspace*{1cm}
\epsfig{figure=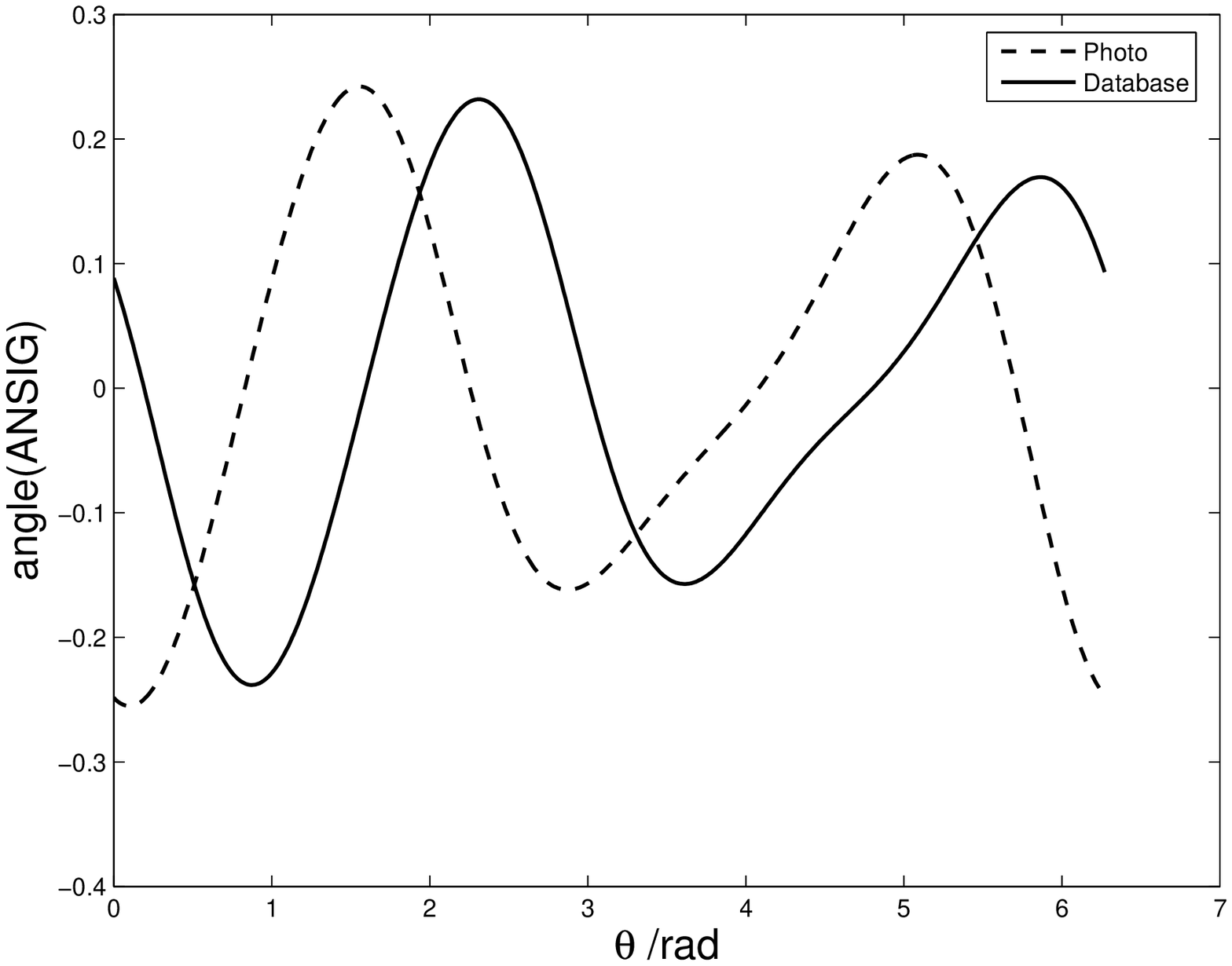,width=5cm}}
\vspace*{-.5cm}\caption{ANSIGs of the shapes in
Fig.~\ref{fig:real_algorithm}. In spite of the unfocused image that
lead to the incomplete edge map, the ANSIGs are very similar (recall
that the circular shift is due to different image orientation and it
is easily taken care of).}
    \label{fig:goodunfocus}
    \end{figure}

\begin{figure}[hbt]
\centerline{\hspace*{-.75cm}\epsfig{figure=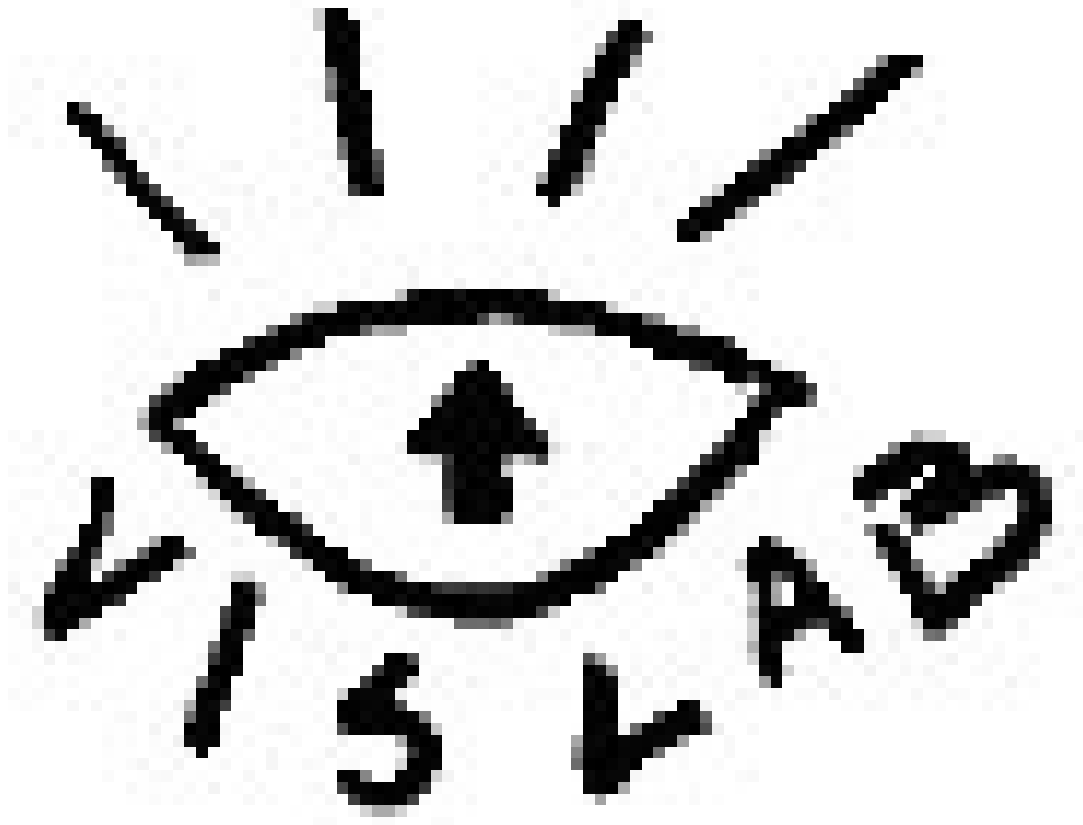,width=3.5cm}
\hspace*{.25cm}\epsfig{figure=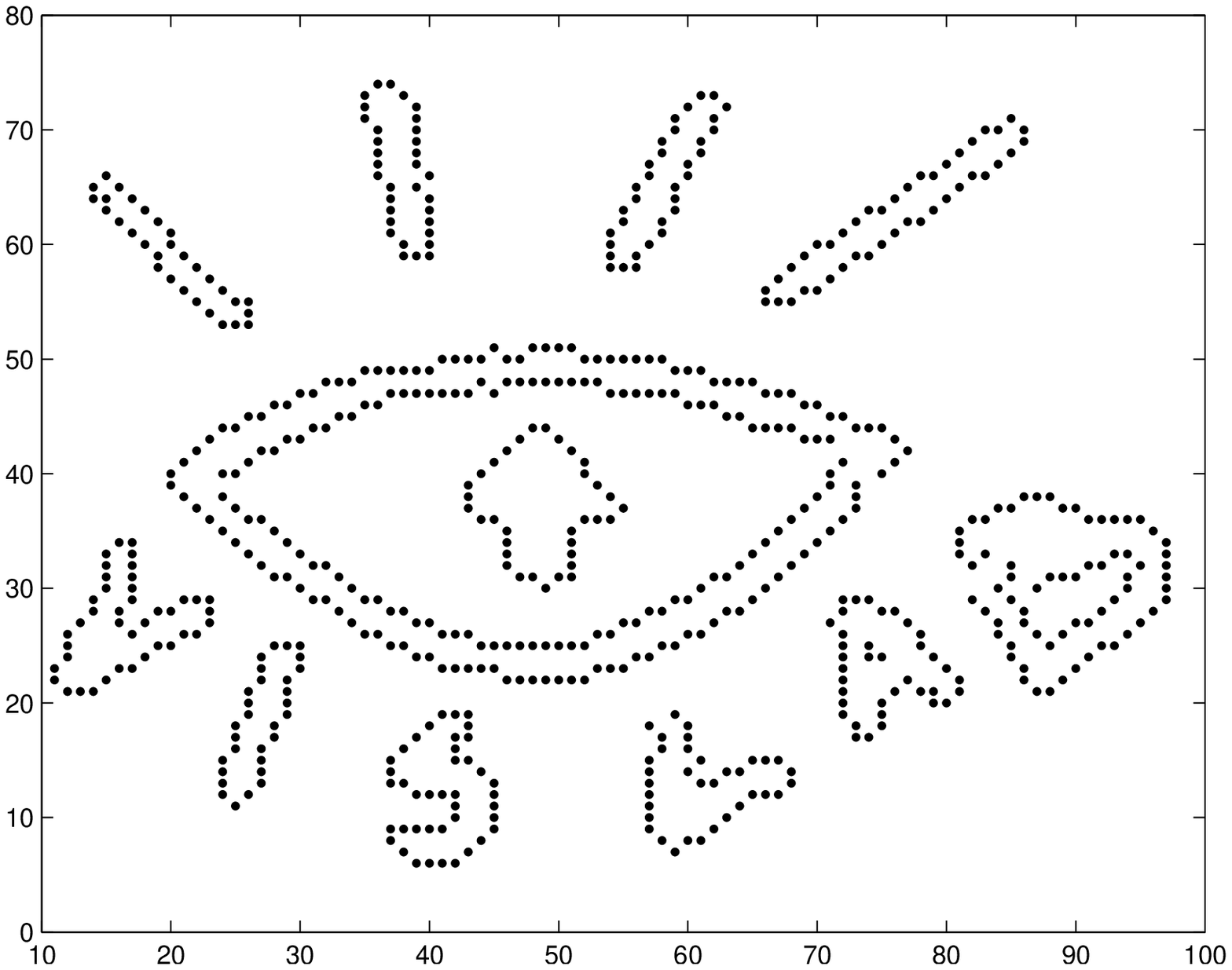,width=3.75cm}
\hspace*{.25cm}\epsfig{figure=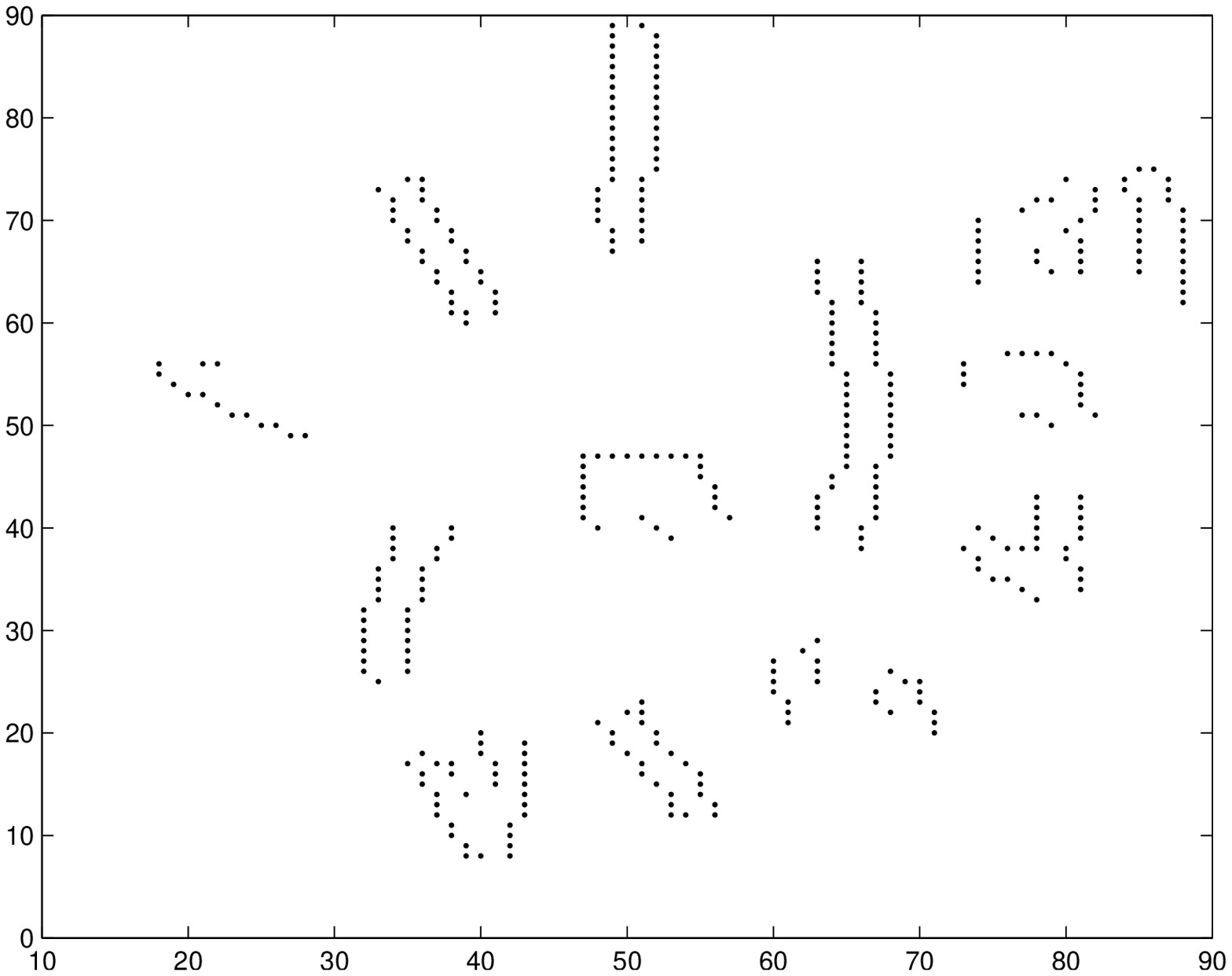,width=3.75cm}
\hspace*{.25cm}\epsfig{figure=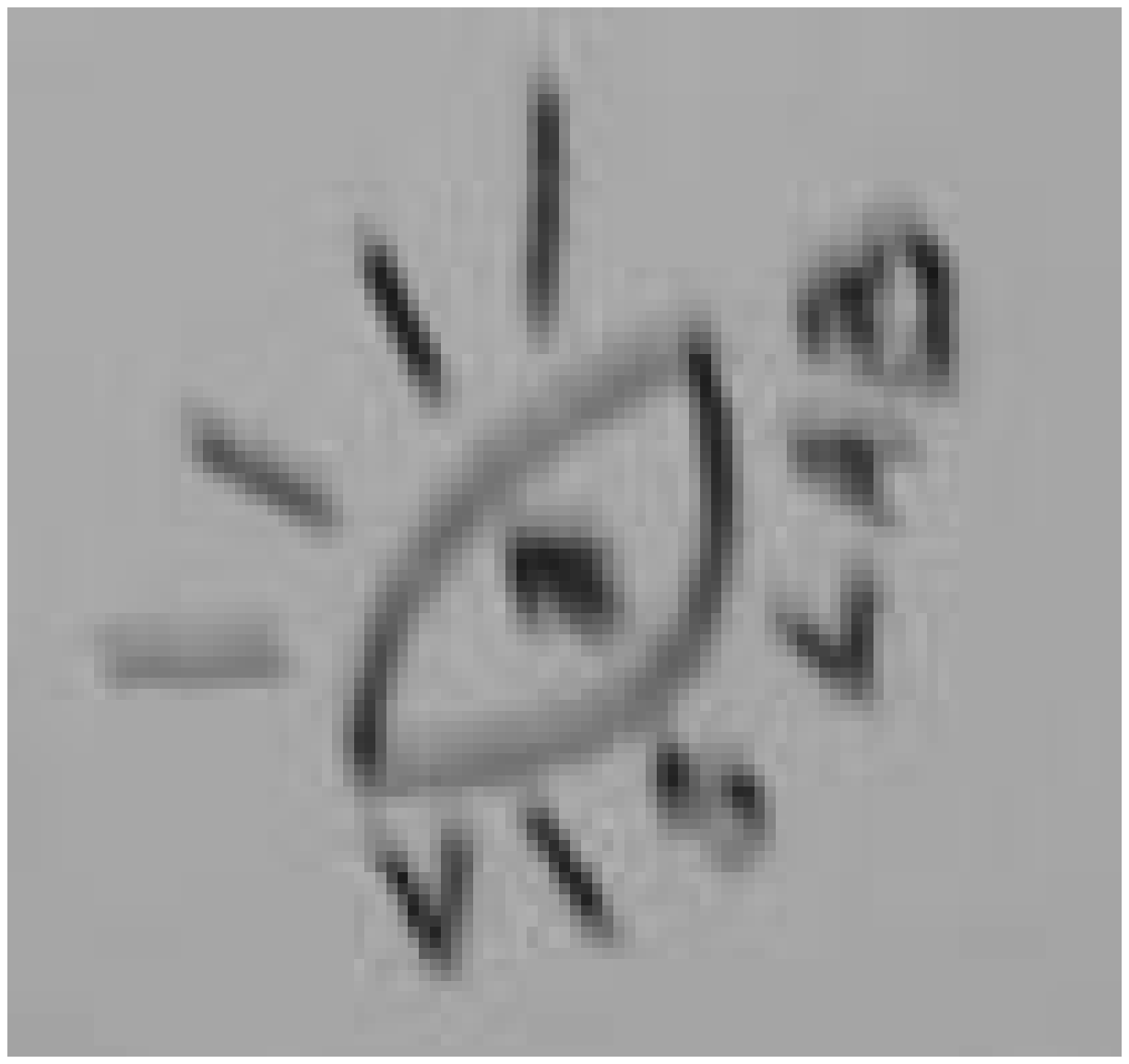,width=3.25cm}}
\vspace*{-.5cm}\caption{Failure in recognizing an out-of-focus
image, due to the highly incomplete edge map.}
    \label{fig:Bad_3}
    \end{figure}

Edge detection failures may also happen when processing out-of-focus
photographic images. Figs.~\ref{fig:real_algorithm}
and~\ref{fig:Bad_3} present two extreme cases. For the out-of-focus
test image of Fig.~\ref{fig:real_algorithm}, some edge points are
missed, but the corresponding ANSIG, represented in
Fig.~\ref{fig:goodunfocus}, results similar to the one in the
database, leading to a correct classification. This is due to the
robustness of the ANSIG representation in what respects to shape
sampling density, recall the derivation in
Section~\ref{sec:geometric} and its illustration in
Figs.~\ref{fig:sampling} and~\ref{fig:sampling2}. In opposition, the
out-of-focus test image of Fig.~\ref{fig:Bad_3} results very smooth,
originating an highly incomplete edge map (compare the middle plots
of Fig.~\ref{fig:Bad_3}). This is the reason for the classification
error in the fifth row of Table~\ref{tab:class_results_ATR2}.

%

\begin{figure}[hbt]
\centerline{\hspace*{-.75cm}\epsfig{figure=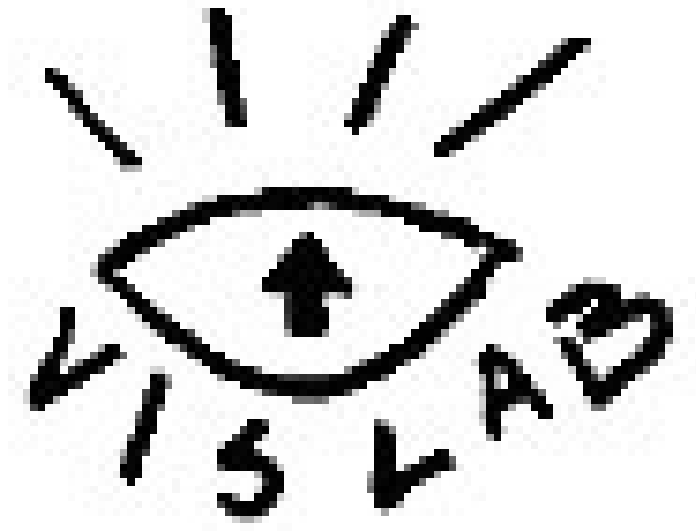,width=3.5cm}
\hspace*{.25cm}\epsfig{figure=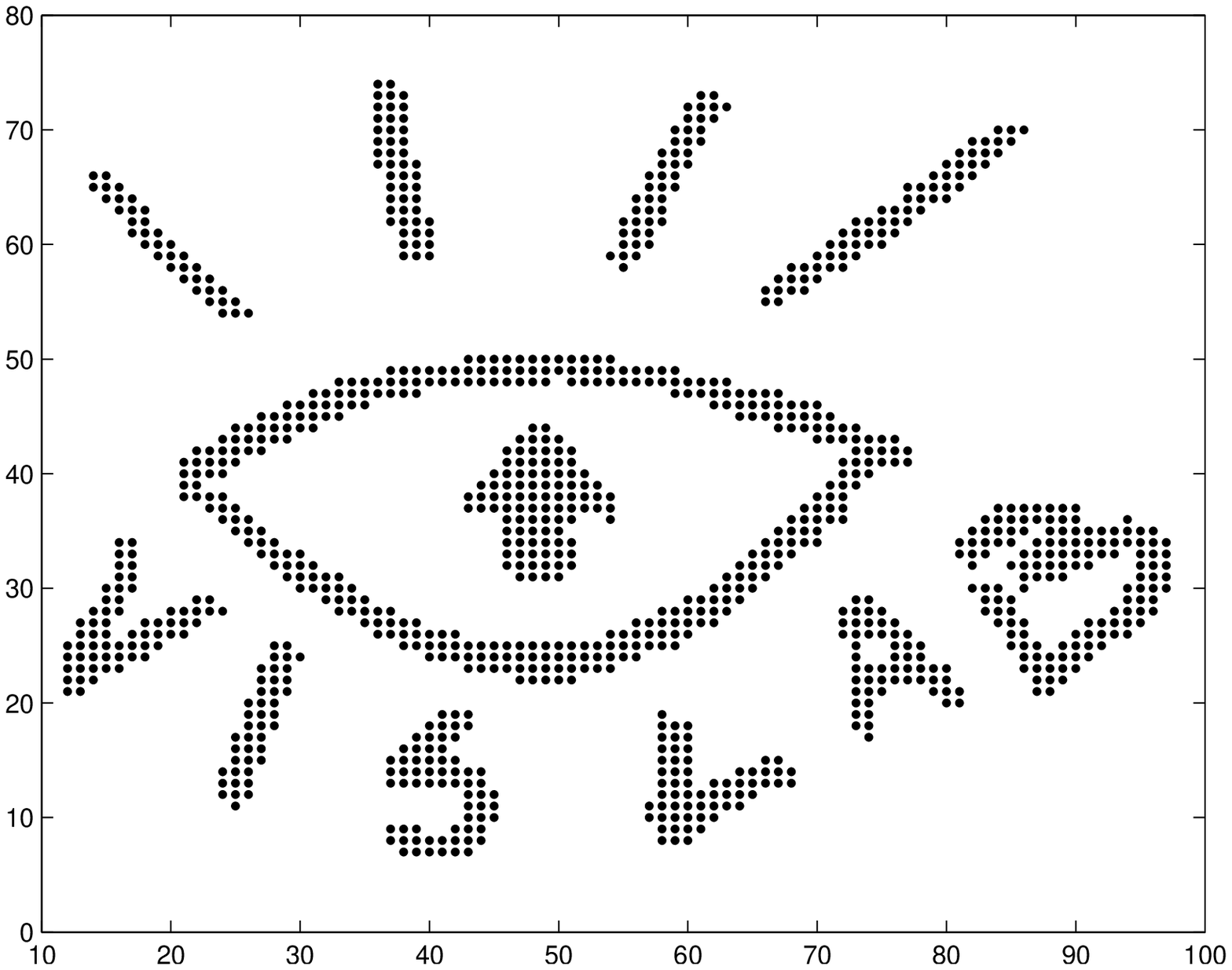,width=3.75cm}
\hspace*{.25cm}\epsfig{figure=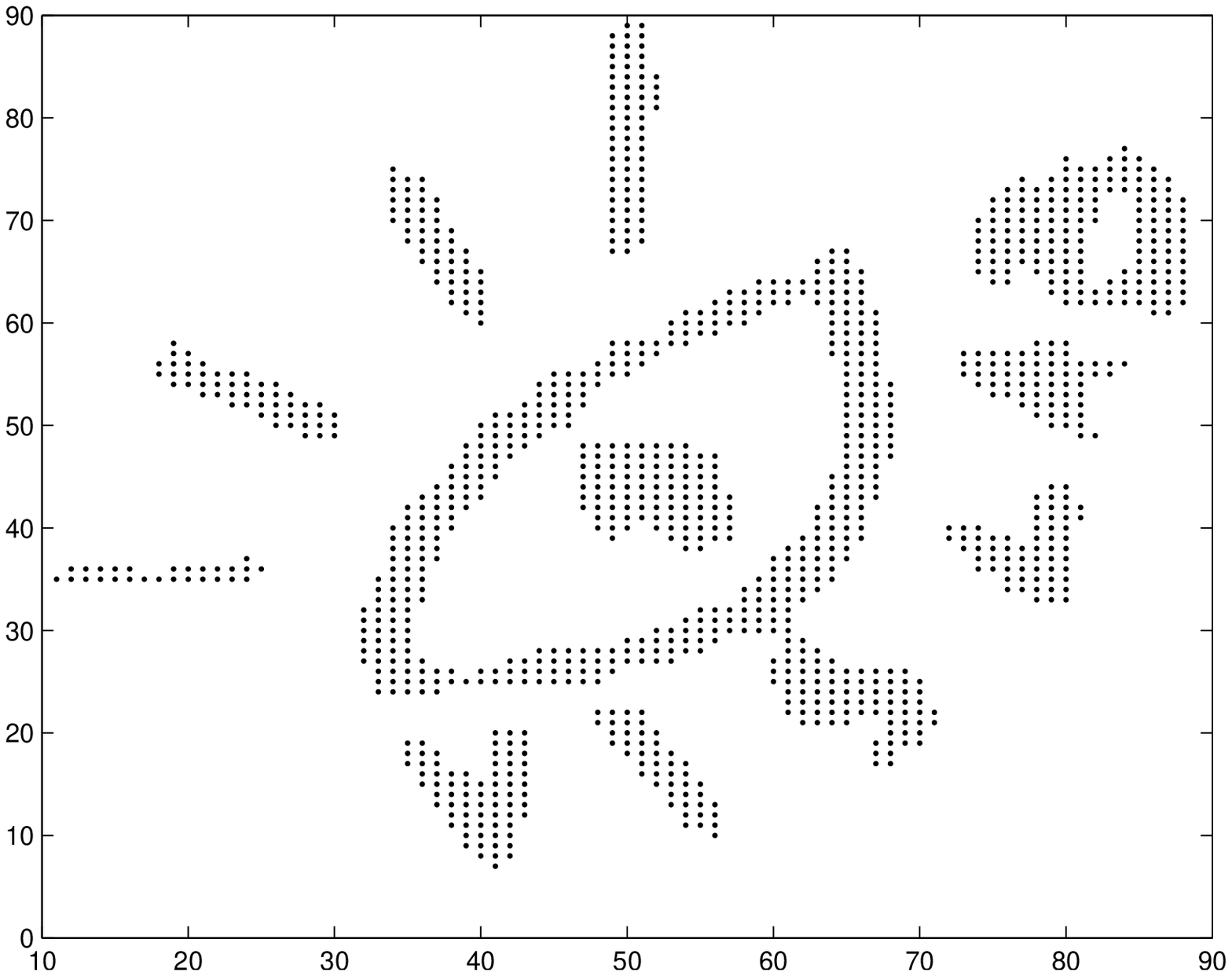,width=3.75cm}
\hspace*{.25cm}\epsfig{figure=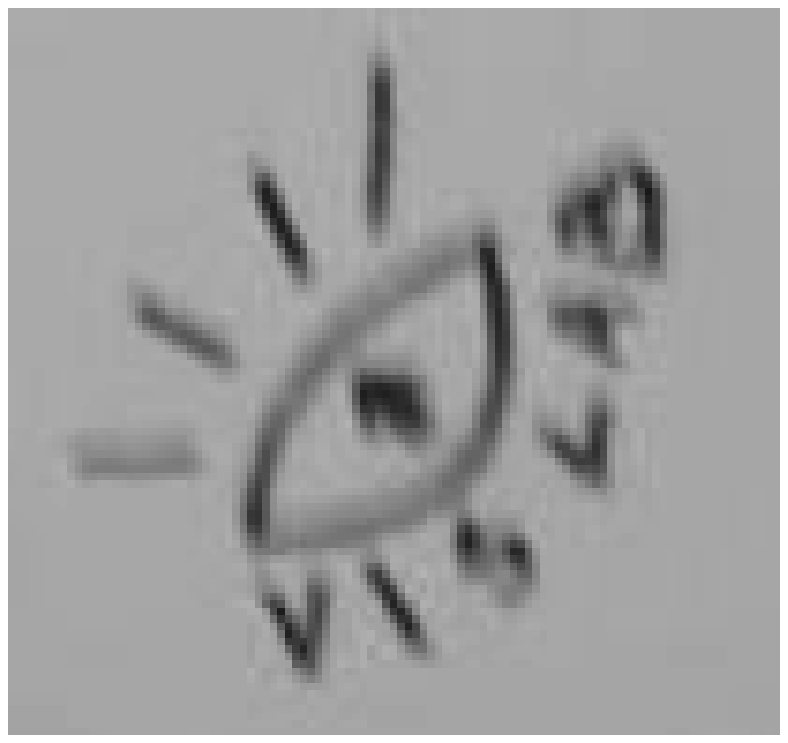,width=3.25cm}}
\vspace*{-.5cm}\caption{Comparing the same pair of images of
Fig.~\ref{fig:Bad_3}, now using intensity thresholding as the
pre-processing step, leads to better results. See also the ANSIGs of
the middle plots in Fig.~\ref{fig:ansigthres}.}
    \label{fig:thres}
    \end{figure}

\begin{figure}[hbt]
\centerline{\epsfig{figure=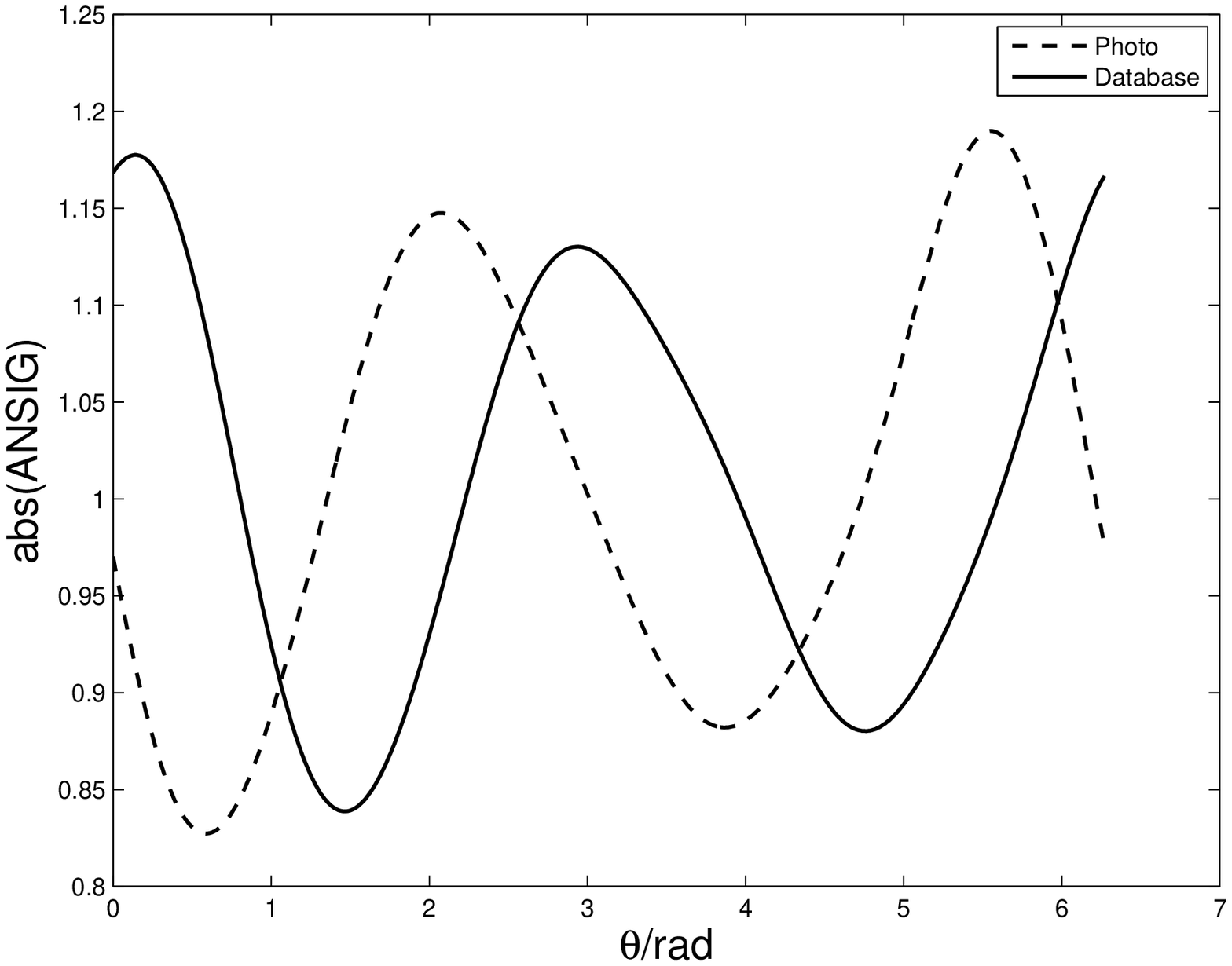,width=5cm}\hspace*{1cm}
\epsfig{figure=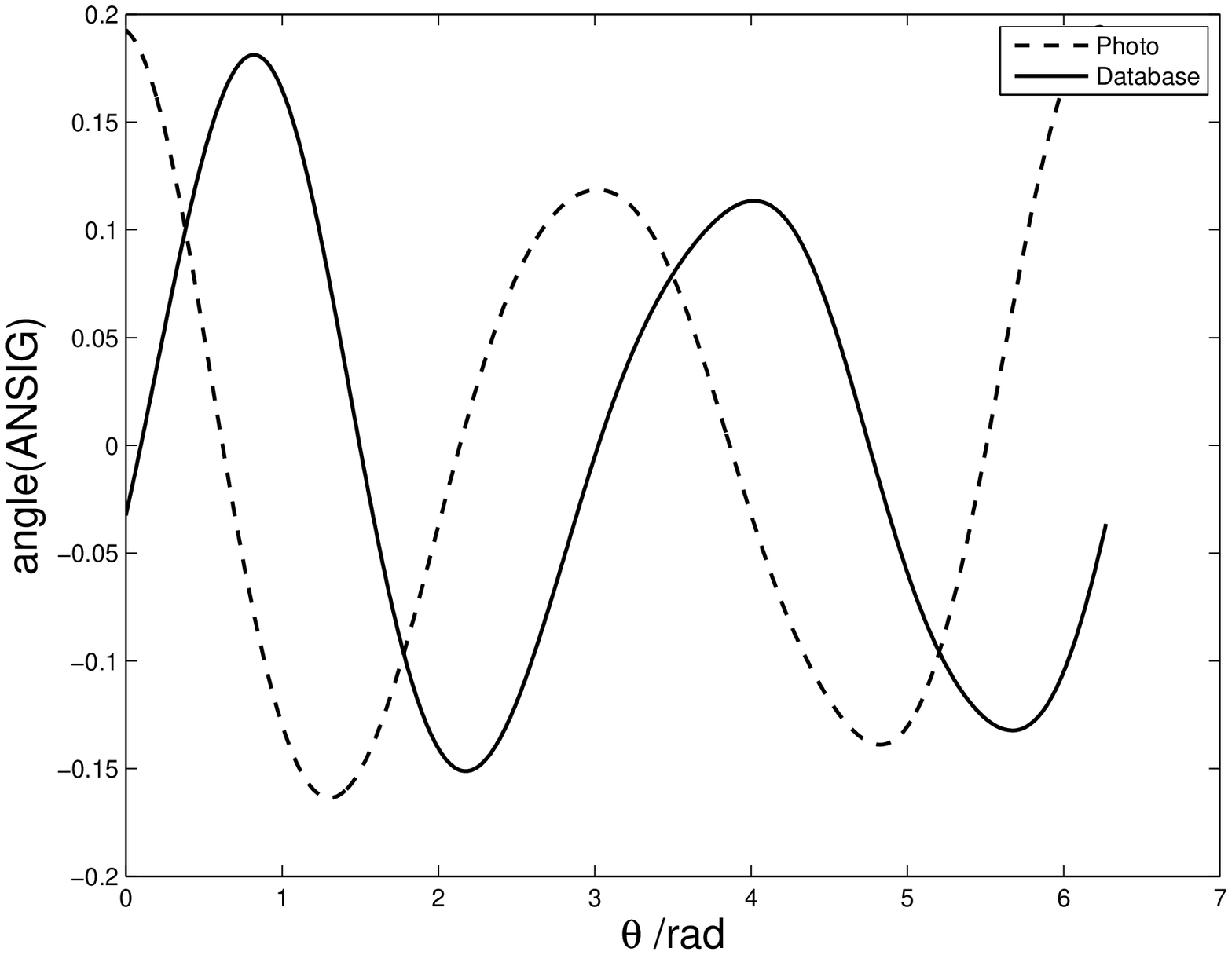,width=5cm}}
\vspace*{-.5cm}\caption{ANSIGs of the shapes in
Fig.~\ref{fig:thres}. In spite of the unfocused image, the ANSIGs
are very similar (again, the circular shift is due to different
image orientation and it is easily taken care of).}
    \label{fig:ansigthres}
    \end{figure}

We emphasize, however, that different pre-processing schemes may
easily improve the results. In fact, one of the advantages of the
shape representation scheme we propose in this paper is precisely
the fact that it does not require edge segments, dealing equally
well with shapes described by {\it arbitrary} sets of points. As an
example, we extracted different shape vectors from the pair of
images in Fig.~\ref{fig:Bad_3}, by using a simple intensity
threshold, see the resulting shapes in the middle plots of
Fig.~\ref{fig:thres}. Since thresholding is much more insensitive to
focusing effects, the shapes result more similar and the ANSIG
successively captures this similarity, see
Fig.~\ref{fig:ansigthres}.

\begin{figure}[hbt]
\centerline{\epsfig{figure=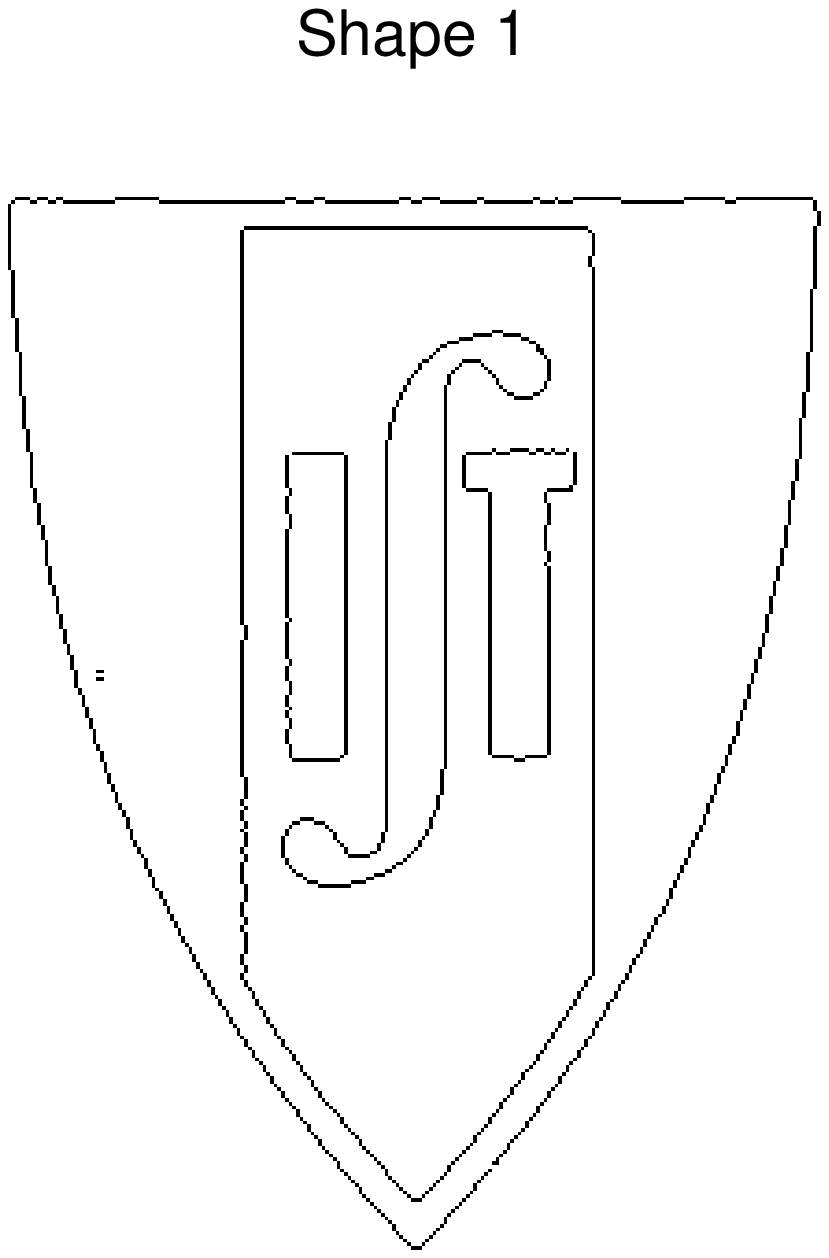,height=4cm}\hspace*{.5cm}
\epsfig{figure=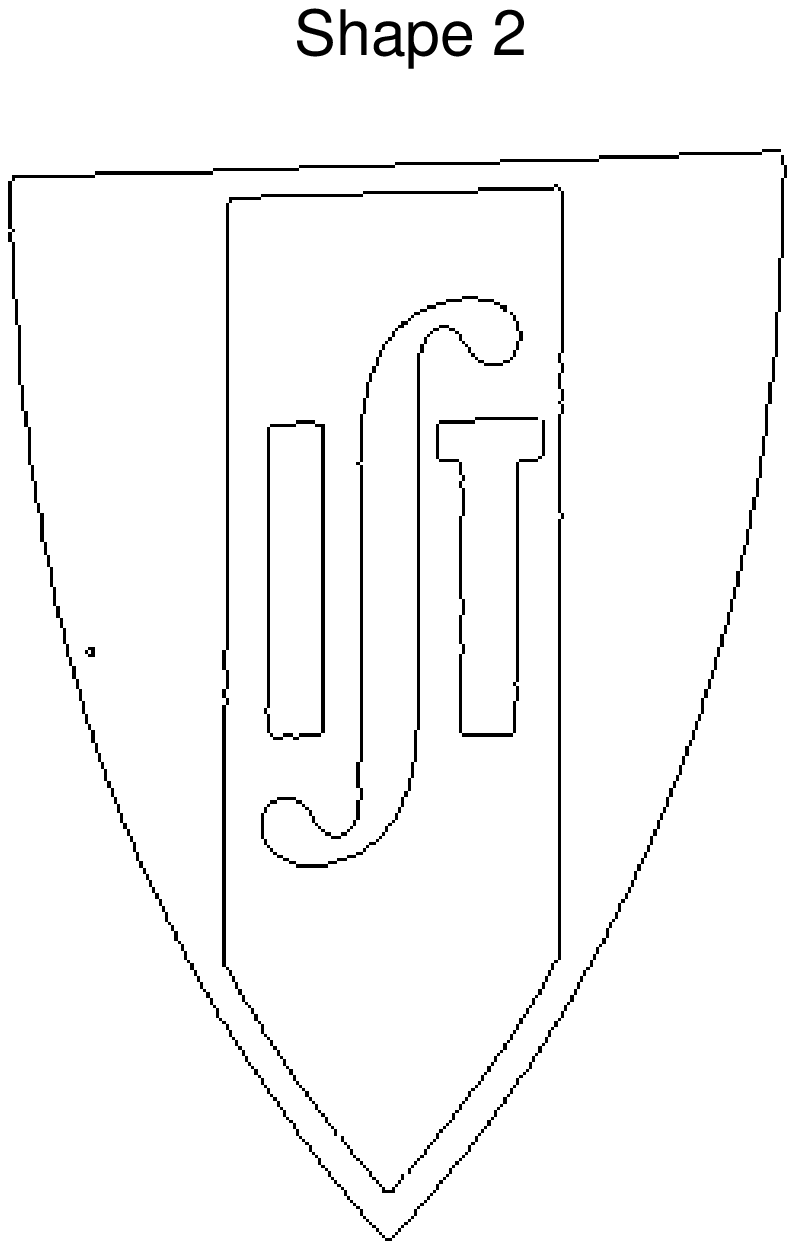,height=4cm}\hspace*{.5cm}
\epsfig{figure=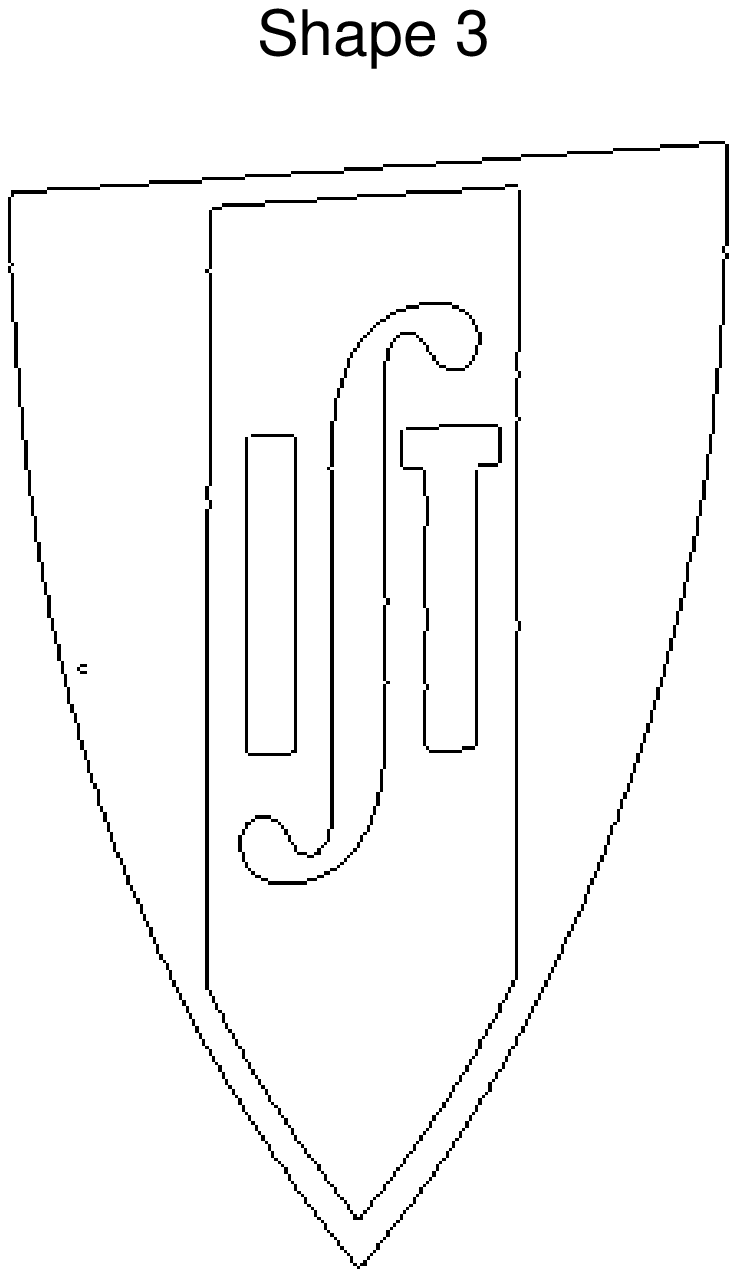,height=4cm}\hspace*{.5cm}
\epsfig{figure=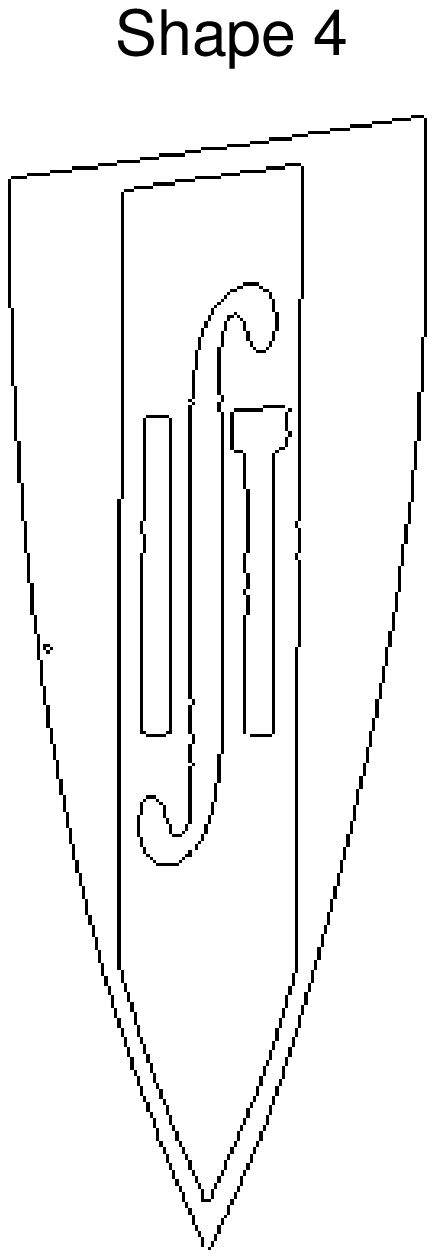,height=4cm}}
    \vspace*{-.5cm}\caption{Edge maps obtained from photographs with increasing perspective distortion.}
    \label{fig:Perspective_1}
    \end{figure}

\begin{figure}[hbt]
    \centerline{\epsfig{figure=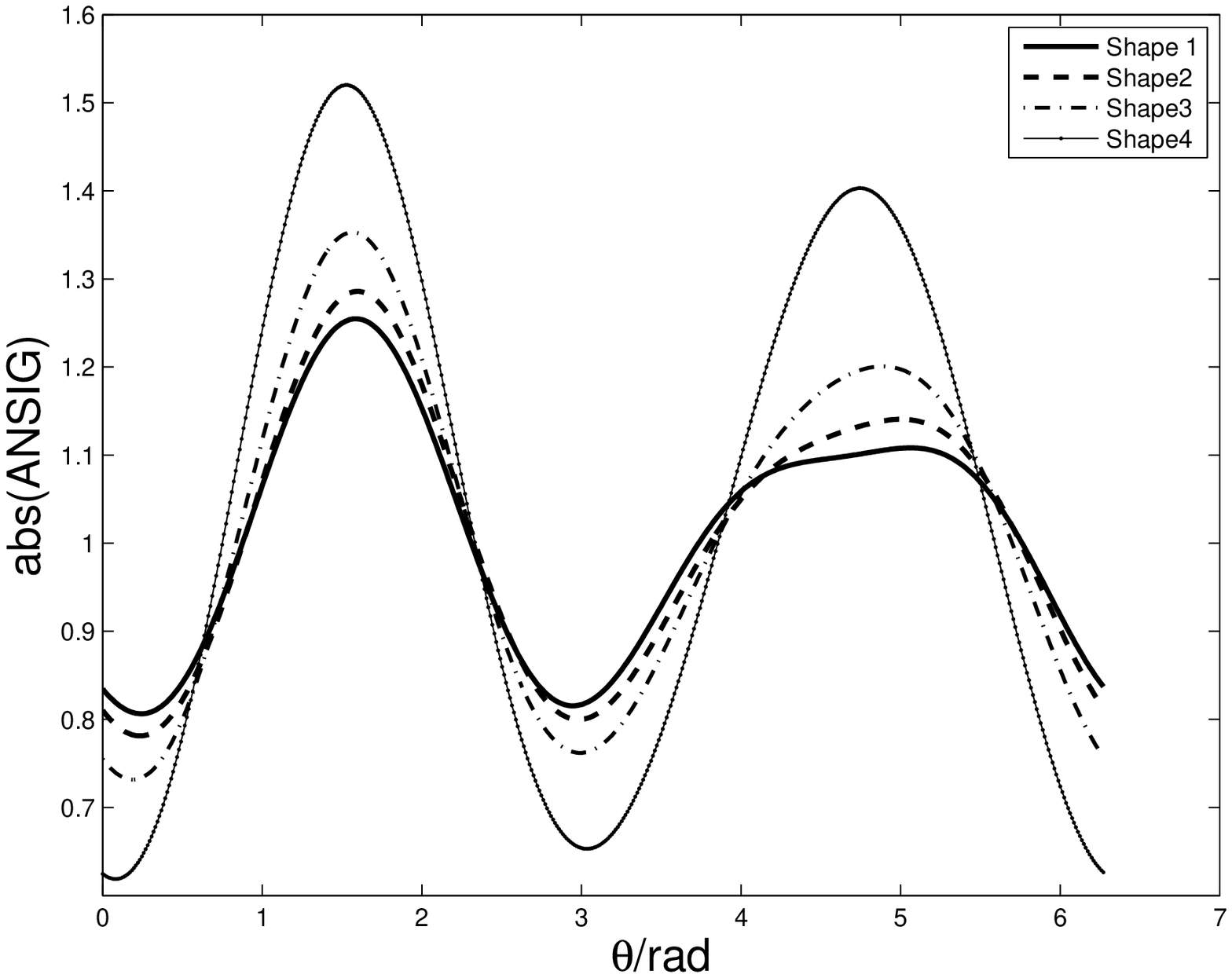,width=7cm}\hspace*{1cm}
    \epsfig{figure=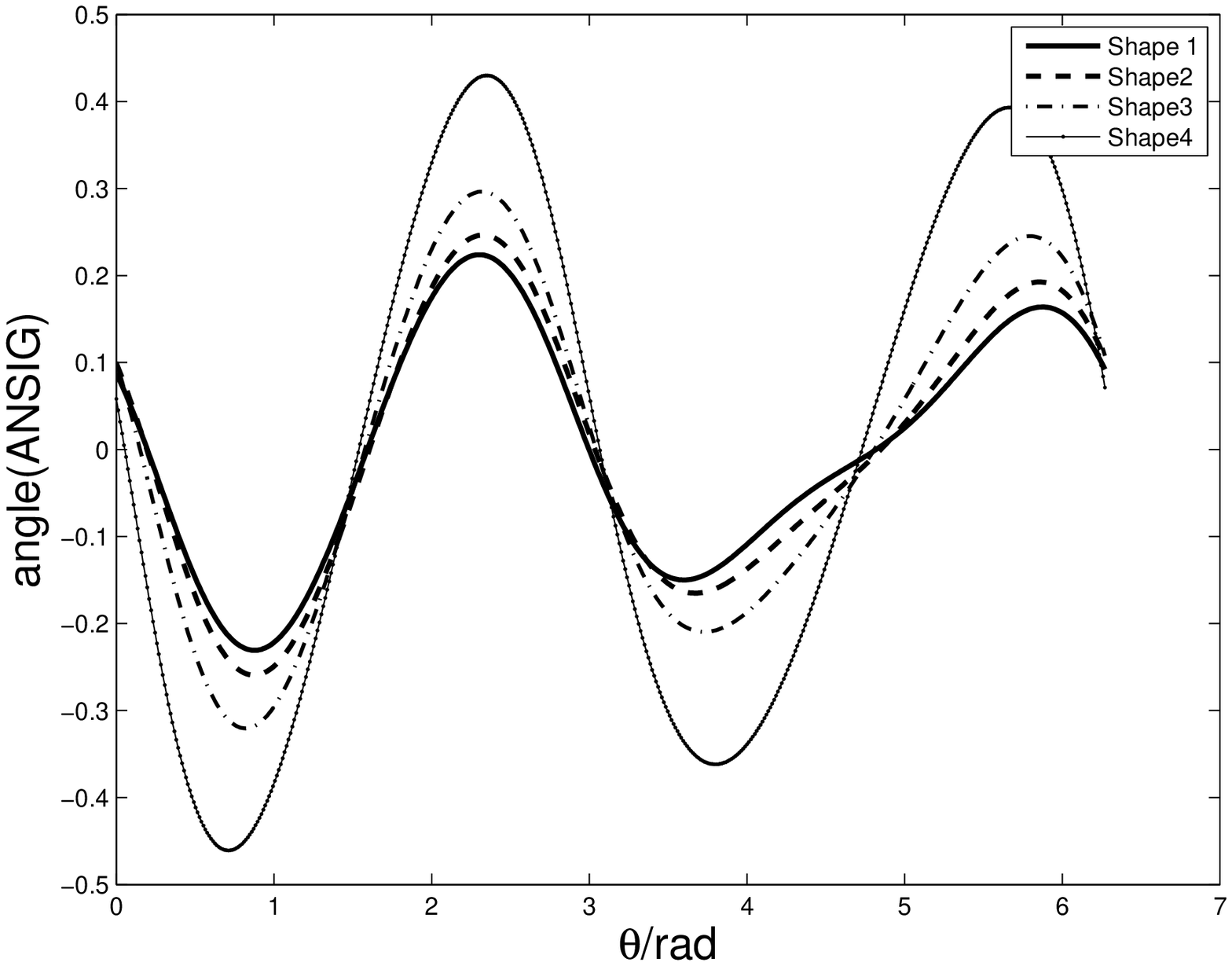,width=7cm}}
\vspace*{-.5cm}\caption{ANSIGs of the perspectively distorted shapes
in Fig.~\ref{fig:Perspective_1}.}
    \label{fig:Perspective_2}
    \end{figure}

\noindent{\bf Perspective distortion.} A different kind of model
violation that may occur when dealing with 2D shapes extracted from
photographic images is due to perspective distortion. In fact, while
the ANSIG representation was developed assuming the shapes are
2D-rigid, this may not be the case, even when only flat objects are
considered, if the camera is not adequately oriented. To illustrate
the behavior of the ANSIG representation when in presence of
perspective distortions, we photographed one of the trademark
images, purposely from directions not perpendicular to the paper.
Some of the corresponding edge maps are shown in
Fig.~\ref{fig:Perspective_1}, with perspective distortion increasing
from left to right. The plots in Fig.~\ref{fig:Perspective_2}
compare the ANSIG of the leftmost (undistorted) shape of
Fig.~\ref{fig:Perspective_1}  with the ANSIGs of the perspectively
distorted ones. We can see that, although this effect was not taken
into account in our modeling, the ANSIG representation can deal with
small perspective distortions (see the similarity of the thick
solid, dashed, and dot-dashed lines in the plots of
Fig.~\ref{fig:Perspective_2}). Naturally, when the distortions are
severe, our representation fails to adequately capture the shape
similarity (see the thin solid lines in the plots of
Fig.~\ref{fig:Perspective_2}).

\section{Conclusion}
\label{sec:conc}

We proposed a new method to represent 2D~shapes, described by a set
of unlabeled points, or landmarks, in the plane. Our method is based
on what we call the analytic signature (ANSIG) of the shape, whose
most distinctive characteristic is its invariance to the way the
landmarks are labeled. This makes the ANSIG particularly suited to
cope with shapes described by large sets of edge points in images.
We illustrated its performance in shape-based classification tasks.

We envisage paths for future research and development based on the
ANSIG representation. In this paper, we store ANSIGs by sampling
them on the unit-circle. A topic that deserves further study is not
only the choice of sampling rate but also the adoption of different
sampling schemes, {\it e.g.}, the use of two or more concentric
circles for robustness. The derivations in this paper are targeted
to the representation and comparison of complete shapes. However, in
many practical scenarios, it is also necessary to recognize a set of
points as being a part, {\it i.e.}, a subset of a given shape. A
good challenge is then to adapt the ANSIG representation to deal
with incomplete shapes. Finally, in our experiments, shapes are
obtained directly from the (noisy) output of the edge detection
process. Naturally, intermediate processing steps, {\it e.g.}, the
popular morphological filtering operations, would lead to ``cleaner"
shapes, thus to more accurate classifications.

\bibliographystyle{IEEEtran}
\bibliography{perminv}

\end{document}